%% file: arxiv.tex
\documentclass[10pt,twocolumn,letterpaper]{article}

\usepackage{cvpr}              

\input{preamble}

\definecolor{cvprblue}{rgb}{0.21,0.49,0.74}
\usepackage[pagebackref,breaklinks,colorlinks,allcolors=cvprblue]{hyperref}

\def\paperID{3037} 
\def\confName{CVPR}
\def\confYear{2026}

\title{Registration-Free Learnable Multi-View Capture of Faces\\ in Dense Semantic Correspondence\vspace{-0.4cm}
}

\author{
\normalsize{Panagiotis P. Filntisis\textsuperscript{1,3,4,\dag} \quad
George Retsinas\textsuperscript{1} \quad
Radek Dane\v{c}ek\textsuperscript{4} \quad} \\
\normalsize{Vanessa Sklyarova\textsuperscript{4, 5} \quad
Petros Maragos\textsuperscript{1,2,3,\dag} \quad
Timo Bolkart\textsuperscript{6}\footnotemark}\\
\vspace{-0.3cm}
\\
\footnotesize{\textsuperscript{1}Institute of Robotics, Athena Research Center, 15125 Maroussi, Greece\vspace{-0.1cm}}\\
\footnotesize{\textsuperscript{2}School of ECE, NTUA, Greece\vspace{-0.1cm}} \quad
\footnotesize{\textsuperscript{3}HERON -- Hellenic Robotics Center of Excellence, Athens, Greece}\\
\footnotesize{\textsuperscript{4}MPI for Intelligent Systems, T\"ubingen, Germany} \quad
\footnotesize{\textsuperscript{5} ETH Zurich, Zurich, Switzerland} \quad
\footnotesize{\textsuperscript{6} Google, Switzerland}\\
}

\begin{document}


\twocolumn[{%
\maketitle
\renewcommand\twocolumn[1][]{#1}
\begin{center}
\vspace{-0.5cm}
    \centering
    \captionsetup{type=figure}
    \includegraphics[width=2.0\columnwidth]{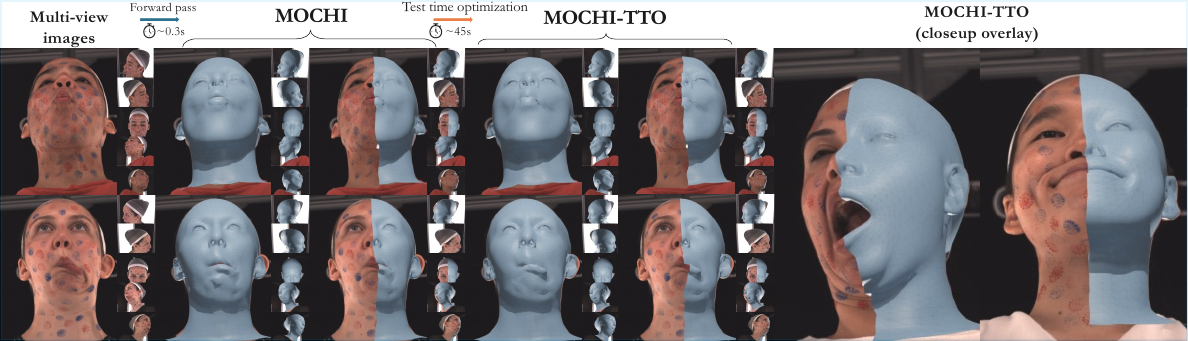}
    \caption{\textbf{\modelName}: Registration-free multi-view face capture in dense correspondence. Given calibrated multi-view images MOCHI predicts a canonical FLAME-topology mesh \textit{without needing precomputed registrations during training}. 
    When a target scan is available at test-time, an optional brief TTO of the model sharpens details and \emph{outperforms classical, manually labor-heavy registration pipelines}.
    }
    \label{fig:teaser}
\end{center}
}]

{
\renewcommand{\thefootnote}{\fnsymbol{footnote}}
\footnotetext[1]{Contributed foundational intellectual property prior to joining Google.}
\footnotetext[2]{The research work of P. P. Filntisis and P. Maragos was supported by the project ``Applied Research for Autonomous Robotic Systems'' (MIS 5200632) which is implemented within the framework of the National Recovery and Resilience Plan ``Greece 2.0'' (Measure: 16618- Basic and Applied Research) and is funded by the European Union- NextGenerationEU. 
}
}
\input{sec_camera_ready/0_abstract}

\input{sec_camera_ready/1_intro}

\input{sec_camera_ready/related}
\input{sec_camera_ready/4_method}
{
    \small
    \bibliographystyle{ieeenat_fullname}
    \bibliography{main}
}

\newpage

\section*{\textbf{Supplementary Material}}

\maketitle
\appendix
\input{sec_camera_ready/suppmat}    


\end{document}

%% file: preamble.tex
\usepackage{graphicx} 

\newcommand{\modelName}{MOCHI\xspace}

\usepackage{graphicx}
\usepackage{multirow}

\newcommand{\qheading}[1]{\noindent\textbf{#1}}

%% file: sec_camera_ready/0_abstract.tex
\begin{abstract}
Recent frameworks like ToFu and TEMPEH provide an automated alternative to classical registration pipelines by predicting 3D meshes in dense semantic correspondence directly from calibrated multi-view images.
However, these learning-based methods rely on the slow, manual registration pipelines they aim to replace for their training supervision.
We overcome this limitation with \modelName (\textbf{M}ulti-view \textbf{O}ptimizable \textbf{C}orrespondence of \textbf{H}eads from \textbf{I}mages), a multi-view 3D face prediction framework trained without requiring registered training data.
MOCHI eliminates the registration data dependency by enforcing topological consistency through a pseudo-linear inverse kinematic solver.
Semantic alignment is guided by dense keypoints from a 2D landmark predictor trained exclusively on synthetic data.
Our analysis further reveals that standard point-to-surface distances induce training instabilities and visual artifacts in registration-free settings.
We propose pointmap- and normal-based losses instead, which provide smoother gradients and superior reconstruction fidelity.
Finally, we introduce a test-time optimization scheme that refines network weights over a few dozen iterations.
This approach bridges the gap between feed-forward efficiency and iterative optimization precision, allowing MOCHI to outperform traditional labor-intensive pipelines in both reconstruction accuracy and visual quality. Code and model are public at: \url{https://filby89.github.io/mochi}.

\end{abstract}

%% file: sec_camera_ready/1_intro.tex
\section{Introduction}


3D face reconstruction in dense correspondence -- that is, producing meshes with a consistent topology across identities and expressions at the highest possible quality -- is a central problem in computer vision. 
It has direct impact on a number of industries such as entertainment (VFX in movies, games), future forms of communication (teleconferencing in AR, VR, Metaverse), healthcare (such as cosmetic surgery pre-visualization), and perhaps future 3D embodied AI assistants. 

Ideally, a face reconstruction method should not only recover accurate geometry but also enforce a shared topology across individuals and frames, enabling semantic consistency and compatibility with downstream tasks such as animation and editing.
Despite tremendous progress in face modeling and reconstruction~\cite{3DMM_survey, tewari2020neuralrendering}, the highest-fidelity systems still rely on multi-view stereo (MVS) capture~\cite{Furukawa2010_MVS, Schonberger2016_SfM} followed by non-rigid registration to bring scans into a predefined topology~\cite{Li2017_FLAME,Booth2018_3DMMFitting, Paysan2009_BFM}. This registration step requires iterative optimization and often manual verification or correction by human experts, making it one of the major bottlenecks for scalable 3D face capture.

Recent learning-based systems aim to simplify this process by predicting meshes in a fixed topology directly from calibrated multi-view images \cite{Bolkart2023_TEMPEH,Li2024_GRAPE,Li2021_ToFu} (see Fig.~\ref{fig:related_work}).
While such methods greatly reduce processing time, they still \emph{require precomputed registrations during training}, and preparing such data remains time-consuming and labor-intensive, and a classical registration pipeline must be in place to provide the training supervision.

To remove the dependency on precomputed registrations for training, we present \modelName, a \emph{registration-free} approach for multi-view 3D face reconstruction in dense correspondence which allows training directly on raw scans.
First, a pseudo-linear inverse kinematic solver~\cite{Shetty2023_PLIKS} recovers 3DMM parameters from the network's vertex outputs. These parameters are then passed through the parametric model to produce a topology-consistent mesh, which is used to regularize the network's free-form predictions via vertex- and edge-level losses.
Second, we observe that the traditionally used loss to train with scans, i.e., the point-to-surface distance, produces noisy gradients due to discrete closest point assignments thus often leading to self-intersections and other artifacts if the mesh is not already tightly aligned to the scan. \
We instead swap this out for differentiable rendering losses: \emph{pointmap} and \emph{surface-normal} losses~\cite{Wang2024_DUSt3R}. These losses provide smoother gradients and train the model more reliably.
Finally, because direct geometric errors often fail to capture semantically salient perceptual effects, we additionally supervise with 2D landmarks predicted by a model trained purely on synthetic data.
At inference, we also introduce a \emph{test-time optimization} (TTO) stage that performs a brief per-instance optimization of the model parameters, leading to registrations \textit{more accurate} than the classical manually intensive registration pipeline.
As a result, \modelName can 
register a new multi-view dataset from scratch, using only the raw scans and calibrated images, effectively minimizing the manual effort 
needed 
(Fig.~\ref{fig:related_work}). 

Our contributions can be summarized as follows:
(1) To our knowledge, the first dense-correspondence multi-view face reconstruction system that does not require precomputed registrations for training, yet surpasses registration-requiring methods such as TEMPEH~\cite{Bolkart2023_TEMPEH} in reconstruction accuracy.
(2) A stable training objective based on pointmap and surface-normal losses, reducing artifacts and improving model training compared to scan-to-mesh supervision.
(3) A lightweight test-time optimization (TTO) of our base model, which when coupled with an existing scan outperforms even classical manually-intensive registration pipelines.

%% file: sec_camera_ready/related.tex
\begin{figure}
    \centerline{
    \includegraphics[width=1.0\columnwidth]{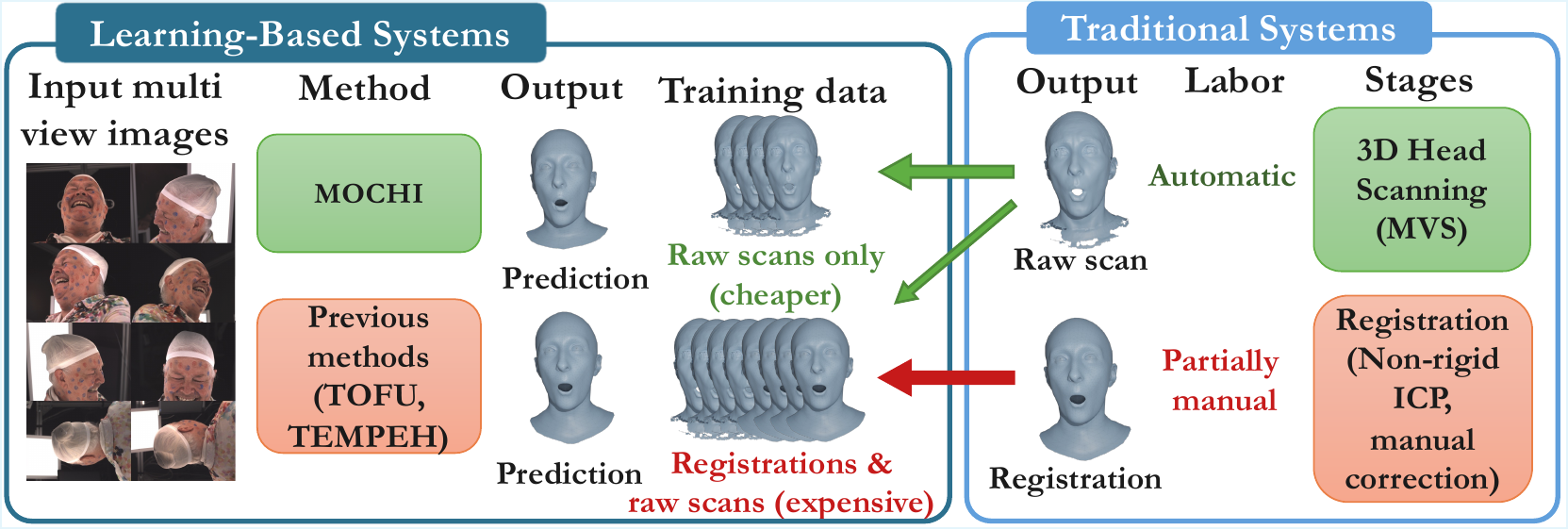}
    }
    \vspace{-0.2cm}
\caption{
    \emph{Traditional capturing systems} (right) rely on a multi-stage approach involving MVS to produce a head scan, followed by registration that requires manual supervision.
    Recent \emph{learning-based systems} (left) simplify the workflow but still require GT registrations from traditional systems for training (red arrow).
    \emph{\modelName} relies only on the fully automatic parts (green arrow), outperforming previous learning-based methods despite removing the need for expensive manual registrations. When coupled with TTO, \modelName surpasses even the classic manual registration pipeline.
}

    \label{fig:related_work}
\end{figure}

\section{Related Work}

Recovering 3D faces from images has advanced dramatically since the seminal work of Blanz and Vetter~\cite{Blanz1999_3DMM}. We review the most relevant lines of work below.

\qheading{Scan Registration.}
Traditional high-fidelity face capture begins with obtaining a 3D scan, most commonly via multi-view stereo (MVS)~\cite{HernandezSchmitt2004,Furukawa2010_MVS,Goesele2006stereo,KolmogorovZabih2002,VogiatzisTorrCipolla2005}, though laser scanning~\cite{levoy2000laser}, photometric stereo~\cite{Ghosh2011_MVSPerformance, ma2007rapid}, and learning-based methods~\cite{Kar2017LSM,Im2019DPSNet,Yao2018MVSNet,Wang2024_DUSt3R, Wang2024_VGGT} are also used; a comprehensive survey~\cite{wang2021multiview_stereo} covers these in detail.
The resulting scans have unstructured topology and may contain holes or artifacts in difficult regions (e.g.\ lips, eyes, facial hair), and must therefore be registered onto a common topology to be useful in production.
A common approach is to first fit a 3DMM to the scan~\cite{Amberg2008ExpressionI3, Blanz2002, li2020learningFormation}, providing a coarse initialization that is then refined with non-rigid ICP~\cite{besl1992icp} to capture finer detail~\cite{Li2017_FLAME, wu2018deep, Cao2014_FaceWarehouse, yang2020facescape, Paysan2009_BFM, Gerig2018, booth2016_3dmmm, booth2018large, li09robust, passalis2011using, Salazar2013FullyAE}.
Other methods skip the model initialization and apply non-rigid ICP directly to deform a template to the scan~\cite{beeler2010singleshot, bradley2010highres, beeler2011hq}.
Some works collapse scanning and registration into a single joint stage~\cite{FyffeNaganoHuynhSaitoBuschJonesLiDebevec2017, Riviere2020singleshot}.
Learning-based approaches train neural networks to map unstructured raw scans directly to a shared template topology, through mechanisms such as surface-to-surface translation~\cite{bahri2021shapemyface}, predicting point-wise displacements~\cite{Liu20193DFM}, or learning continuous implicit deformation fields~\cite{Zheng2022ImFace, zheng2025imfacePlus}. These are faster than iterative optimization but remain prone to failures on noisy inputs.
Overall, the above registration pipelines yield excellent fidelity but require heavy computation and substantial manual cleanup (e.g.\ resolving teeth artifacts, fixing failures on extreme expressions, and tuning hyperparameters per subject). In contrast, \modelName removes the need for any registration pipeline by learning directly from raw scans.

\qheading{Learnable Multi-View Mesh Prediction.}
A recent line of deep-learning-based methods~\cite{Li2021_ToFu, liu2022refa, Bolkart2023_TEMPEH, Li2024_GRAPE} aims to collapse the multi-stage, partly manual traditional pipeline into a single neural network that can be run without manual correction (see Fig.~\ref{fig:related_work}).
ToFu~\cite{Li2021_ToFu} pioneered this approach, using a volumetric deep feature map to predict vertices of a predefined template from multi-view calibrated images.
ReFA~\cite{liu2022refa} instead uses deep features to refine 3D position and normal maps in UV space.
TEMPEH~\cite{Bolkart2023_TEMPEH} builds upon ToFu by considering surface properties and visibility for multi-view feature fusion, and is supervised mostly with unstructured scans, but still needs at least coarse registrations for regularization.
GRAPE~\cite{Li2024_GRAPE} builds upon TEMPEH, proposing a visual hull and visibility-aware 3D feature aggregation to generalize to novel camera rigs.
Xiao et al.~\cite{xiao2022detailed} learn an implicit function to recover detailed facial geometry from multi-view images, supervised with GT displacement maps derived from registered templates.
All of the above methods require GT registrations and hence still need the traditional pipeline to create them (see Fig.~\ref{fig:related_work}).
With \modelName, we go one step further: we employ a pseudo-linear inverse kinematic solver~\cite{Shetty2023_PLIKS} to estimate FLAME~\cite{Li2017_FLAME} parameters, outputting a mesh that is used as a regularizer. This step, combined with our differentiable rendering losses and semantic guidance from dense 2D landmarks, \textit{effectively removes the need for GT registrations}. Note that while dense landmarks have been used previously in 3D face reconstruction~\cite{wood20223d, taubner2024flowface}, in our case they do not drive the geometric accuracy but act as a semantic regularizer for ambiguous regions such as eyes and lips.
%
Finally, some methods sidestep explicit meshes altogether, using neural representations learned directly from images~\cite{Lombardi:2019, Lombardi21, Ma2021PixelCA, Gao2022nerfblendshape, xu2023gaussianheadavatar, giebenhain2024npga, aneja2025scaffoldAvatar} to represent geometry via neural volumes, radiance fields, or Gaussian primitives. While these can produce high-fidelity renderings, they lack the editability and compatibility with standard graphics engines that explicit mesh-based representations, including ours, provide.

\qheading{Image-based Face Reconstruction.}
A large body of work focuses on simplifying the capture setup from calibrated multi-view to uncalibrated single-view images; recent surveys~\cite{3DMM_survey, tewari2020neuralrendering} give a comprehensive overview.
Early works focused on optimization-based fitting of 3DMMs to images using ``analysis-by-synthesis''~\cite{Blanz1999_3DMM, Vetter1998EstimatingC3, Cao2013_RealtimeFaceTracking, thies2016face, wu2016anyma}.
Many deep-learning works repurposed this paradigm for self-supervised systems~\cite{Tewari2017, Tewari2018, Ploumpis2020, Sanyal2019RingNet, Feng2021_DECA, Guo2020towards_3DDFA_V2, Shang2020_MGCNET, Danecek2022EMOCA, filntisis2022spectre, Zhang2023TokenFace, Retsinas2024_SMIRK}, where a neural network maps image features onto 3DMM and scene parameters.
A few recent works predict landmarks~\cite{taubner2024flowface} or depth and UV maps~\cite{giebenhain2023nphm} with post-hoc 3DMM fitting.
Some methods bypass 3DMMs entirely and train on 3D GT to output meshes~\cite{Deng2020_RetinaFace,Dou2017,Feng2018,Guler2017,Jung2021,Ruang2021_SADRNet,Sela2017,Szabo2019,Wei2019,Zeng2019_DF2Net,Wu2020}, voxels~\cite{Jackson2017}, or SDFs~\cite{Park2019_DeepSDF}, but these unstructured outputs require post-hoc registration.
Despite the progress, none of these methods reach the quality of studio-based capture.

%% file: sec_camera_ready/4_method.tex
\section{Method}
\input{figures/method_figure}

\subsection{Preliminaries}

\qheading{FLAME.} We adopt FLAME~\cite{Li2017_FLAME} as our parametric head model for representing facial geometry. It generates a mesh $\mathbf{V} \in \mathbb{R}^{n_v \times 3}$ with $n_v = 5023$ vertices as a function of shape $\boldsymbol{\beta}$, expression $\boldsymbol{\psi}$, and pose parameters $\boldsymbol{\theta}$. The mesh is computed by applying linear blend skinning (LBS) to a base template deformed by learned corrective bases:
$$
\mathbf{V} = \text{LBS}\left(\bar{\mathbf{V}} + \mathbf{B}_{\text{id}} \boldsymbol{\beta} + \mathbf{B}_{\text{exp}} \boldsymbol{\psi} + \mathbf{B}_{\text{pose}}(\boldsymbol{\theta}), \boldsymbol{\theta}, \mathcal{J}, \mathcal{W} \right),
$$
where $\bar{\mathbf{V}}$ is the mean template, $\mathbf{B}_{\text{id}}$ and $\mathbf{B}_{\text{exp}}$ are identity and expression bases, and $\mathbf{B}_{\text{pose}}$ are pose-dependent corrective blendshapes. The LBS operator uses joint locations $\mathcal{J}$ and skinning weights $\mathcal{W}$ to apply articulated pose deformations. 
\subsection{Methodology}



We follow the multi-view capture setting of TEMPEH~\cite{Bolkart2023_TEMPEH} but train \emph{without} registrations.
Given $K$ calibrated views $\{\mathcal{I}_i\}_{i=1}^{K}$ with corresponding cameras $\{C_i\}_{i=1}^{K}$, 
\textbf{\modelName} predicts a canonical FLAME-topology mesh 
$M = (\mathbf{V}, \mathbf{T})$, 
where $\mathbf{V} \in \mathbb{R}^{n_v\times3}$ are the $n_v$ predicted vertex coordinates and $\mathbf{T}$ denotes the fixed set of triangular faces defining mesh connectivity. 
The mesh $\hat{M}$ should be spatially aligned to the captured scan geometry while maintaining the canonical FLAME topology.

To enable training directly from raw scans, we introduce three key extensions: 
(i) a pseudo-linear inverse-kinematics branch that recovers FLAME parameters from per-vertex predictions to \emph{enforce} topology; 
(ii)
dense 2D landmarks from a synthetic-data-trained detector that act as a semantic regularizer, constraining articulated regions; and
(iii) fully differentiable point-map and surface-normal losses that replace scan-to-mesh distances, providing smoother gradients and improved reconstruction accuracy.

At inference, we further introduce an optional \emph{test-time optimization} (TTO) stage that refines the predicted mesh $\hat{M}$.
This refinement minimizes the same differentiable geometric losses used during training, allowing the network predictions to be locally adapted to the specific input scans without requiring explicit correspondences.

\subsection{Dense Landmark Detector}
\label{sec:tracker}

Previous methods like TEMPEH~\cite{Bolkart2023_TEMPEH} 
rely on manual registrations to supervise mesh prediction, since point-to-surface distances often fail in regions with partial or noisy geometry (e.g., around lips, teeth, and eyelids), unless the mesh is already tightly aligned to the scan. To overcome this, we introduce a dense landmark detector to predict per-vertex 2D correspondences aligned with the FLAME topology, offering semantic supervision where geometric cues are weak.

We train the detector on $25{,}000$ synthetic images rendered in Blender, using randomized shape, expression, pose, albedo, HDR lighting, and natural-image backgrounds. Hair is added via randomly sampled HAAR~\cite{Sklyarova2024haar} styles for added realism. The detector builds on a DINOv3-Large backbone with a LoRA branch and is trained via $\ell_2$ regression to predict dense 2D landmarks $\mathbf{U} \in \mathbb{R}^{n_v \times 2}$ from an input image $\mathbf{I}$. Generalization is validated on held-out synthetic data and real FaMoS scans; see Sup.~Mat.

\subsection{Coarse Stage}
\label{sec:coarse}

The coarse stage follows the volumetric fusion design of \textbf{TEMPEH}~\cite{Bolkart2023_TEMPEH}, but integrates our registration-free objectives. It predicts an initial canonical mesh from calibrated multi-view images by aggregating features into a localized 3D volume and reading out vertex positions probabilistically. We denote the full model as:
\begin{equation}
f_{\text{coarse}}: \{\mathbf{I}_i\}_{i=1}^K \;\longmapsto\; \mathbf{V}_c,
\end{equation}
where $\mathbf{V}_c$ are the vertices of the predicted mesh $M_c$ in canonical FLAME topology. The model comprises the following steps:

\noindent \textit{Feature Extraction and Fusion.}  
A shared encoder–decoder extracts per-view features $\mathbf{F}_i$. These are unprojected into a 3D grid $\mathbf{G}_c$ and fused via per-channel mean and variance, forming a coarse feature volume $\mathbf{Q}_c$.

\noindent \textit{Localization.}  
A spatial transformer estimates anisotropic scale, rotation, and translation to normalize the head region, producing a localized feature volume $\mathbf{Q}_c'$.

\noindent \textit{Vertex Readout.}  
A 3D U-Net predicts a probability volume $\mathbf{Q}_p$ with one channel per vertex. Vertex positions are obtained by spatial soft-argmax over $\mathbf{Q}_p$, yielding the vertices $\textbf{V}_c$ of the coarse mesh $M_c$.

\qheading{Rendered Geometric Losses.}
\label{sec:rendering-loss}
Conventional scan-to-mesh losses rely on nearest-neighbor search (e.g., point-to-surface), which introduce \emph{non-smooth gradients} due to discrete closest-point assignments. This often destabilizes training and may cause mesh artifacts or self-intersections, especially under large deformations.
Instead, we supervise geometry using \emph{rendered geometric maps} from all calibrated views, which provide smooth and spatially consistent gradients.
Given the ground-truth scan $\mathbf{S}$, we render for each camera view $i$:
(1) the surface normals $\mathbf{N}_{\text{gt}, i}$ and  
(2) the per-pixel 3D point maps $\mathbf{P}_{\text{gt}, i}$.  
From a predicted mesh $(\mathbf{V}_{\text{pred}}, \mathbf{T})$, 
we differentiably render the corresponding maps $\mathbf{N}_{\text{pred}, i}$ and $\mathbf{P}_{\text{pred}, i}$.  
We then minimize a robust geometric discrepancy along all views, using the Geman–McClure penalty \cite{geman1987statistical} to mitigate scan noise and outliers:
\begin{equation}
\label{eq:geom}
\mathcal{L}_{\text{geom}} = \sum_{i=1}^K
\rho_{\text{GM}}\!\big(\|\mathbf{N}_{\text{pred,i}}-\mathbf{N}_{\text{gt,i}}\|_2\big)
+ \rho_{\text{GM}}\!\big(\|\mathbf{P}_{\text{pred,i}}-\mathbf{P}_{\text{gt,i}}\|_2\big),
\end{equation}
where  $\rho_{\text{GM}}(x)=\frac{x^2}{x^2+\sigma^2}$ ($\sigma$ is set to 10).

\noindent\textbf{Recovery of 3DMM Parameters.} To anchor predicted meshes to a canonical topology, we apply a differentiable inverse module based on the PLIKS layer~\cite{Shetty2023_PLIKS}.  
Given a predicted mesh $\mathbf{V}_{\text{pred}} \in \mathbb{R}^{n_v \times 3}$, the solver first estimates a set of rigid transformations $\{\mathbf{R}_s\}_{s=1}^S$ for FLAME’s $S$ skinning segments via Procrustes alignment between template vertices and their predicted locations (using dominant segment assignments).  
For efficiency, we perform this estimation only once, without iterative refinement.
With rotations fixed, PLIKS solves a least-squares problem to estimate shape $\boldsymbol{\beta}$, expression $\boldsymbol{\psi}$, and translation $\mathbf{t}$:
$$
[\boldsymbol{\beta},\boldsymbol{\psi},\mathbf{t}]
= \arg\min \;\big\|
\mathbf{R} \big(\bar{\mathbf{V}} 
+ \mathbf{B}_{\text{id}}\boldsymbol{\beta} 
+ \mathbf{B}_{\text{exp}}\boldsymbol{\psi} \big) 
+ \mathbf{t} - \mathbf{V}_{\text{pred}} \big\|_2^2.
$$
Here, $\mathbf{R}$ denotes the piecewise rigid motion applied per-vertex, using the rotation of the corresponding dominant segment. This compact notation is a simplification for clarity; the underlying system accounts for per-vertex transformations and remains fully linear in the unknowns.  
We then regularize the recovered parameters to encourage in-distribution identity and expression:
$$
\mathcal{L}_{\text{PLIKS-reg}} =
\lambda_\beta \|\boldsymbol{\beta}\|_2^2 +
\lambda_\psi \|\boldsymbol{\psi}\|_2^2.
$$


\qheading{Re-forward and Topology Enforcement.}
The recovered parameters are re-forwarded through a FLAME layer~\cite{Li2017_FLAME} to obtain a topology-respecting set of vertices $\mathbf{V}_{\text{fl}}$, correcting distortions from the unconstrained regression.
We then enforce vertex-level consistency and edge regularization:
\begin{equation}
\label{eq:align}
\mathcal{L}_{\text{PLIKS-align}} =
\lambda_v \big\|\mathbf{V}_{\text{fl}}-\mathbf{V}_{\text{pred}}\big\|_2^2
\;+\;
\lambda_e \,\mathcal{L}_{\text{edge}}\!\big(\mathbf{V}_{\text{fl}},\,\mathbf{V}_{\text{pred}}\big),
\end{equation}
where $\mathcal{L}_{\text{edge}}$ penalizes relative changes of edge lengths, given the mesh triangles $\textbf{T}$. 


With these terms, PLIKS serves as an \emph{implicit topology regularizer}: it softly aligns unconstrained vertex predictions to valid FLAME topology, while allowing gradients to flow through both representations. This bidirectional consistency encourages $\mathbf{V}_{\text{pred}}$ to follow plausible anatomical structure, and drives $\mathbf{V}_{\text{fl}}$ to reflect view-dependent details from free-form predictions. 

\noindent\textit{Note:} One might ask why we do not regress FLAME parameters directly. Learning identity parameters robustly would require many more subjects than typical multi-view datasets provide, leading to overfitting. Moreover, predicting FLAME parameters would constrain reconstructions strictly to its shape space. Our implicit regularization instead nudges $\mathbf{V}_{\text{pred}}$ toward FLAME’s canonical topology while still allowing deviations from the 3DMM manifold.


\qheading{Landmark Loss.}
%
Topology-aware and geometric losses alone cannot resolve ambiguities caused by partial, noisy, or semantically imprecise scans.  
To address this, we supervise the model using dense 2D landmarks from the detector (see Section~\ref{sec:tracker}), which provide a semantically rich signal over key facial regions.

Given tracked 2D landmarks $\mathbf{U}_i \in \mathbb{R}^{n_v \times 2}$ for view $i$ and projection function $\Pi_i$, we define a reprojection loss for a set of 3D vertices $\mathbf{V} \in \mathbb{R}^{n_v \times 3}$ as
$
\mathcal{D}_i(\mathbf{V}) = \left\| \left( \Pi_i(\mathbf{V}) - \mathbf{U}_i \right) \right\|_2^2
$.
We apply this loss to both the canonical mesh $\mathbf{V}_{\text{fl}}$ and the free-form output $\mathbf{V}_{\text{pred}}$, encouraging semantic consistency in both regressed parameters and raw vertex predictions: 
$$
\mathcal{L}_{\text{lm}} = \sum_{i=1}^{K} \left[ \mathcal{D}_i(\mathbf{V}_{\text{fl}}) + \mathcal{D}_i(\mathbf{V}_{\text{pred}}) \right].
$$


\qheading{Total Training Objective.}
The overall training loss combines geometric, landmark, and topology consistency terms:
\begin{align}
\mathcal{L}_{\text{total}} \;=\;
& \;\lambda_{\text{geom}} \mathcal{L}_{\text{geom}}
\;+\; \lambda_{\text{lm}}\,\mathcal{L}_{\text{lm}} \nonumber\\
& \;+\; \lambda_{\text{align}}\,\mathcal{L}_{\text{PLIKS-align}}
\;+\; \lambda_{\text{reg}}\,\mathcal{L}_{\text{PLIKS-reg}},
\label{eq:total_loss}
\end{align}
where each $\lambda$ term controls 
the strength of its term.

\newcommand{\best}[1]{\textbf{\textcolor{cvprblue}{#1}}}

\begin{table*}[t]
\centering
\setlength{\tabcolsep}{6pt}
\footnotesize
\resizebox{\textwidth}{!}{%
\begin{tabular}{ll lccc ccc ccc ccc}
\toprule
& & & \multicolumn{3}{c}{\textbf{Total}} &
  \multicolumn{3}{c}{\textbf{Face}} &
  \multicolumn{3}{c}{\textbf{Lips}} &
  \multicolumn{3}{c}{\textbf{Neck}} \\
\cmidrule(lr){4-6}\cmidrule(lr){7-9}\cmidrule(lr){10-12}\cmidrule(lr){13-15}

\textbf{Dataset} & & \textbf{Method} 
& Median $\downarrow$ & Mean $\downarrow$ & Std $\downarrow$
& Median $\downarrow$ & Mean $\downarrow$ & Std $\downarrow$
& Median $\downarrow$ & Mean $\downarrow$ & Std $\downarrow$
& Median $\downarrow$ & Mean $\downarrow$ & Std $\downarrow$ \\

\midrule
\multirow{4}{*}{FaMoS}
& \multirow{2}{*}{Images Only}
& \textit{TEMPEH} 
& 0.36 & 0.63 & 1.41 
& 0.32 & 0.49 & 1.02 
& 0.47 & 0.74 & \best{1.10} 
& 0.63 & 1.41 & \best{2.24} \\

& & \textit{MOCHI} 
& \best{0.26} & \best{0.48} & \best{1.38} 
& \best{0.23} & \best{0.36} & \best{0.98} 
& \best{0.37} & \best{0.69} & 1.23 
& \best{0.46} & \best{1.22} & 2.36 \\

\cmidrule(lr){2-15}

& \multirow{2}{*}{Images + Scan}
& \textit{Classic Registrations} 
& 0.10 & 0.24 & 1.40 
& 0.09 & 0.15 & \best{0.89} 
& 0.09 & 0.28 & \best{1.12} 
& 0.23 & 0.98 & 2.79 \\

& & \textit{MOCHI TTO} 
& \best{0.07} & \best{0.21} & \best{1.31} 
& \best{0.06} & \best{0.13} & 0.92 
& \best{0.06} & \best{0.20} & 1.22 
& \best{0.14} & \best{0.53} & \best{1.78} \\

\midrule

\multirow{4}{*}{CoMA}
& \multirow{2}{*}{Images Only}
& \textit{TEMPEH} 
& 0.81 & 1.40 & \best{1.69} 
& 0.69 & 1.13 & 1.32 
& 1.34 & 1.90 & 1.98 
& 1.26 & \best{2.27} & \best{2.71} \\

& & \textit{MOCHI} 
& \best{0.53} & \best{1.09} & 1.70 
& \best{0.45} & \best{0.79} & \best{1.10} 
& \best{0.69} & \best{1.21} & \best{1.69} 
& \best{1.02} & 2.31 & 3.21 \\

\cmidrule(lr){2-15}

& \multirow{2}{*}{Images + Scan}
& \textit{Classic Registrations} 
& 0.10 & 0.23 & 0.70 
& 0.09 & 0.16 & 0.39 
& 0.11 & 0.40 & 1.39 
& 0.25 & 0.79 & 1.79 \\

& & \textit{MOCHI TTO} 
& \best{0.07} & \best{0.17} & \best{0.67} 
& \best{0.06} & \best{0.10} & \best{0.31} 
& \best{0.06} & \best{0.23} & \best{1.12} 
& \best{0.15} & \best{0.53} & \best{1.44} \\

\bottomrule
\end{tabular}%
}

\caption{\textbf{Quantitative evaluation on FaMoS and CoMA.} We report Median, Mean, and Std point-to-surface error (mm) per region and for the full head (excluding scalp). Lower is better. FaMoS results use the official test split; CoMA results are zero-shot. MOCHI outperforms TEMPEH on both datasets despite training without registrations. MOCHI TTO further surpasses the classic registration pipeline.}
\label{tab:main_results}
\end{table*}

\subsection{Refinement Stage}
\label{sec:local}

\qheading{Refinement Stage.}
%
To enhance geometric detail, the refinement stage adjusts each coarse vertex based on localized multi-view features and surface cues.
We denote the full refinement module as 
$$
f_{\text{refine}}: (\mathbf{V}_{\text{pred}}, \{\mathbf{F}_i\}_{i=1}^K) \mapsto \mathbf{V}_{\text{ref}},
$$
where $\mathbf{V}_{pred}$ is the predicted coarse mesh and $\mathbf{F}_i$ are the already extracted per-view image features.

Following TEMPEH~\cite{Bolkart2023_TEMPEH}, we perform these steps:\\
\noindent \textit{Surface-aware fusion.}  
For each vertex $\mathbf{v}_{p} \in \mathbf{V}_{pred}$, multi-view features are sampled over a local 3D grid aligned with its surface normal and aggregated using visibility-aware weights.

\noindent \textit{Probabilistic refinement.}  
A local 3D U-Net predicts a spatial distribution over the sampling grid, and the refined position $\mathbf{v}_r$ is then obtained via soft-argmax.

To train the refinement module, now that we have a coarse registration  $\mathbf{V}_{\text{ref}}$ we can use it as a geometric anchor. The total loss $\mathcal{L_{\text{ref}}}$ combines: (1) a multi-view landmark loss $\sum_{i=1}^{K}\mathcal{D}_i(\mathbf{V}_{\text{ref}})$; (2) edge-length regularization $\mathcal{L}_{\text{edge}}(\mathbf{V}_{\text{ref}},\mathbf{V}_{\text{pred}})$; (3) an eyeball constraint $\|\mathbf{M}^{\text{eyes}}\,\odot\,(\mathbf{V}_{\text{ref}}-\mathbf{V}_{\text{pred}})\|_2^2$, where $\mathbf{M}^{\text{eyes}}$ is an eyeball-vertex mask; and (4) a geometric loss as defined in Eq.~\ref{eq:geom}.



\subsection{Test-Time Optimization}
\label{sec:tto}



While \modelName\ generalizes well to unseen subjects and expressions, we can further tailor a reconstruction to a \emph{specific} capture at inference via a short \emph{test-time optimization} (TTO) procedure. Given a target scan, we fine-tune \emph{only} the refinement module for a fixed number of iterations (typically 50) using the same losses as in Sec.~\ref{sec:local}. This per-scan adaptation improves high-frequency detail and corrects residual misalignments -- especially around challenging regions like the lips and eyelids -- \emph{without} requiring any precomputed registrations. 
This test-time optimization benefits from the learned priors of the local refinement module, requiring only a few iterations to converge. Unlike vertex-level fitting from scratch, our TTO operates in a learned latent space, providing semantic regularization and robustness to scan noise or partial geometry. As a result, the refinement is stable, fast, and complements the coarse-to-fine design by tailoring predictions to each specific scan. 

\section{Experiments}
\label{sec:experiments}

\begin{figure*}[t]
    \centering
    \includegraphics[width=1.0\linewidth]{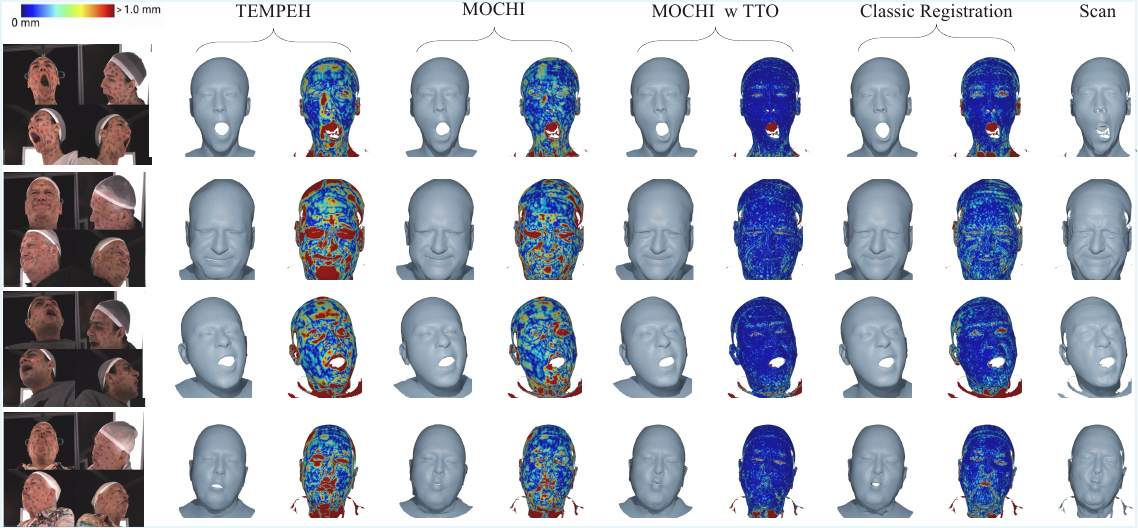}
    \caption{\textbf{Qualitative results on FaMoS (8 views).} From Left to Right: \emph{TEMPEH}~\cite{Bolkart2023_TEMPEH}, \emph{MOCHI}, \emph{MOCHI + TTO} (50 iters), \emph{Classic Registration}, and the \emph{Scan}. For each method we show the predicted mesh (left) and the point-to-surface error heatmap on the scan (right; red $\geq$ 1.0 mm, lower is better). MOCHI improves over TEMPEH on unseen subjects even without using registrations for training; adding TTO further reduces error and can surpass the classic manual registration pipeline.}
    \label{fig:qualitative}
\end{figure*}

\qheading{Implementation Details.} 
We train MOCHI on one NVIDIA A100 (80GB). We pretrain using only 2D landmarks for 150,000 iterations. Next, the coarse stage is trained for 300,000 iterations, and the refinement stage for 300,000 iterations, totaling $\approx$1 week.

\qheading{Datasets.}  
We train and evaluate \modelName on the \textbf{FaMoS} multi-view head capture dataset \cite{Bolkart2023_TEMPEH}. FaMoS is a dynamic multi-view head dataset with 8 color cameras, intrinsic and extrinsic parameters, and raw scans. It includes 94 subjects, each performing 28 sequences of facial expressions and head motions; all subjects wear hair nets. Following the official split \cite{Bolkart2023_TEMPEH}, we use 70 subjects for training (87,000 multi-view samples), 8 for validation (1,118 multi-view samples), and 16 for testing (8,800 multi view samples). To the best of our knowledge, FaMoS is the only publicly available multi-view dataset that provides both calibrated camera parameters and raw scans. We additionally evaluate on \textbf{CoMA}~\cite{coma2018ranjan}, which shares the same 3dMD rig type as FaMoS but is much smaller in scale (12 subjects, 12 expressions, 6 views vs.\ 8 in FaMoS). We apply TEMPEH, MOCHI, and MOCHI-TTO directly on CoMA \emph{without any fine-tuning}, serving as a zero-shot generalization test.

\qheading{Evaluation Metrics.}  \
We follow TEMPEH \cite{Bolkart2023_TEMPEH} and GRAPE \cite{Li2024_GRAPE} and quantify reconstruction accuracy on the FaMoS test set by computing the \emph{point-to-surface} distance from each vertex of the ground-truth scan to its closest point on the predicted 3D head surface.  
We report the \textbf{median}, \textbf{mean}, and \textbf{standard deviation} of this distance, separately for the \emph{head without scalp}, \emph{face}, \emph{lips}, and \emph{neck} regions.  
We omit the scalp region from evaluation, since the subjects in the dataset wear head caps to conceal hair, resulting in unreliable geometry in those areas (see Sup.~Mat.).

\subsection{Quantitative Evaluations}
\qheading{Generalization on New Images.}
We evaluate the ability of \modelName to generalize to new multi-view samples by comparing it against TEMPEH~\cite{Bolkart2023_TEMPEH}, which to the best of our knowledge is the only learning-based registration method with an open-source implementation. We present quantitative results on the FaMoS test set in Table~\ref{tab:main_results}. As can be seen, \modelName outperforms TEMPEH, achieving lower reconstruction errors (approximately a 28\% drop in the median error for the full head without scalp), despite having removed supervision with registrations.

\qheading{Dataset Registration.}
We further evaluate the ability of \modelName to register an entire dataset by comparing it to a traditional registration-based FLAME fitting pipeline that relies on optimization-based alignment with semi-manual supervision \cite{Bolkart2023_TEMPEH,Li2017_FLAME}. In Table~\ref{tab:main_results} we report results obtained by running test-time optimization (TTO) for 50 iterations on the FaMoS test set. MOCHI TTO outperforms the traditional, labor-intensive registration pipeline.

\qheading{Zero-Shot Generalization on CoMA.}
To assess generalization beyond FaMoS, we evaluate all methods zero-shot on CoMA~\cite{coma2018ranjan}, which uses the same 3dMD rig type but with 6 views instead of 8 and no fine-tuning. As shown in Table~\ref{tab:main_results}, MOCHI outperforms TEMPEH by a significant margin (0.53 vs.\ 0.81 median error for the full head), confirming that the gains transfer across datasets. MOCHI TTO again surpasses the classic registration pipeline.

\subsection{Qualitative Evaluation}
Figure~\ref{fig:qualitative} presents qualitative results of both MOCHI, and MOCHI w/ TTO (50 iterations) on FaMoS test subjects, comparing against TEMPEH \cite{Bolkart2023_TEMPEH} and classic registrations. As it can be seen in the color coded point-to-surface errors, in unseen subjects MOCHI achieves a more accurate reconstruction compared to TEMPEH, even though the registration is not used for training. Additionally, MOCHI with TTO significantly improve the reconstruction result - even outperforming the classic manual registration pipeline.

\subsection{Ablation Studies}

\begin{figure}
    \centering
    \includegraphics[width=0.85\linewidth]{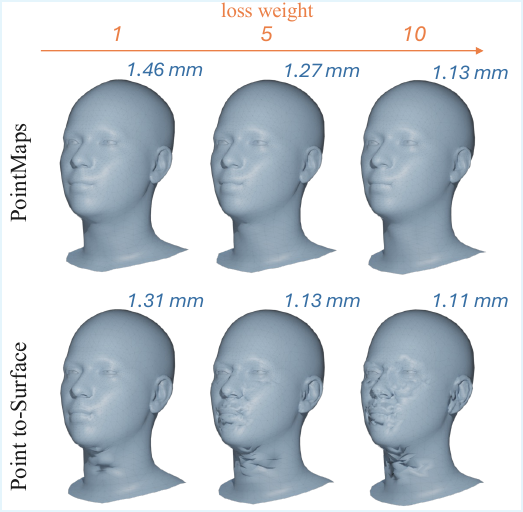}
    \caption{\textbf{Effect of loss type/weight.} Training the coarse model with point maps (top) and point-to-surface loss (bottom). When starting from a very coarse prediction using only landmark pretraining, increasing the point-to-surface loss weight leads to artifacts, whereas point-map supervision remains stable. Blue numbers show the point-to-surface error (mm).}
    \label{fig:s2m}
\end{figure}

\qheading{Point-to-surface vs.~Point maps.}
In Fig.~\ref{fig:s2m} we compare training the coarse stage with the conventional \emph{point-to-surface} loss against our differentiable \emph{point-map} losses. 
As the point-to-surface weight increases, reconstructions develop visible artifacts and self-intersections, bypassing the topology enforcement from our regularizers. On the contrary, supervision with point-maps shows a much smoother behavior, preserving clean geometry even in larger weights.

\begin{figure}
    \centering
    \includegraphics[width=1.0\linewidth]{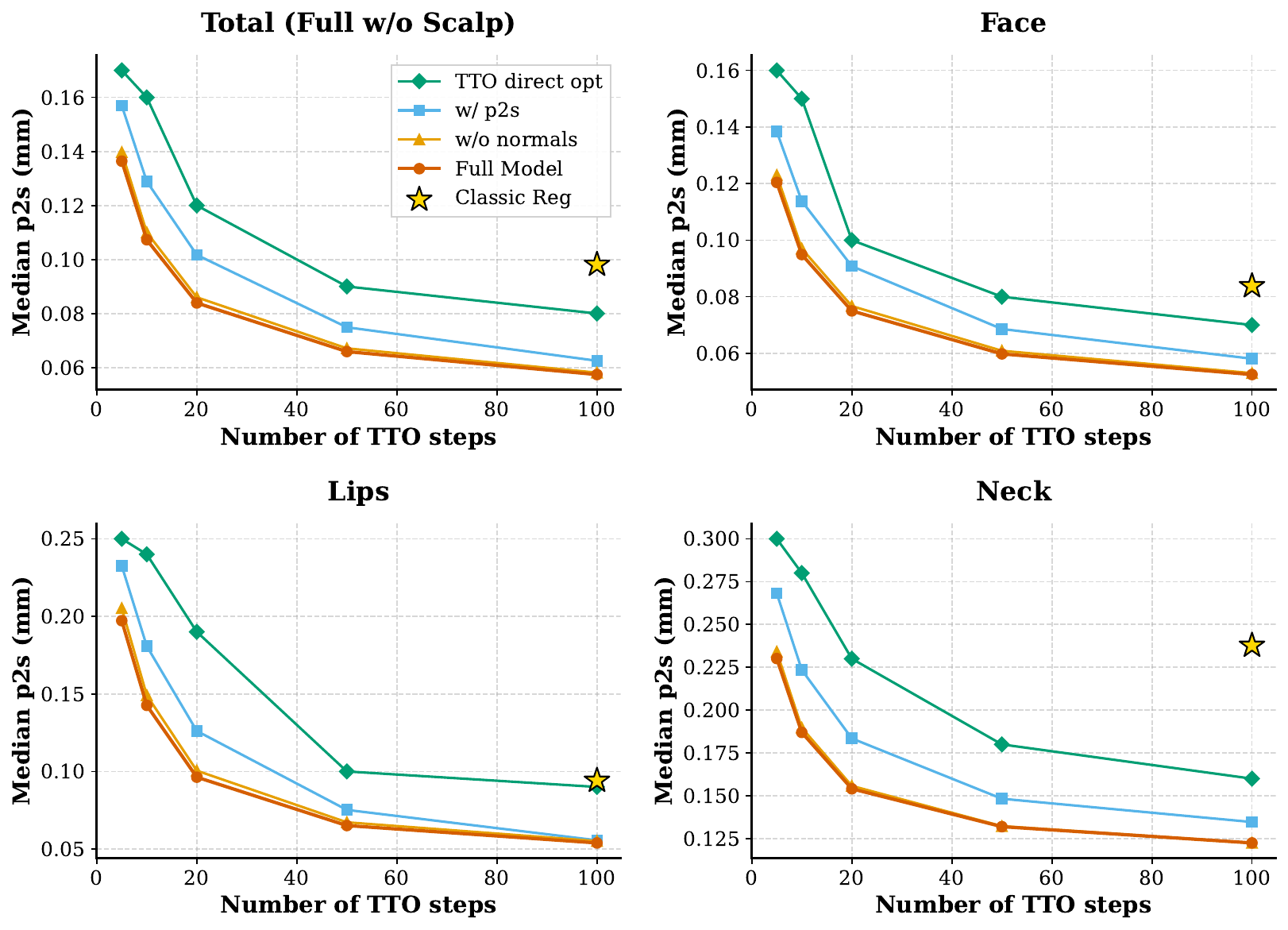}
    \caption{\textbf{TTO ablation on the FaMoS validation set.} We ablate (1) the number of TTO steps, (2) training without normals, and (3) training with the point-to-surface loss. Best results are obtained when using with both point-map and normal losses. The black star shows classic registration metrics. Full table in Sup.~Mat.}
    \label{fig:tto_ablations}
\end{figure}

In a similar fashion,
the TTO ablations in Fig.~\ref{fig:tto_ablations} show that point maps achieve \emph{lower} point-to-surface error than training directly with the point-to-surface objective. We believe that this is a direct result of the non-differentiable closest triangle selection in the point-to-surface loss, making the loss less suitable for optimization. 
\begin{figure}
    \centering
    \includegraphics[width=0.9\linewidth]{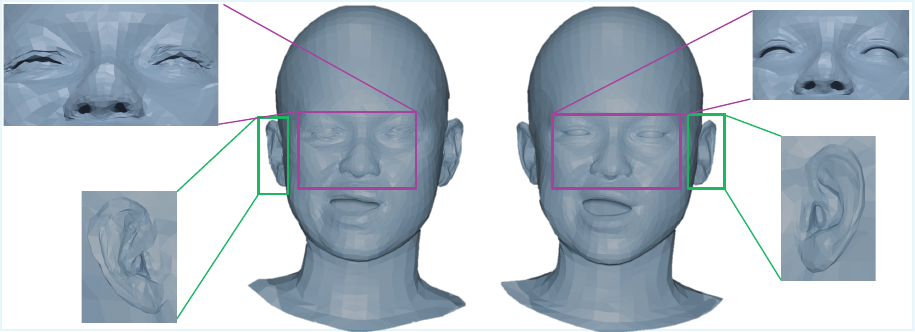}
    \caption{\textbf{Effect of inverse parameter recovery.} 
    \textit{Left:} only 2D landmark regularization. 
    \textit{Right:} 2D landmarks + PLIKS. 
    While 2D landmarks alone can approximate the facial topology, sensitive areas such as eyes, nose, and ears exhibit artifacts and noisy triangles. Adding PLIKS enforces anatomical consistency.}
    \label{fig:pliks}
\end{figure}
%


\qheading{Inverse Recovery of Parameters.}
Fig.~\ref{fig:pliks} shows the effect of using only dense 2D landmarks versus using landmarks \emph{plus} PLIKS-based inverse parameter recovery. Landmarks provide semantic guidance but do not constrain the 3D topology resulting in reconstructions with self-intersecting mesh triangles and severe noise in sensitive regions (eyelids, nose/alar base, ear helix). 
Adding PLIKS enforces a canonical topology, suppresses artifacts, and stabilizes edges, yielding anatomically consistent eyes, nose, and lips.
Note that a quantitative comparison is not meaningful since without PLIKS, landmarks alone effectively ``spray'' triangles onto the scan surface, yielding deceptively low errors despite severe anatomical failures. Also, omitting landmarks and supervising only with point-map losses plus PLIKS  produces plausible results, but causes systematic misalignments in expression-sensitive regions (e.g., lips, eyes).


\qheading{Normal Maps.} Fig.~\ref{fig:tto_ablations} also presents results when performing TTO with and without normal maps. As can be seen, normal maps provide a small additional boost to reconstruction accuracy, especially in the early steps.

\qheading{Direct Vertex Optimization.} Instead of fine-tuning the network parameters, vertex coordinates are optimized directly to minimize the reconstruction loss. Before fine-tuning, the mesh is initialized by a single-step MOCHI prediction. 
For fairness, we tuned the learning rate specifically for this baseline to ensure optimal convergence. As seen in Fig.~\ref{fig:tto_ablations},  direct vertex optimization also yields higher error, supporting our hypothesis that latent-space optimization provides implicit semantic regularization. Unlike direct vertex manipulation, it preserves a strong geometric prior and ensures robustness to scan noise.
Note that optimizing the refinement module rather than raw vertices effectively traverses the solution manifold of plausible facial geometries, enabling fast and stable convergence with only a few dozen steps. In the Sup. Mat. we also show visually that direct vertex opt is more prone to artifacts.

\qheading{Test-Time Optimization.}
We show results on our quantitative ablation study on test-time optimization in Fig~\ref{fig:tto_ablations}. As we see, with as little as 20 iterations of TTO we are able to surpass the reconstruction accuracy of a classic manual registration pipeline. Further increasing the iterations can help improve the reconstruction result even more. Interestingly, as we already said previously, starting from the same checkpoint and training with point-to-surface for TTO results in suboptimal reconstruction.  
More extensive ablation studies and the full table can also be found in the Suppl. Mat.

\qheading{Limitations.}
%
Our rendering losses outperform the point-to-surface loss but require extra computation to render point and normal maps for predictions (scan maps can be pre-rendered).
On FaMoS (8 views), this adds $\approx60$ ms per iteration, compared to using point-to-surface loss. 
Additionally, the 
PLIKS solver
adds $\approx150$ ms per iteration. 
All timings were calculated with Kaolin \cite{jatavallabhula2019kaolin} on an NVIDIA A100 (80GB).
Furthermore, despite our strong topology enforcement scheme, 
in rare cases TTO can fail. 
For instance, when scans include a large region of teeth,
the model may latch onto teeth geometry and distort the lips (see Sup.~Mat.).



%

\section{Conclusions}
We presented \modelName, a registration-free framework for multi-view 3D face reconstruction in dense correspondence. By moving the registration step inside the network, \modelName learns directly from raw multi-view scans. This is achieved by combining a pseudo-linear inverse kinematic solver to enforce a consistent FLAME topology, stable differentiable pointmap and surface-normals losses, and semantic guidance from a dense 2D landmark detector. As a result, \modelName can be directly utilized to register new multi-view datasets from scratch using only raw scans and camera calibrations, removing the need for pre-existing manual registrations. Furthermore, a lightweight test-time optimization (TTO) can be applied to further refine the output for any given scan, leading to a final registration that is more accurate than those produced by classical, manually-intensive pipelines.

\clearpage
\section*{Acknowledgments}
\begin{minipage}{0.7\linewidth}
The work of P. P. Filntisis and P. Maragos is also partially funded by the European Union under Horizon Europe (grant No. 101136568 -- HERON).
\end{minipage}%
\hfill
\begin{minipage}{0.20\linewidth}
    \includegraphics[width=\linewidth]{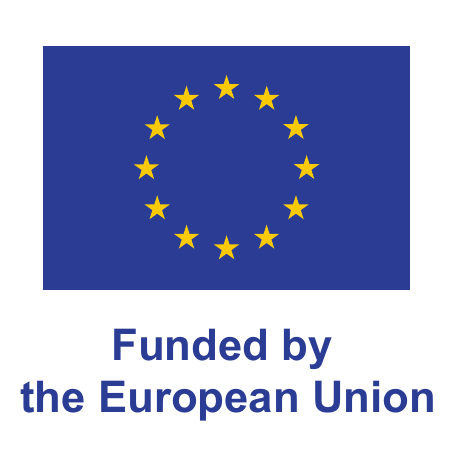}
\end{minipage}
\\
{\raggedright Vanessa Sklyarova is supported by the Max Planck ETH Center for Learning Systems. \par}

%% file: figures/method_figure.tex
\begin{figure*}
    \centerline{
    \includegraphics[width=2.0\columnwidth]{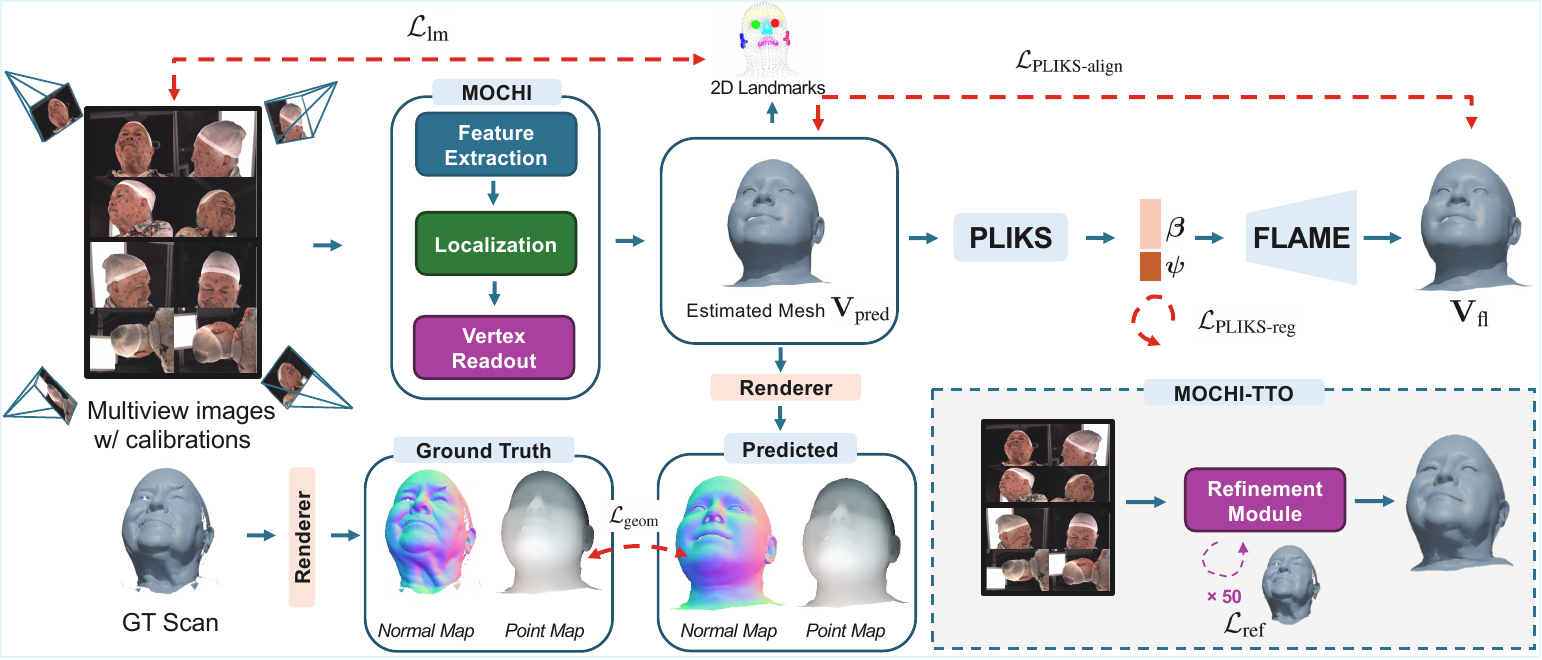}}
    \vspace{-0.2cm}
    \caption{
        Given calibrated multi-view images, \modelName predicts the vertex positions \(\mathbf{V}_{\text{pred}}\) of a FLAME-topology mesh \(\hat{M} = (\mathbf{V}_{\text{pred}}, \mathbf{T})\) that is spatially aligned to the raw scan. A pseudo-linear inverse kinematic solver is used to recover the FLAME identity and expression parameters that best explain \(\mathbf{V}_{\text{pred}}\), which are then passed through a differentiable FLAME layer to obtain a canonical mesh with vertices \(\mathbf{V}_{\text{fl}}\). We regularize the predicted mesh against \(\mathbf{V}_{\text{fl}}\) via edge- and vertex-position losses, and impose additional regularization on the recovered FLAME parameters. For both \(\mathbf{V}_{\text{fl}}\) and \(\mathbf{V}_{\text{pred}}\), we further apply a projected 2D landmark loss using the predictions of a tracker trained purely on synthetic data. Finally, we differentiably rasterize the mesh into point maps and normal maps and supervise them with the corresponding maps rendered from the ground-truth scan.
         At test time, an optional TTO stage (gray box, right) fine-tunes the refinement module for a small number of iterations using the same geometric losses, further improving reconstruction accuracy.
    }    
    \label{fig:method_scheme}
\end{figure*}




%% file: sec_camera_ready/suppmat.tex


This supplementary material provides additional details and results for MOCHI.
\vspace{-.1cm}

\section{Implementation Details}
\begin{figure}[h]
    \centering
    \includegraphics[width=0.98\linewidth]{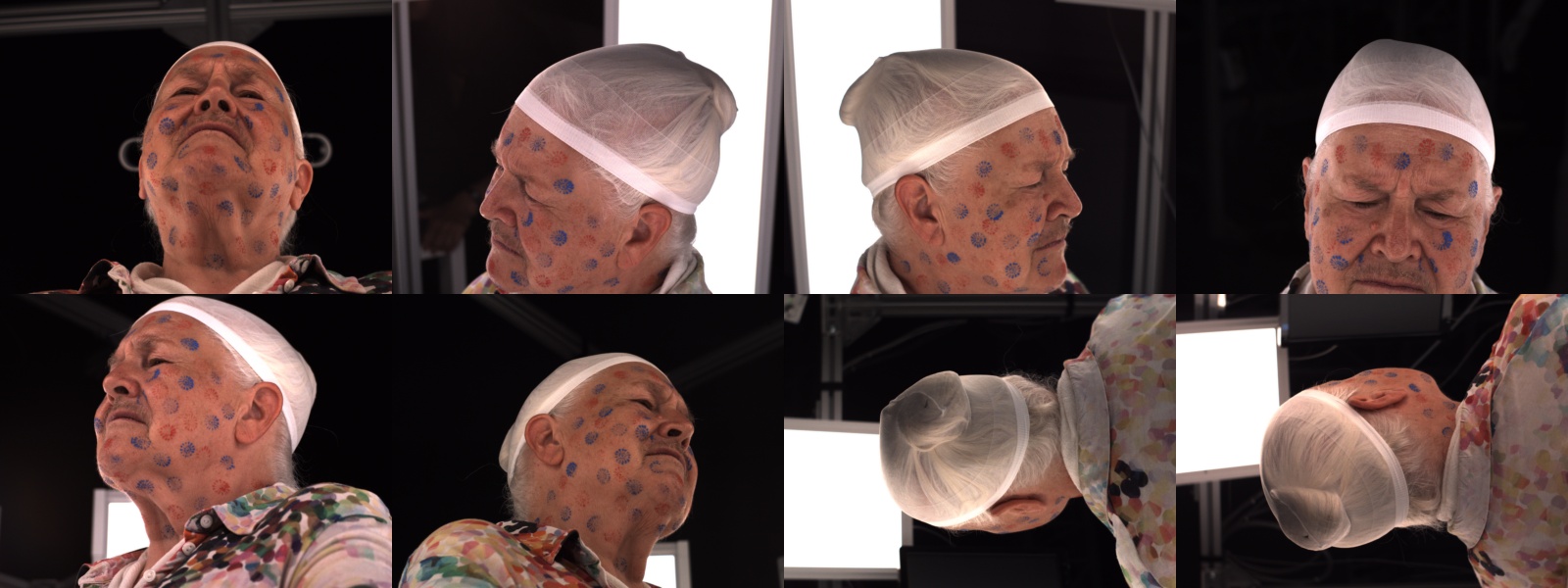}
    \caption{\textbf{Multi-view input setup.} Example views from the 8 RGB cameras used in the FaMoS dataset. We utilize only these RGB streams for both MOCHI and the TEMPEH baseline.}
    \label{fig:famos_multiview}
\end{figure}

\paragraph{FaMoS Setup}
The FaMoS dataset acquisition setup comprises both grayscale stereo pairs (8 pairs) and RGB cameras (8 cameras). For training MOCHI, we rely exclusively on the 8 RGB cameras; an example of the input views from this configuration is shown in Fig.~\ref{fig:famos_multiview}. To ensure a fair comparison, we also retrained the baseline method TEMPEH\cite{Bolkart2023_TEMPEH} using its official open-source implementation, configuring it to use the same set of 8 RGB cameras instead of the stereo pairs. For more information on FaMoS please see \cite{Bolkart2023_TEMPEH}.

\vspace{-.4cm}
\paragraph{Exclusion of Scalp Region from Evaluation}

\begin{figure}[h]
    \centering
    \includegraphics[width=0.98\linewidth]{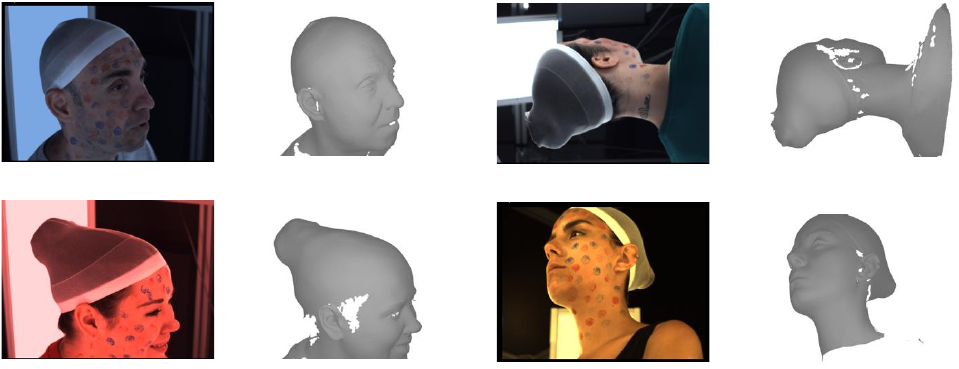}
    \caption{\textbf{Impact of hair caps on ground-truth scans.} We show example input images and their corresponding raw scans from the FaMoS dataset. Depending on the fit of the hair cap, the resulting scans exhibit significant distortion in the scalp region, often including the geometry of the cap itself. As a result, this region does not faithfully represent the subject's actual scalp and is excluded from our quantitative metrics.}
    \label{fig:scalp}
\end{figure}

In the quantitative evaluation presented in the main text, we explicitly exclude the scalp region from our evaluation protocol. This exclusion is necessary because subjects in the FaMoS dataset wore hair caps (or hair nets) to conceal their hair during capture. These caps introduce noise into the reconstructed raw scans—often capturing the geometry of the fabric rather than the subject's actual anatomical shape. Consequently, the raw scans in this region do not serve as a reliable ground truth for modeling the scalp. We illustrate examples of this in Fig.~\ref{fig:scalp}.

\begin{figure}[h]
    \centering
    \includegraphics[width=.6\linewidth]{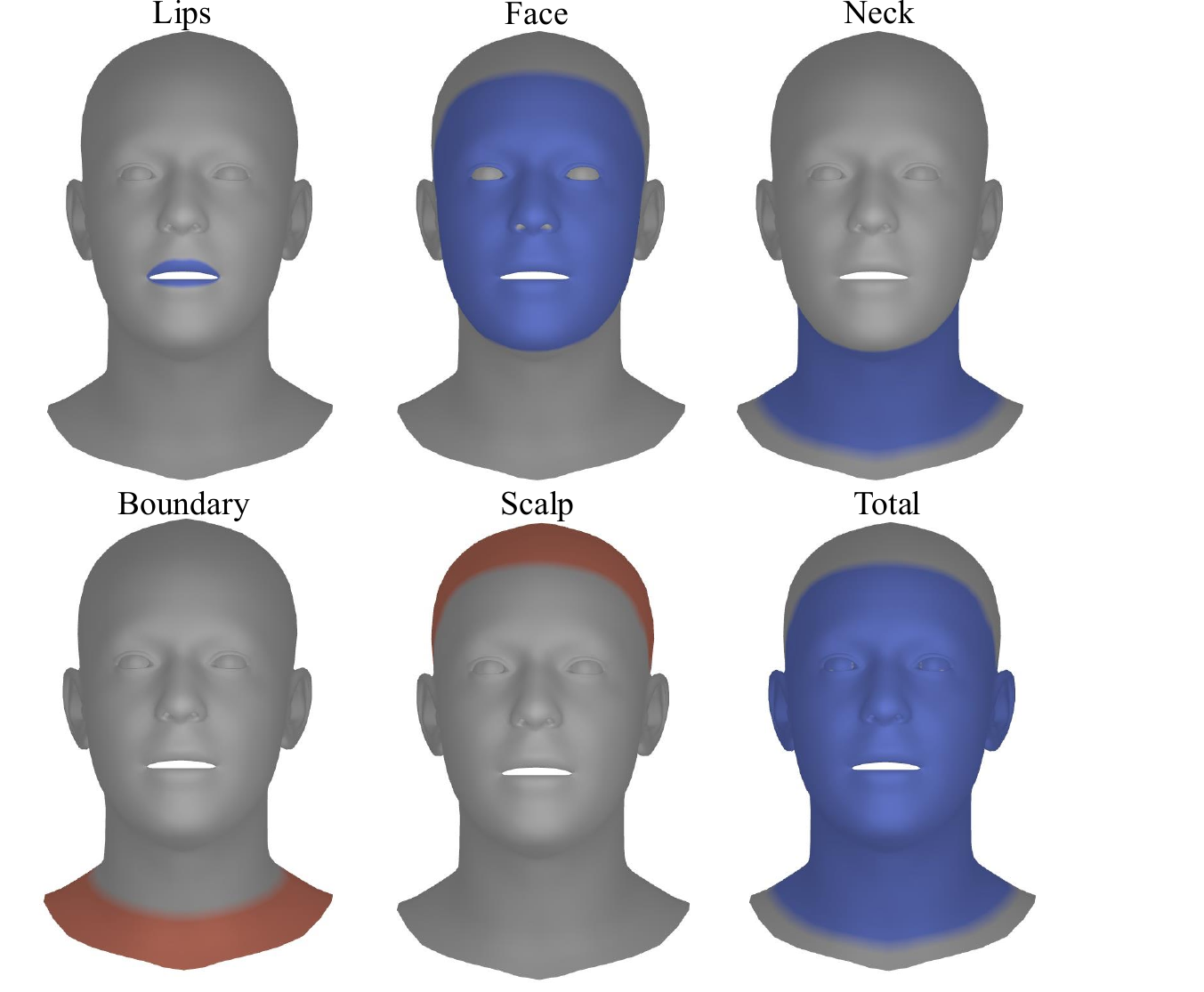}
    \caption{\textbf{Visualization of evaluation masks.} We utilize standard FLAME vertex masks to compute regional metrics. \textbf{Blue regions} indicate vertices included in the evaluation: (Top Row) Lips, Face, and Neck; (Bottom Right) the combined Head mask. \textbf{Red regions} indicate excluded vertices: (Bottom Left) the lower boundary and (Bottom Center) the scalp region, which are removed to ensure robust evaluation against capture noise and to ignore regions non-relevant to face reconstruction.}
    \label{fig:flame_masks}
    \vspace{-.5cm}
\end{figure}

\begin{table*}[t!]
\centering
\setlength{\tabcolsep}{6pt}
\footnotesize
\resizebox{\textwidth}{!}{%
\begin{tabular}{l|ccc|ccc|ccc|ccc}
\toprule
&
  \multicolumn{3}{c}{\textbf{Full w/o Scalp}} &
  \multicolumn{3}{c}{\textbf{Face}} &
  \multicolumn{3}{c}{\textbf{Lips}} &
  \multicolumn{3}{c}{\textbf{Neck}} \\
\cmidrule(lr){2-4}\cmidrule(lr){5-7}\cmidrule(lr){8-10}\cmidrule(lr){11-13}
\textbf{Name}
& Median $\downarrow$ & Mean $\downarrow$ & Std $\downarrow$
& Median $\downarrow$ & Mean $\downarrow$ & Std $\downarrow$
& Median $\downarrow$ & Mean $\downarrow$ & Std $\downarrow$
& Median $\downarrow$ & Mean $\downarrow$ & Std $\downarrow$ \\
\midrule
w/ p2s t=5            & 0.16                & 0.31              & 1.61             & 0.14                & 0.25              & 1.52             & 0.23                & 0.47              & 1.23             & 0.27                & 0.64              & 1.31             \\
w/ p2s t=10           & 0.13                & 0.27              & 1.58             & 0.11                & 0.21              & 1.50             & 0.18                & 0.38              & 1.15             & 0.22                & 0.56              & 1.24             \\
w/ p2s t=20           & 0.10                & 0.23              & 1.57             & 0.09                & 0.18              & 1.50             & 0.13                & 0.28              & 0.99             & 0.18                & 0.49              & 1.16             \\
w/ p2s t=50           & 0.07                & 0.18              & 1.55             & 0.07                & 0.15              & 1.49             & 0.08                & 0.19              & 0.89             & 0.15                & 0.42              & 1.08             \\
w/ p2s t=100          & \textbf{0.06}       & \textbf{0.16}     & 1.54             & 0.06                & 0.14              & 1.49             & 0.06                & \textbf{0.15}     & \textbf{0.85}    & 0.13                 & 0.39              & 1.04             \\ \midrule
w/o normals t=5       & 0.14                & 0.31              & 1.60             & 0.12                & 0.23              & 1.47             & 0.21                & 0.47              & 1.27             & 0.23                & 0.58              & 1.29             \\
w/o normals t=10      & 0.11                & 0.26              & 1.57             & 0.10                & 0.20              & 1.44             & 0.15                & 0.39              & 1.22             & 0.19                & 0.49              & 1.16             \\
w/o normals t=20      & 0.09                & 0.22              & 1.55             & 0.08                & 0.17              & 1.43             & 0.10                & 0.31              & 1.19             & 0.16                & 0.43              & 1.08             \\
w/o normals t=50      & 0.07                & 0.19              & 1.52             & 0.06                & 0.15              & 1.41             & 0.07                & 0.23              & 1.13             & 0.13                & 0.39              & 1.04             \\
w/o normals t=100     & \textbf{0.06}       & 0.17              & 1.51             & \textbf{0.05}       & \textbf{0.13}     & 1.40             & 0.06                & 0.20              & 1.10             & \textbf{0.12}       & \textbf{0.37}     & \textbf{1.03}    \\ \midrule
no-pretrain t=5       & 1.13                & 1.46              & 1.49             & 1.06                & 1.31              & 1.14             & 1.21                & 1.58              & 1.59             & 1.66                & 2.61              & 2.97             \\
no-pretrain t=10      & 1.02                & 1.38              & 1.50             & 0.94                & 1.23              & 1.15             & 1.12                & 1.55              & 1.63             & 1.51                & 2.54              & 3.03             \\
no-pretrain t=20      & 0.75                & 1.16              & 1.47             & 0.68                & 1.02              & 1.12             & 0.89                & 1.36              & 1.58             & 1.12                & 2.23              & 2.94             \\
no-pretrain t=50      & 0.44                & 0.94              & 1.46             & 0.39                & 0.81              & \textbf{1.10}    & 0.56                & 1.12              & 1.56             & 0.69                & 1.85              & 2.80             \\ 
no-pretrain t=100 & 0.32 & 0.89 & 1.48 & 0.28 & 0.77 & 1.15 & 0.38 & 1.02 & 1.56 & 0.5 & 1.67 & 2.73 \\ \midrule
TTO direct opt t=5   & 0.17 & 0.36 & 1.57 & 0.16 & 0.27 & 1.41 & 0.25 & 0.60 & 1.37 & 0.30 & 0.97 & 1.91 \\
TTO direct opt t=10  & 0.16 & 0.33 & 1.56 & 0.15 & 0.26 & 1.41 & 0.24 & 0.56 & 1.34 & 0.28 & 0.92 & 1.90 \\
TTO direct opt t=20  & 0.12 & 0.29 & 1.55 & 0.10 & 0.23 & 1.41 & 0.19 & 0.50 & 1.31 & 0.23 & 0.85 & 1.88 \\
TTO direct opt t=50  & 0.09 & 0.26 & 1.53 & 0.08 & 0.22 & 1.41 & 0.10 & 0.40 & 1.23 & 0.18 & 0.76 & 1.88 \\
TTO direct opt t=100 & 0.08 & 0.28 & 1.54 & 0.07 & 0.26 & 1.46 & 0.09 & 0.37 & 1.14 & 0.16 & 0.82 & 2.17 \\  \midrule
MOCHI TTO (full model) t=5        & 0.14                & 0.30              & 1.60             & 0.12                & 0.23              & 1.49             & 0.20                & 0.46              & 1.27             & 0.23                & 0.57              & 1.26             \\
MOCHI TTO (full model) t=10       & 0.11                & 0.26              & 1.58             & 0.09                & 0.20              & 1.46             & 0.14                & 0.37              & 1.22             & 0.19                & 0.48              & 1.14             \\
MOCHI TTO (full model) t=20       & 0.08                & 0.22              & 1.56             & 0.08                & 0.17              & 1.45             & 0.10                & 0.30              & 1.19             & 0.15                & 0.42              & 1.08             \\
MOCHI TTO (full model) t=50       & 0.07                & 0.18              & 1.53             & 0.06                & 0.14              & 1.43             & 0.07                & 0.23              & 1.13             & 0.13                & 0.38              & 1.04             \\
MOCHI TTO (full model) t=100      & \textbf{0.06}       & \textbf{0.16}     & 1.52             & \textbf{0.05}       & \textbf{0.13}     & 1.42             & \textbf{0.05}       & 0.19              & 1.09             & \textbf{0.12}       & \textbf{0.37}     & \textbf{1.03}    \\ \midrule
Classic Registrations & 0.10                & 0.24              & \textbf{1.44}    & 0.08                & 0.16              & 1.21             & 0.09                & 0.36              & 1.39             & 0.24                & 0.90              & 2.39             \\
\bottomrule
\end{tabular}%
}
\caption{\textbf{Quantitative ablation of Test-Time Optimization (TTO).} We evaluate the impact of different loss functions, supervision signals, and optimization strategies on the FaMoS validation set. We report the \textbf{Median}, \textbf{Mean}, and \textbf{Std} of the point-to-surface error (mm) for the full head (excluding scalp) and specific facial regions. \emph{Direct Opt} refers to initializing with MOCHI and optimizing vertices directly—rather than fine-tuning the parameters of the refinement model—while \emph{No-Pretrain} optimizes the network from random initialization. Our full model (MOCHI TTO) outperforms all baselines and surpasses the quality of classical registrations.}
\label{tab:tto_ablation}
\end{table*}

\paragraph{Evaluation Regions.}
We adopt the official vertex masks provided with the FLAME model~\cite{Li2017_FLAME} to define our regions of interest. Fig.~\ref{fig:flame_masks} visualizes the specific vertex subsets used in our quantitative analysis. The regions highlighted in \textbf{blue} (Lips, Face, Neck, and the combined Head) are used for calculating reconstruction errors. Conversely, the regions highlighted in \textbf{red} are excluded from evaluation. Specifically, we omit the scalp region due to the capture artifacts discussed previously, and the lower mesh boundary that are not relevant to facial reconstruction quality and can be affected by non-relevant to the head scan noise.

\begin{figure*}
    \centering
    \includegraphics[width=1.0\linewidth]{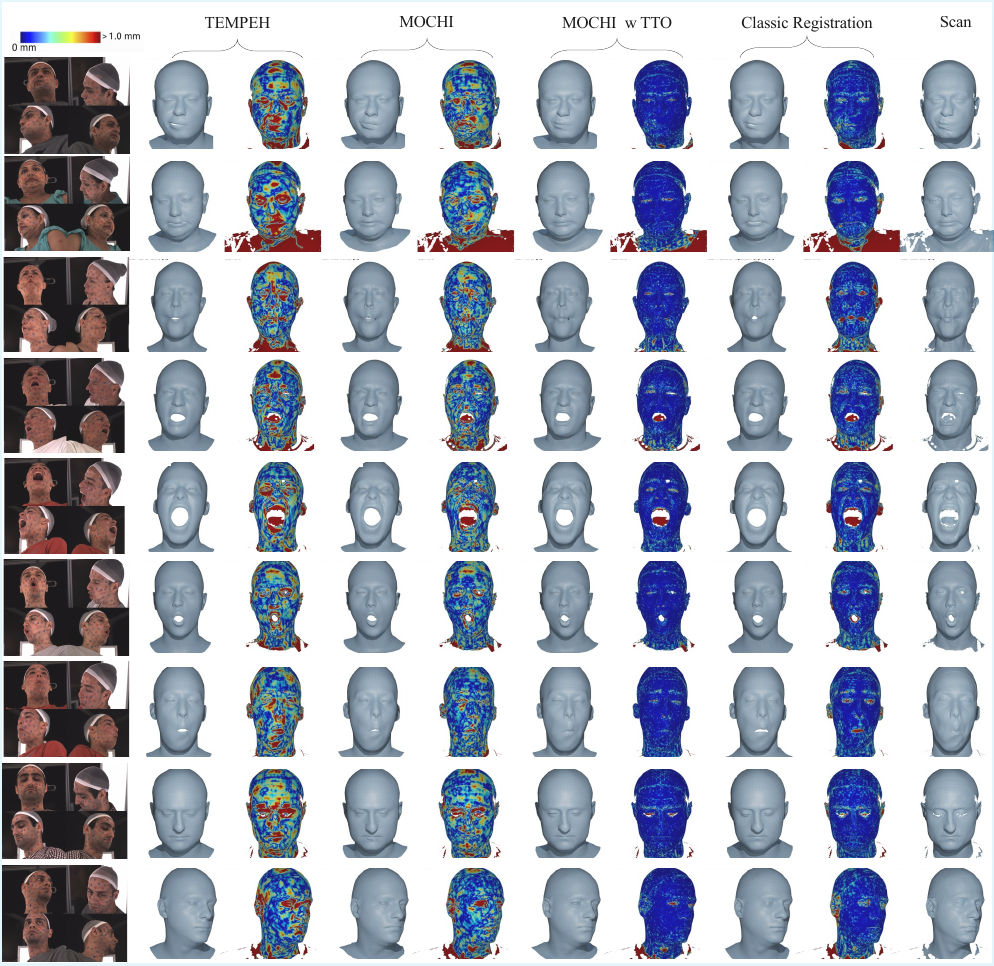}

    \caption{\textbf{Qualitative results on FaMoS validation and test sets (we show 4 of 8 input views).} From Left to Right: \emph{TEMPEH}~\cite{Bolkart2023_TEMPEH}, \emph{MOCHI}, \emph{MOCHI + TTO} (50 iters), \emph{Classic Registration}, and the \emph{Scan}. For each method we show the predicted mesh (left) and the point-to-surface error heatmap on the scan (right; red $\geq$ 1.0 mm, lower is better). MOCHI improves over TEMPEH on unseen subjects even without using registrations for training; adding TTO further reduces error and can surpass the classic manual registration pipeline.}
    \label{fig:famos_qual}
\end{figure*}

\section{Extended Ablation Studies}

\begin{figure*}
    \centering
    \includegraphics[width=1.0\linewidth]{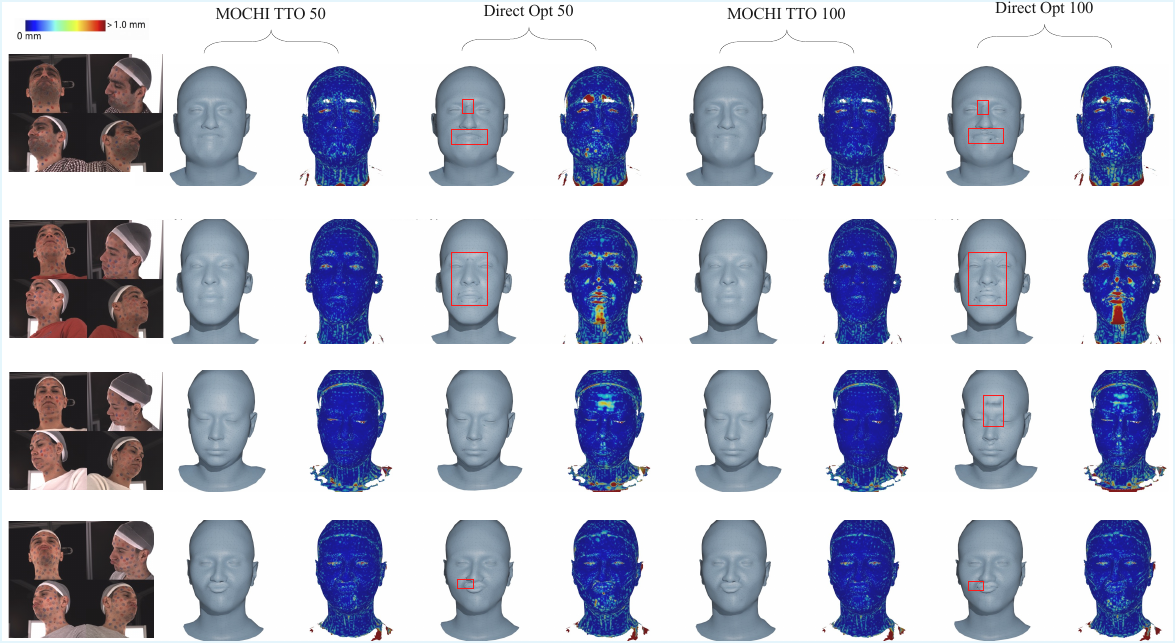}

    \caption{\textbf{Robustness of optimization strategies.} We compare our latent-space optimization (MOCHI TTO) against directly optimizing mesh vertices (Direct Opt) at 50 and 100 iterations. While Direct Opt reduces the numerical point-to-surface error, it is prone to  artifacts, spikes, and surface irregularities (highlighted by red boxes). In contrast, MOCHI TTO leverages the network's learned prior, ensuring the reconstruction remains smooth and anatomically plausible while accurately fitting the target scan.}    \label{fig:robust_tto}
\end{figure*}

\begin{figure*}
    \centering
    \includegraphics[width=1.0\linewidth]{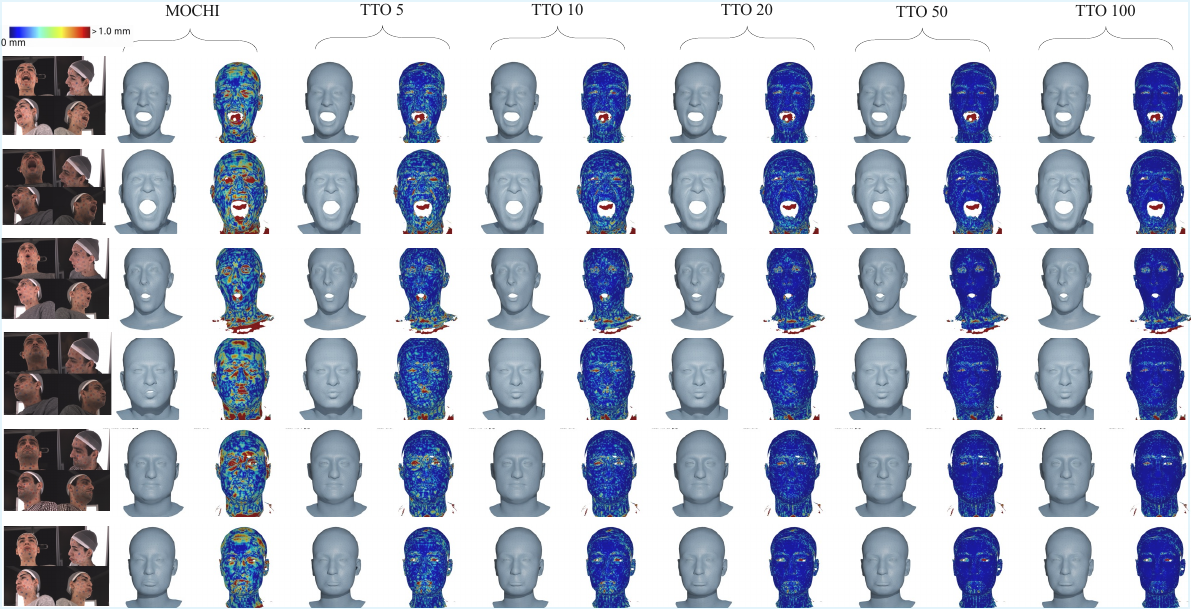}
    \caption{\textbf{Qualitative progression of Test-Time Optimization.} This figure shows the evolution of mesh refinement over 100 iterations on example images from FaMoS test set.}
        \label{fig:tto}
\end{figure*}
\paragraph{Extra Qualitative Comparisons.}
Figure~\ref{fig:famos_qual} presents additional qualitative results on FaMoS validation and test subjects, comparing MOCHI (with and without TTO at 50 iterations) against TEMPEH~\cite{Bolkart2023_TEMPEH} and classic registrations. We show a broad range of examples and expressions, further demonstrating MOCHI’s accuracy and generalization.

\paragraph{Test-Time Optimization (TTO) Strategies.}
In Table~\ref{tab:tto_ablation}, we provide the complete set of numerical results corresponding to Fig.~7 in the main text. Furthermore, we introduce the \textbf{No-Pretrain} baseline where the local refinement module is randomly initialized and optimized from scratch for each scan, instead of being initialized from pre-trained MOCHI weights.

Without MOCHI initialization (No-Pretrain), performance degrades significantly, underlining the importance of learned priors. 

Additionally, Fig.~\ref{fig:tto} shows how reconstruction quality improves progressively from step 5 to 100. 


\paragraph{Direct Vertex Optimization}
Fig.~\ref{fig:robust_tto} illustrates the robustness of our TTO scheme. Under identical conditions, direct vertex optimization leads to high-frequency artifacts, while our latent TTO produces smooth, topologically valid surfaces.




\paragraph{Chamfer Distance}
While prior methods~\cite{Bolkart2023_TEMPEH,Li2024_GRAPE} primarily rely on point-to-surface error, the Chamfer distance is another prevalent metric in 3D vision. However, similar to point-to-surface error, the Chamfer distance relies on non-differentiable nearest-neighbor selection, which can cause artifacts and can make training more unstable. 

\begin{figure}[h]
    \centering
    \includegraphics[width=0.8\linewidth]{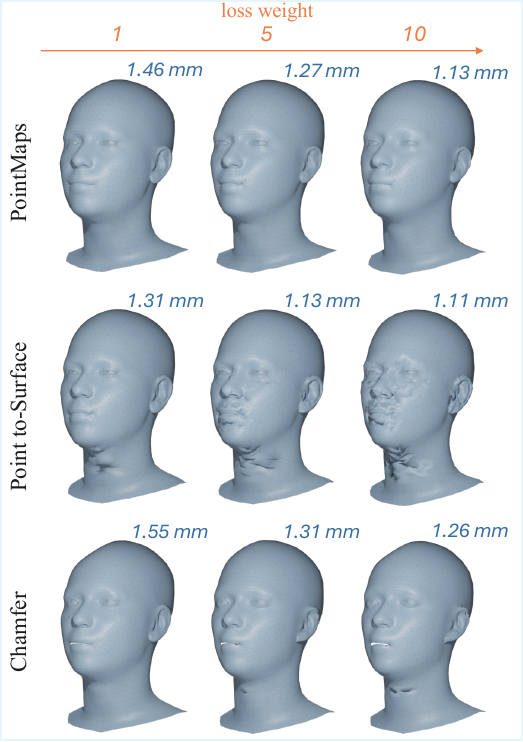}
    
    \caption{\textbf{Impact of loss type and weight.} We visualize the training of the coarse model using point maps (top), point-to-surface loss (middle), and Chamfer loss (bottom). When starting from a coarse prediction initialized only with landmark pretraining, increasing the point-to-surface loss weight leads to severe artifacts. While Chamfer distance results in fewer artifacts, it can still introduce errors (e.g., in the ears); in contrast, point-map supervision remains stable and topology-consistent. Blue numbers indicate the point-to-surface error (mm) relative to the ground-truth scan.}
    \label{fig:s2m_chamfer}
\end{figure}

In Fig.~\ref{fig:s2m_chamfer}, we compare the training dynamics of the coarse stage over 500 iterations using three different objectives: the conventional point-to-surface loss, the Chamfer loss, and our differentiable point-map loss. As the weight of the point-to-surface loss increases, reconstructions rapidly develop visible artifacts and self-intersections, effectively bypassing the topology enforcement of our regularizers. Similarly, while the Chamfer distance avoids the severe topological collapses seen with point-to-surface loss, it still induces artifacts and noise (e.g., in the ears and eyeballs). In contrast, supervision with point-maps exhibits significantly smoother behavior, preserving clean geometry and stable gradients even at higher loss weights.

\begin{figure}[h]
    \centering
    \includegraphics[width=0.6\linewidth]{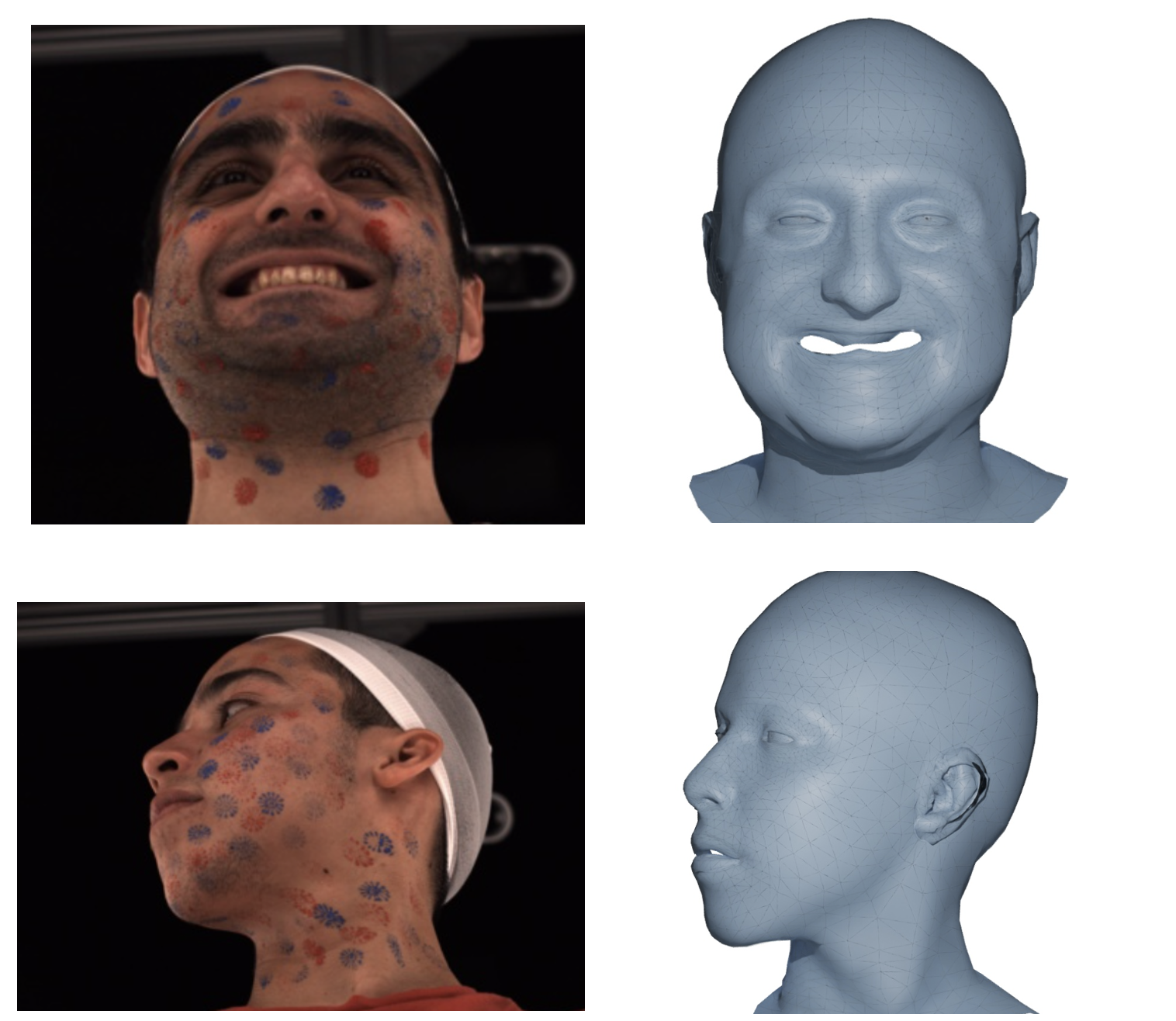}
    \caption{\textbf{Failure cases.} \textbf{Top Row (MOCHI TTO):} Since the FLAME topology lacks explicit teeth geometry, test-time optimization may incorrectly distort the lips by pulling them inward to minimize the point-to-surface distance to the teeth present in the raw scan. \textbf{Bottom Row (MOCHI):} In cases of large head poses, the base model may occasionally present with artifacts.}
    \label{fig:failures}
\end{figure}

\qheading{Limitations}
We visualize typical failure cases of our method in Fig.~\ref{fig:failures}. While MOCHI TTO generally refines geometry, it can encounter difficulties when the raw scan contains teeth, which are not represented in the FLAME model. Additionally, large head poses can occasionally lead to artifacts in the base model. Such cases can happen more often when we train MOCHI without 2D supervision from the predicted landmarks of the dense landmark tracker.

\section{Additional Dataset Evaluation}
\label{sec:additional_datasets}

\begin{figure}[h]
    \centering
    \includegraphics[trim=0 50 0 0, clip,width=1.0\linewidth]{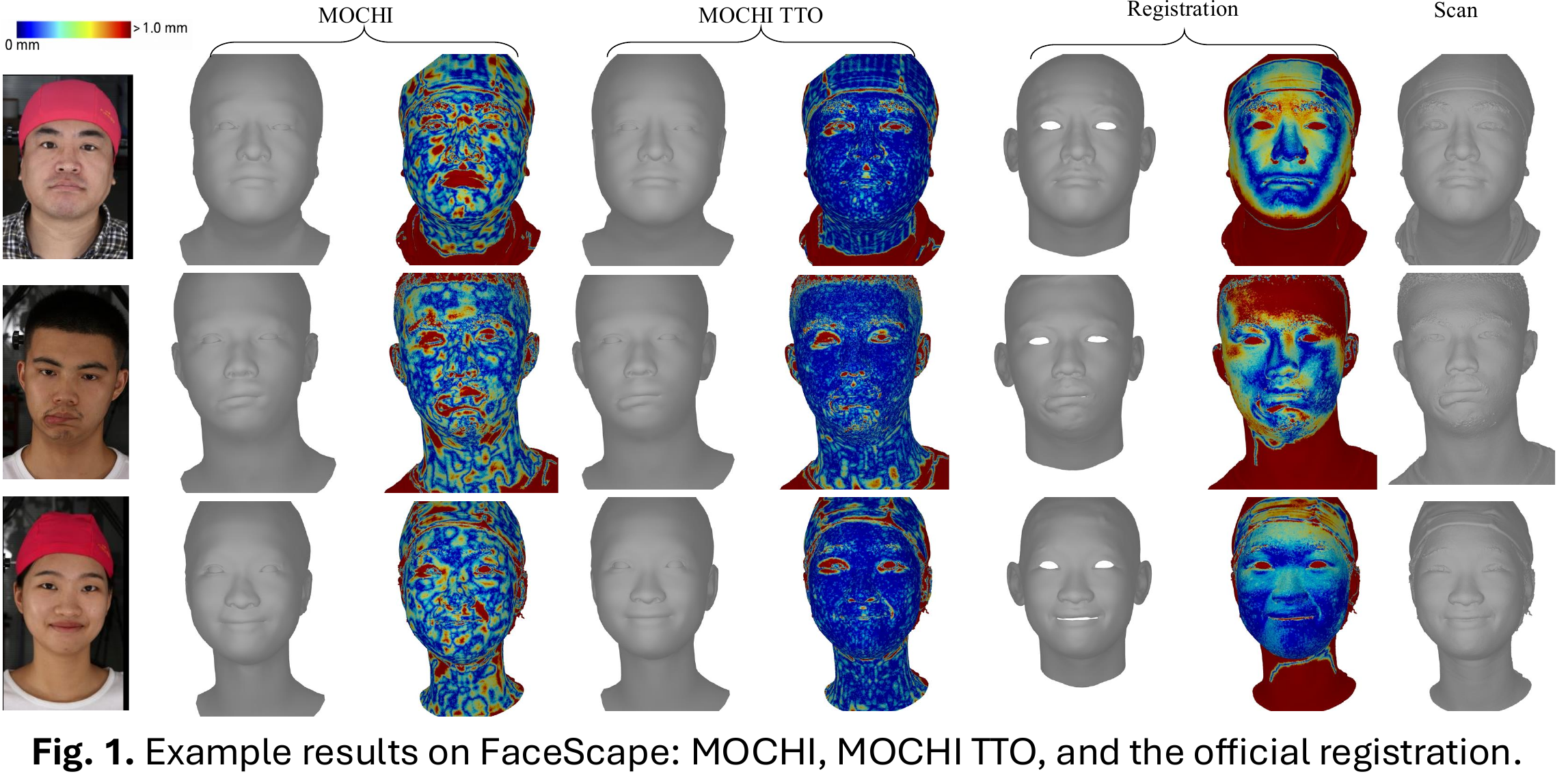}
    \caption{\textbf{Qualitative results on FaceScape.} From left to right: \emph{MOCHI}, \emph{MOCHI TTO}, the official \emph{FaceScape registration}, and the \emph{Scan}. For each method we show the predicted mesh (left) and the point-to-surface error heatmap (right; red $\geq$ 1.0\,mm, lower is better).}
    \label{fig:facescape}
\end{figure}

\paragraph{FaceScape.}
To further validate generalization, we evaluate MOCHI on FaceScape~\cite{yang2020facescape}, which uses a different capture rig with per-sample self-calibration via structure-from-motion. This results in inconsistent camera parameters and arbitrary scale across samples. To evaluate on FaceScape, we align scans to a canonical frame and fine-tune our model. Since the official registrations are also unaligned, we rigidly align them to the scans for fair comparison. Qualitative results comparing MOCHI, MOCHI-TTO, and the official FaceScape registration are shown in Fig.~\ref{fig:facescape}. As can be seen, MOCHI-TTO produces registrations that are more accurate than the official FaceScape registrations, confirming that our method generalizes beyond the FaMoS capture setup.

\begin{figure*}[h]
    \centering

    \includegraphics[width=0.23\columnwidth]{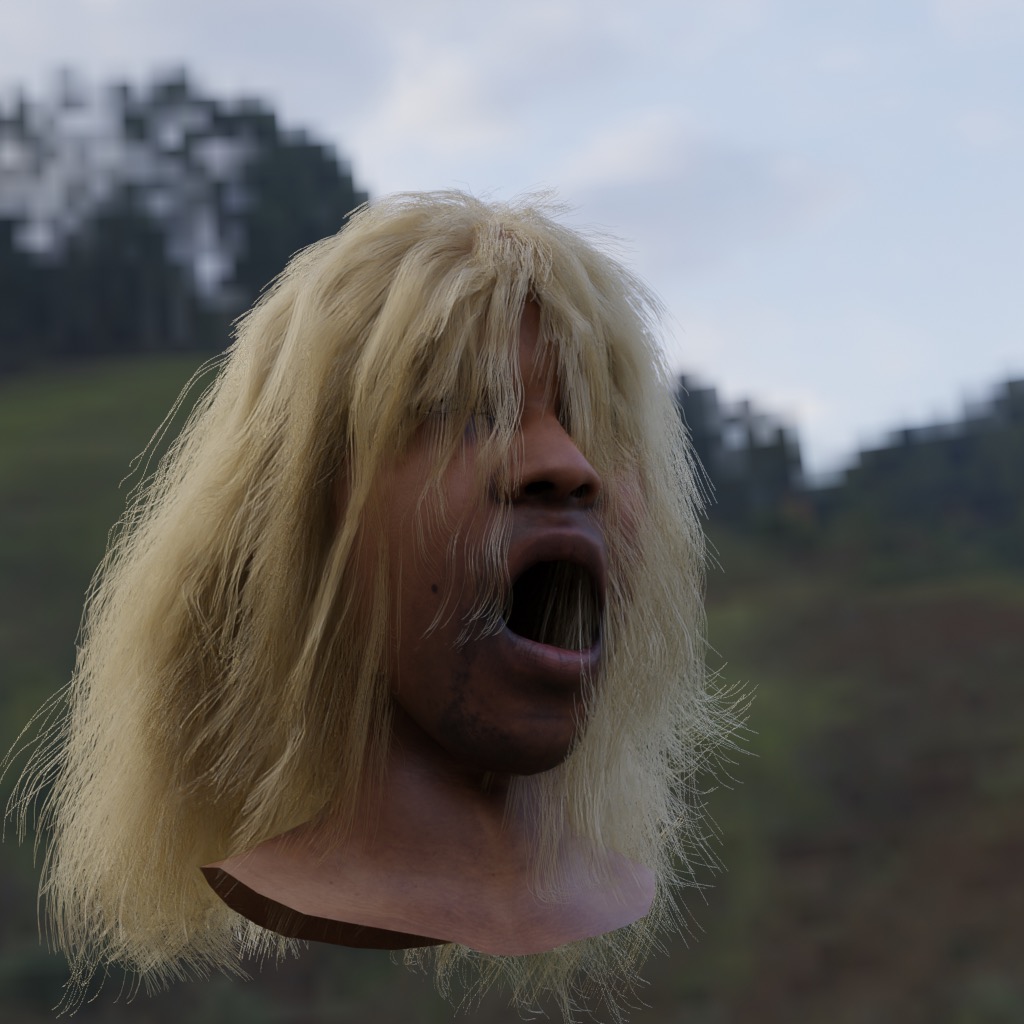}
    \includegraphics[width=0.23\columnwidth]{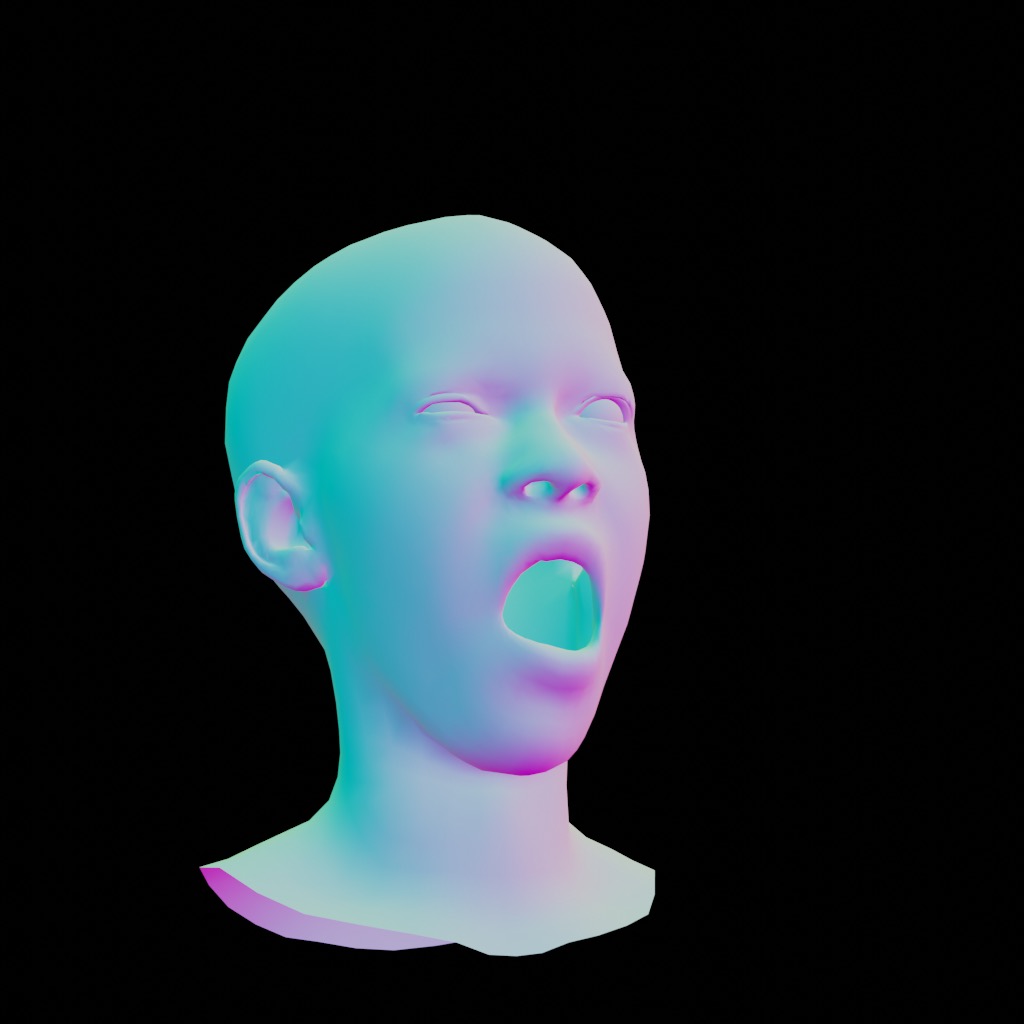}
    \includegraphics[width=0.23\columnwidth]{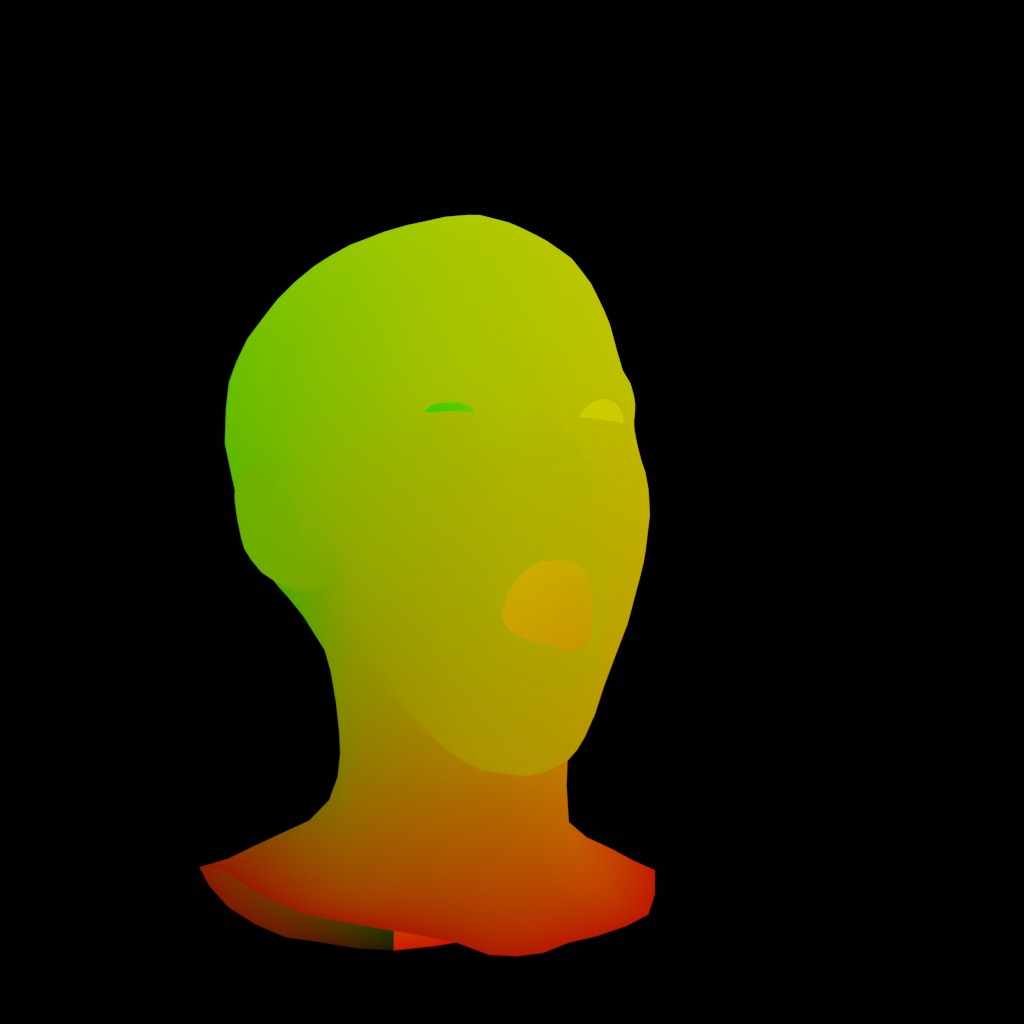}
    \includegraphics[width=0.23\columnwidth]{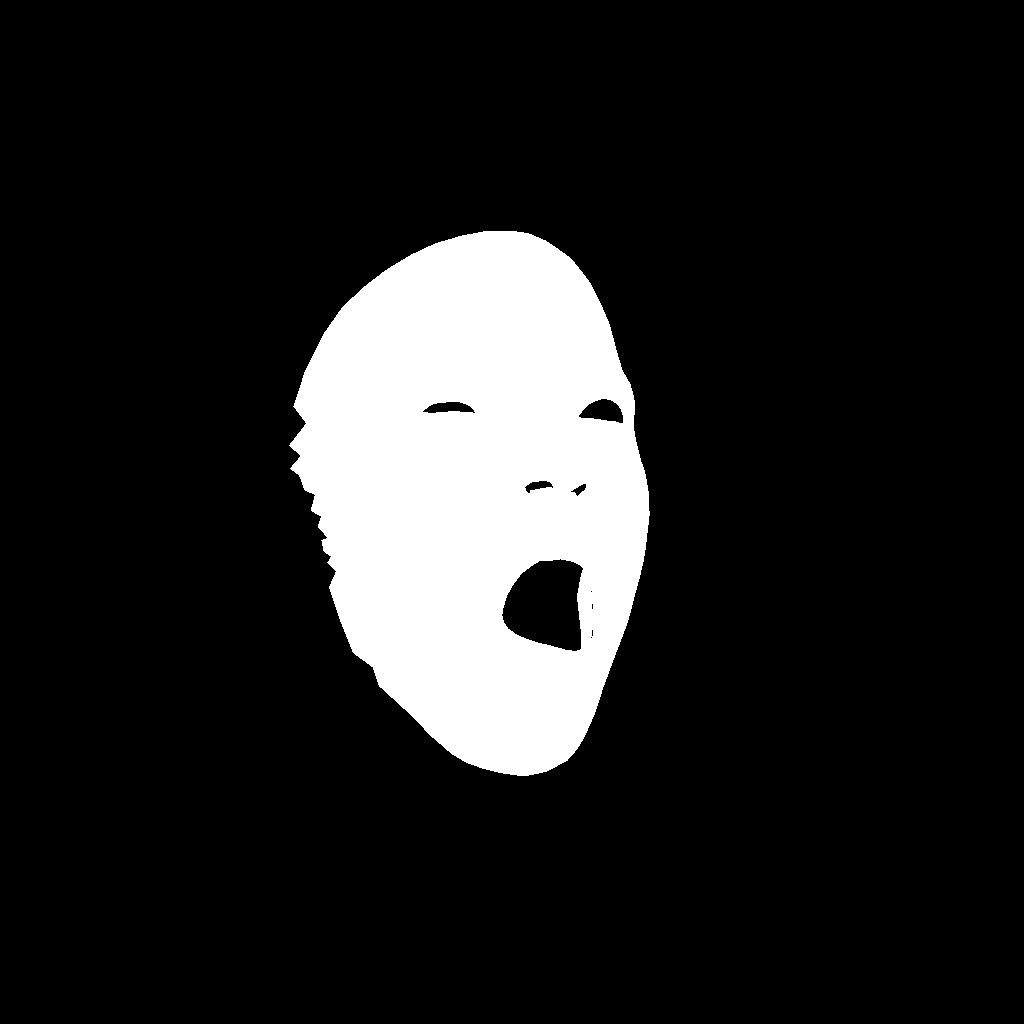}
    \includegraphics[width=0.23\columnwidth]{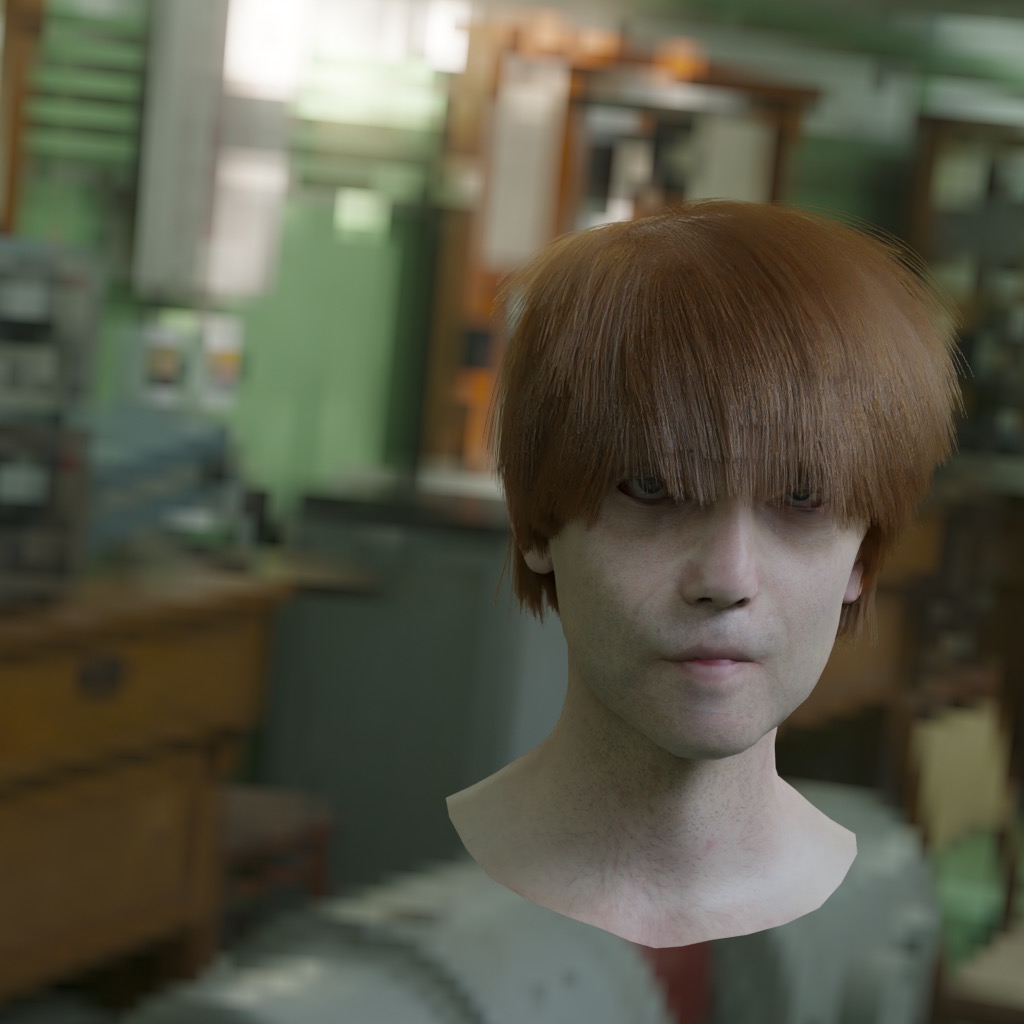}
    \includegraphics[width=0.23\columnwidth]{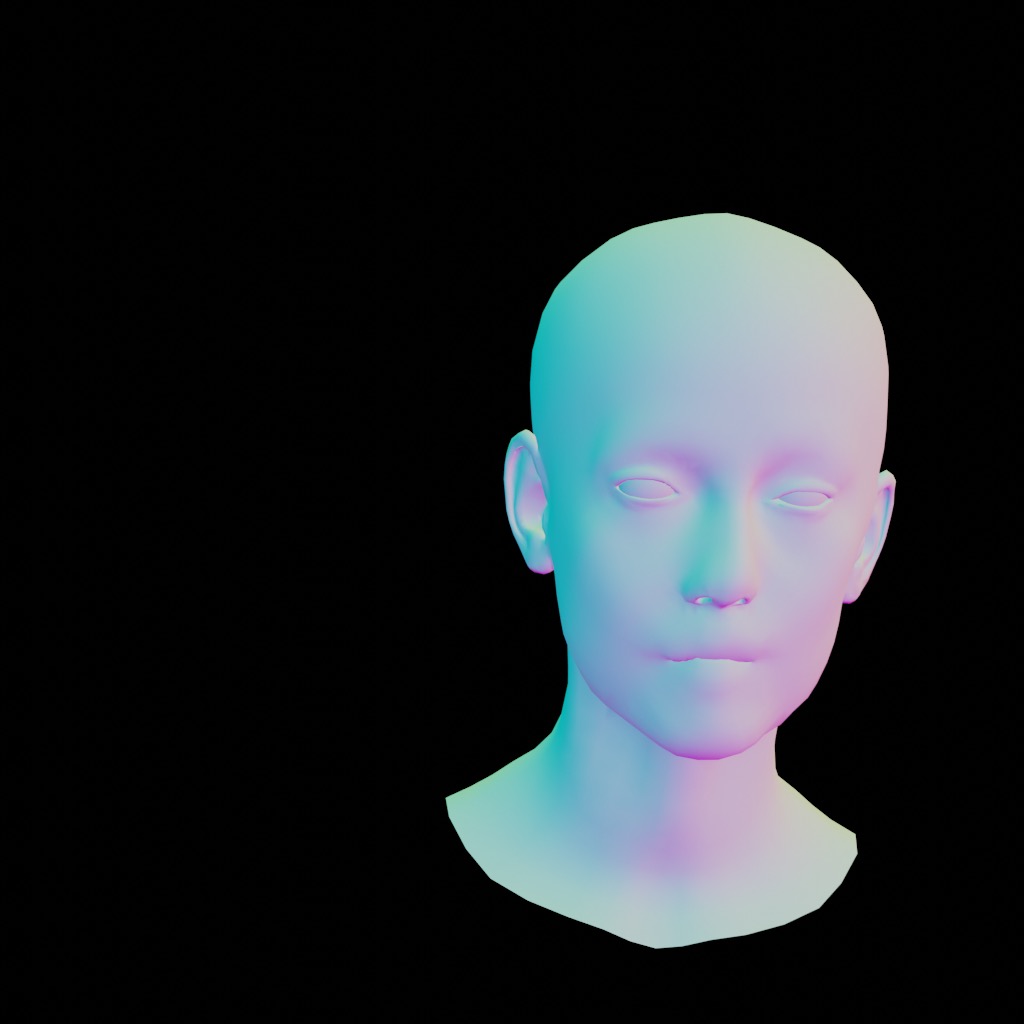}
    \includegraphics[width=0.23\columnwidth]{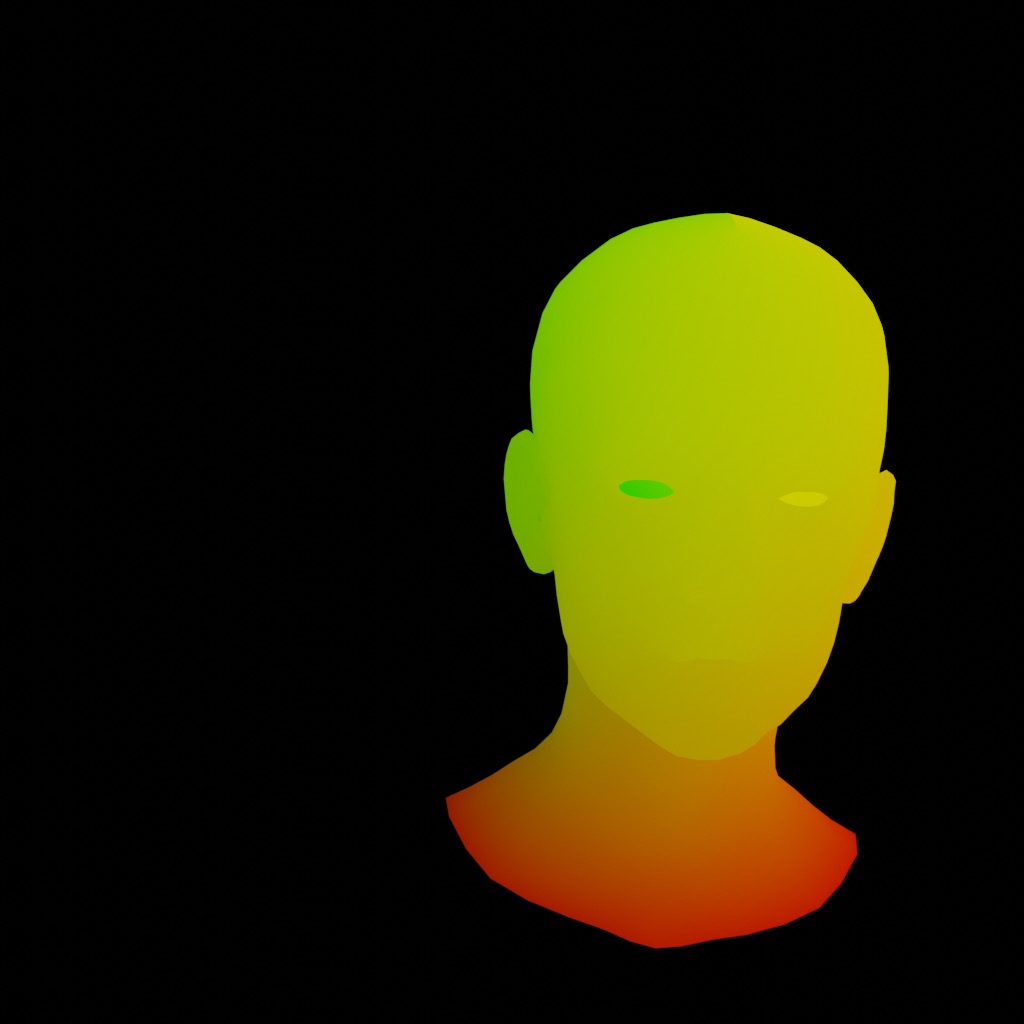}
    \includegraphics[width=0.23\columnwidth]{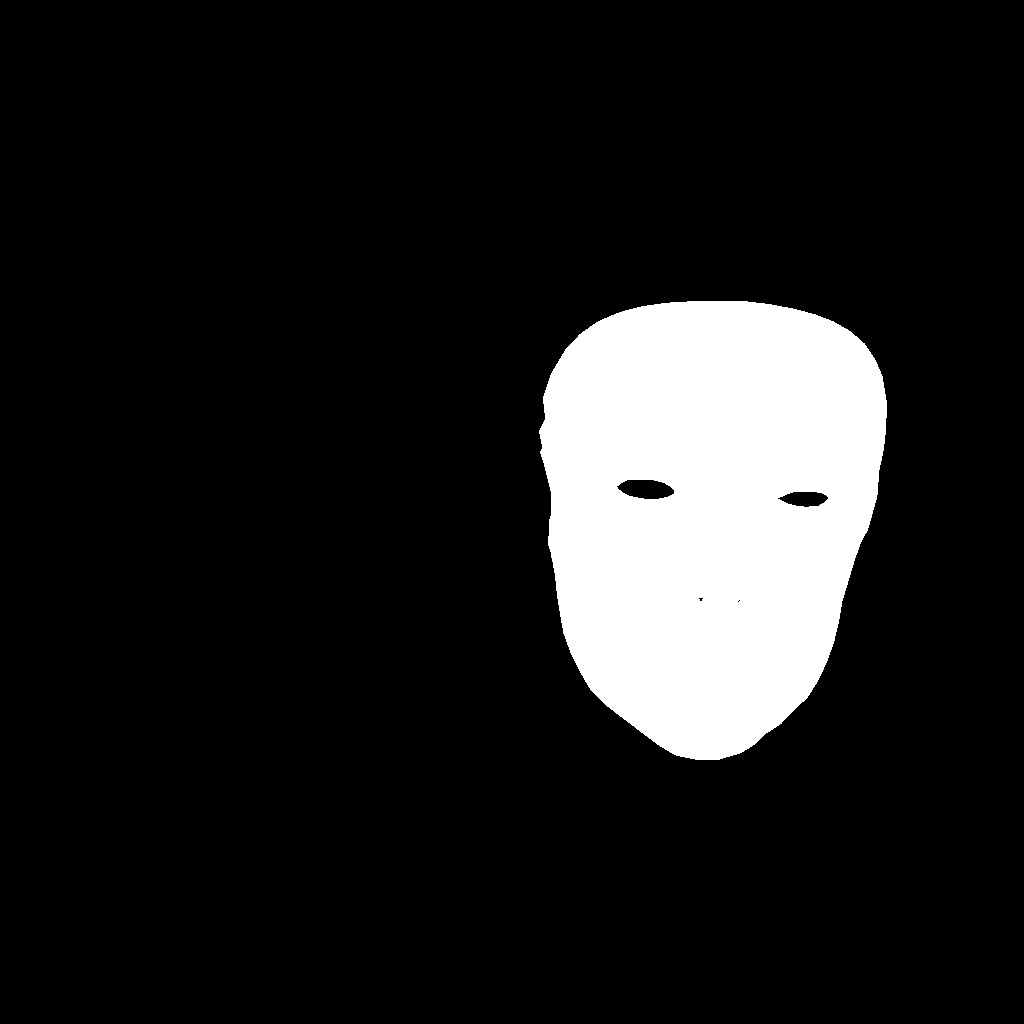}
    \includegraphics[width=0.23\columnwidth]{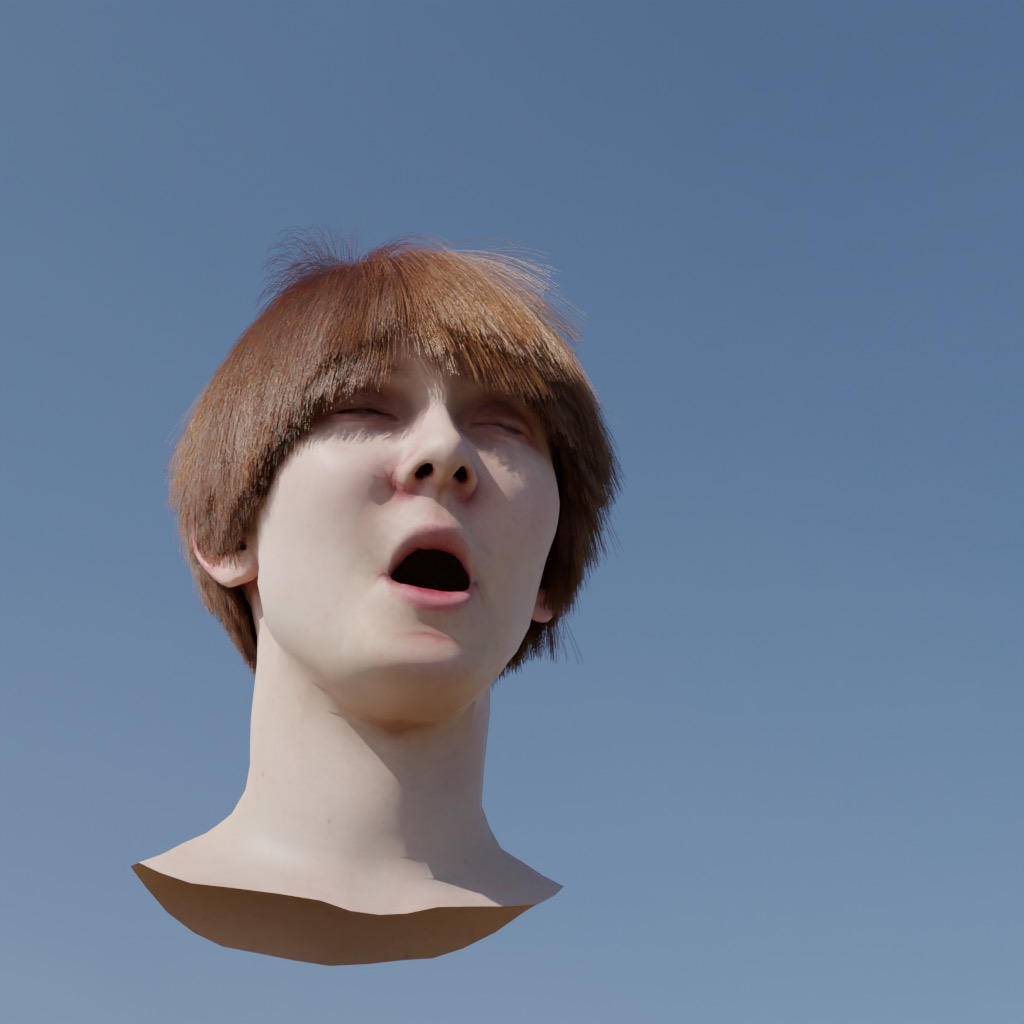}
    \includegraphics[width=0.23\columnwidth]{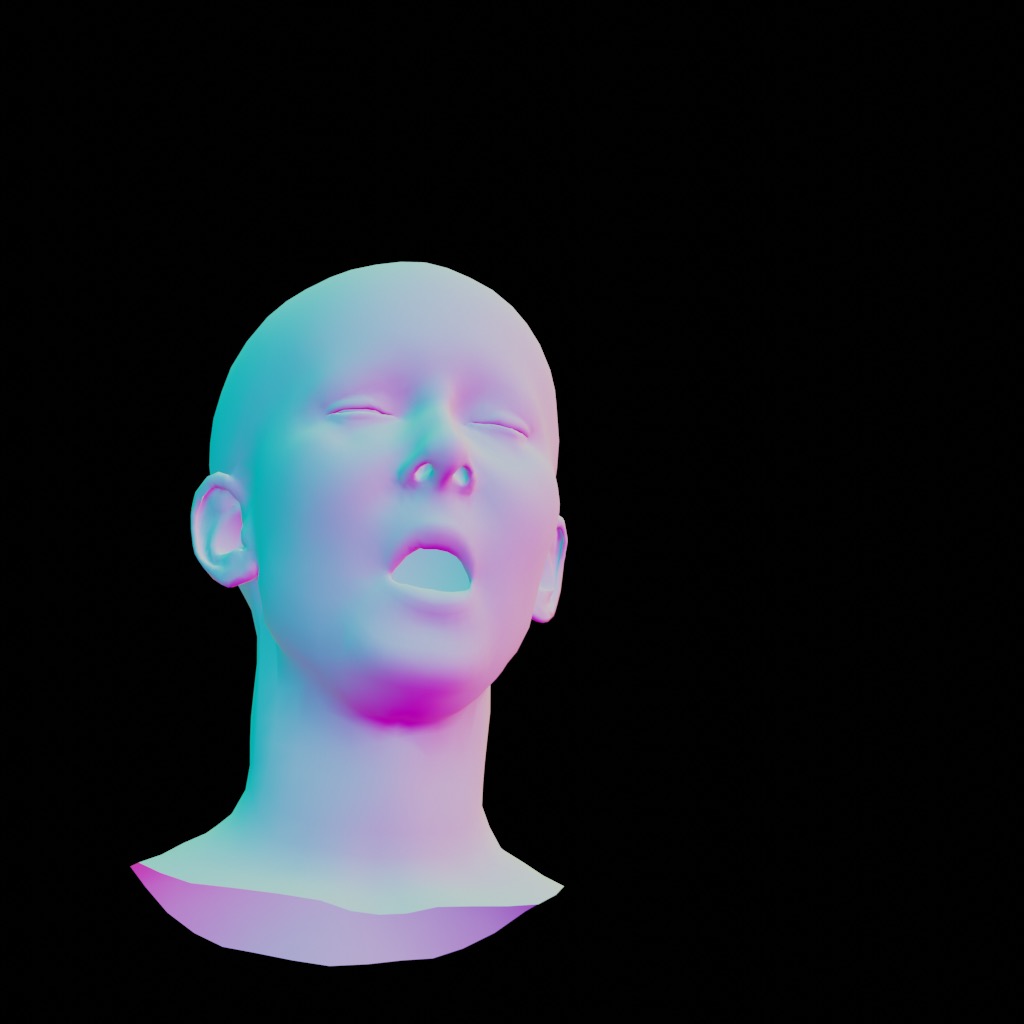}
    \includegraphics[width=0.23\columnwidth]{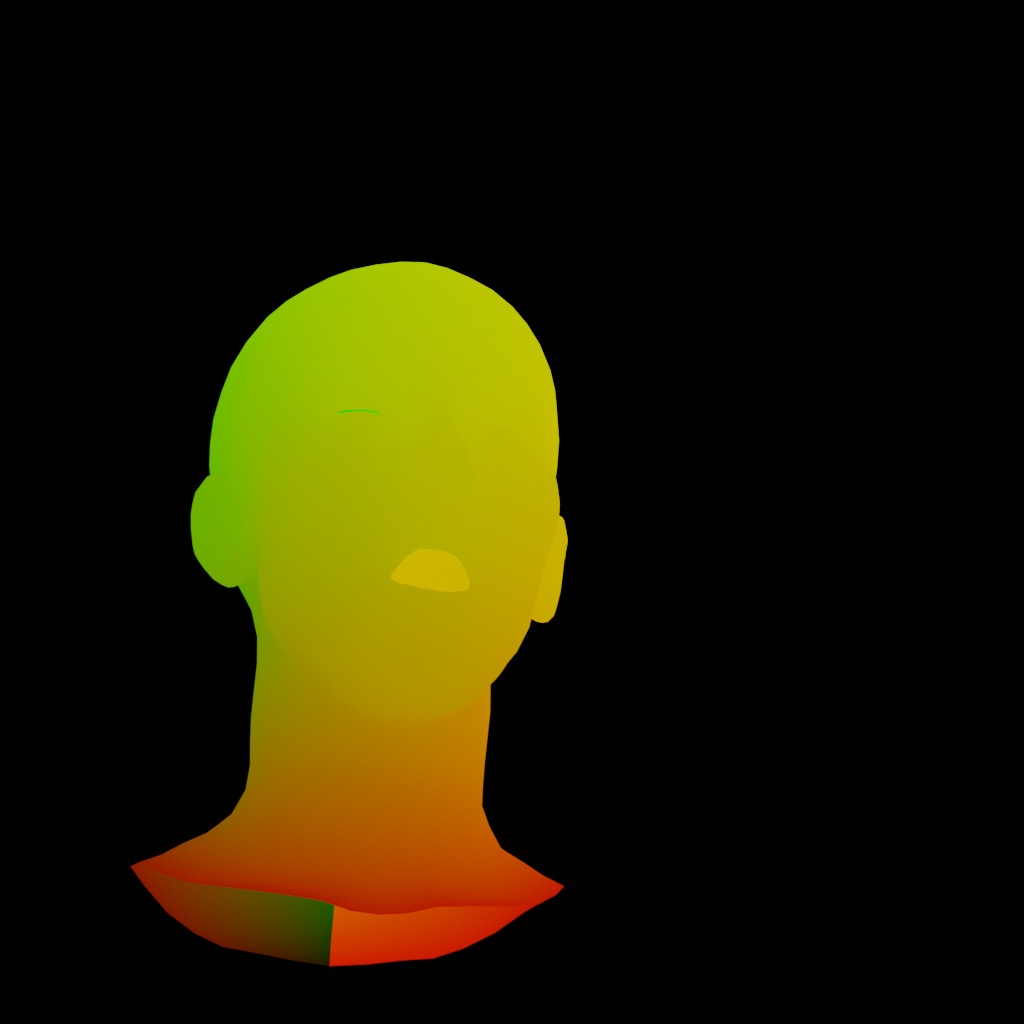}
    \includegraphics[width=0.23\columnwidth]{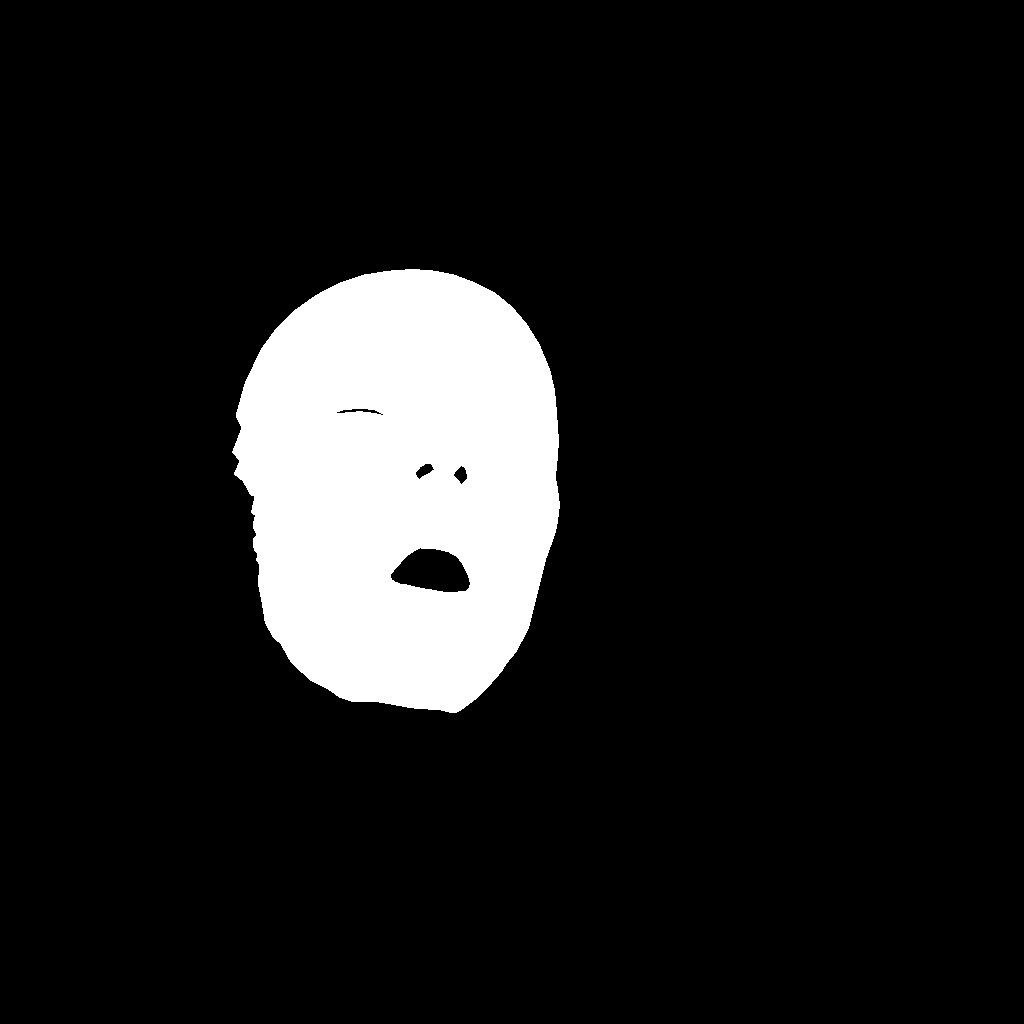}
    \includegraphics[width=0.23\columnwidth]{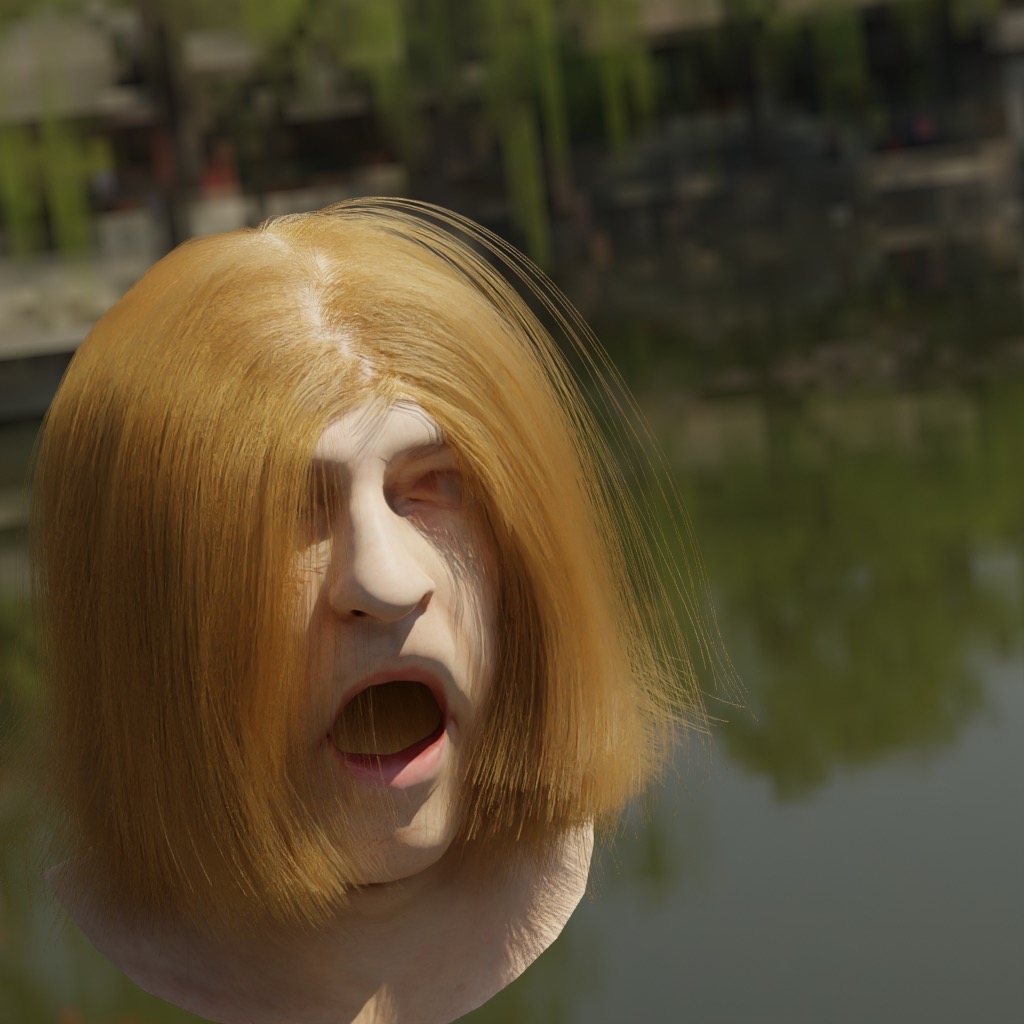}
    \includegraphics[width=0.23\columnwidth]{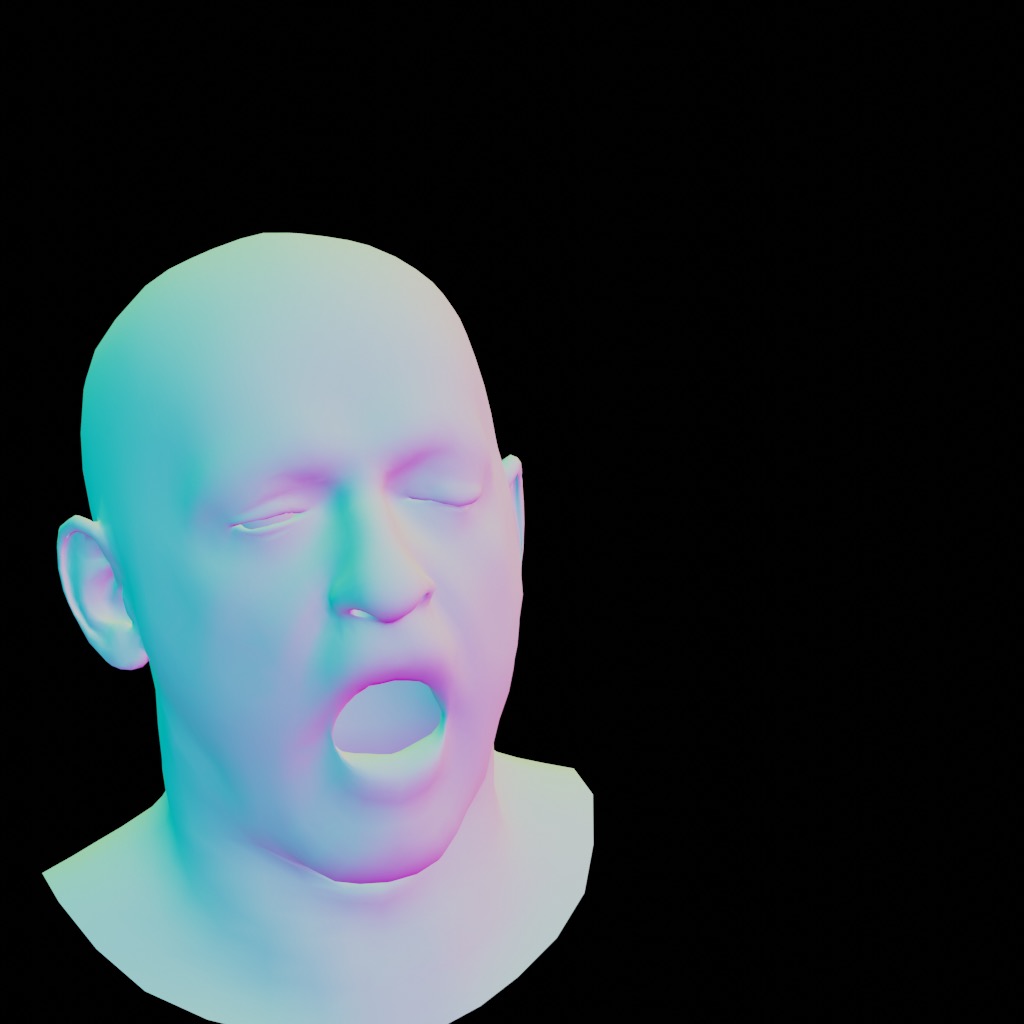}
    \includegraphics[width=0.23\columnwidth]{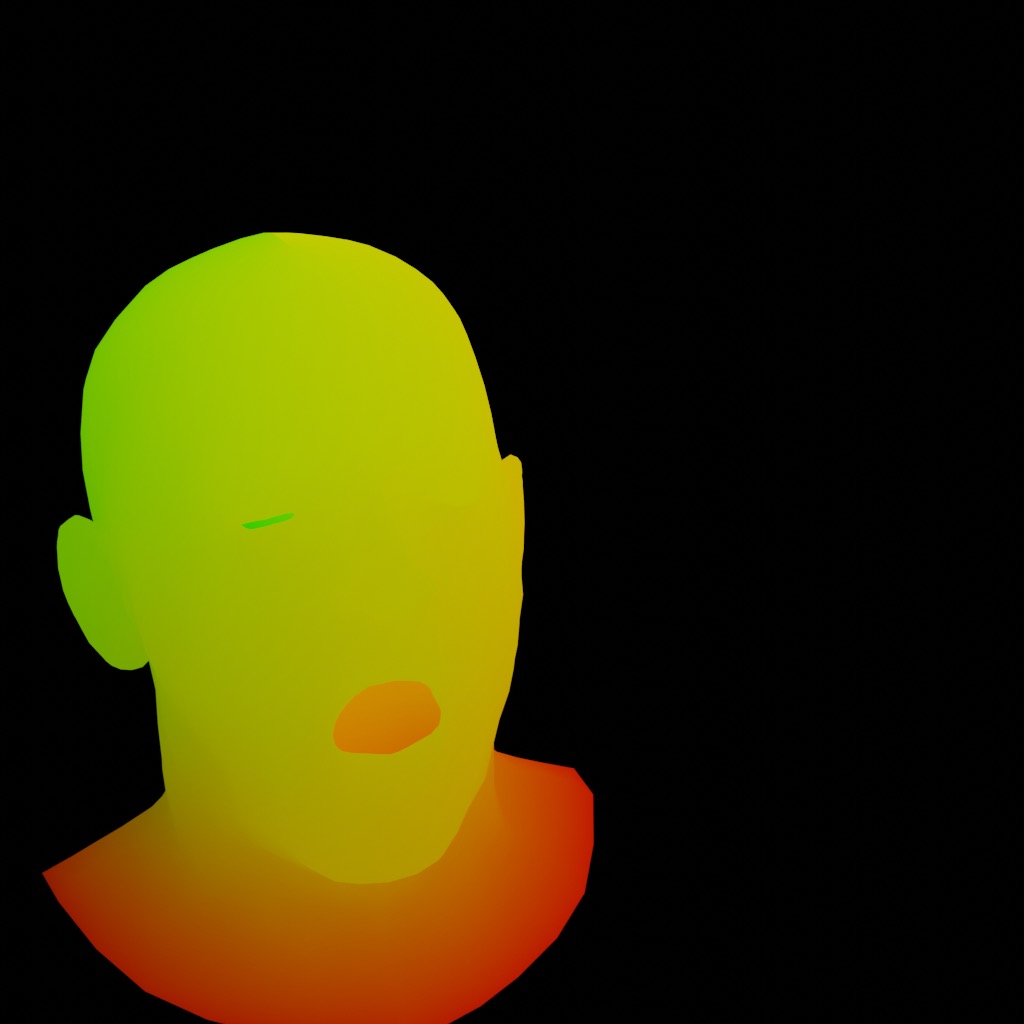}
    \includegraphics[width=0.23\columnwidth]{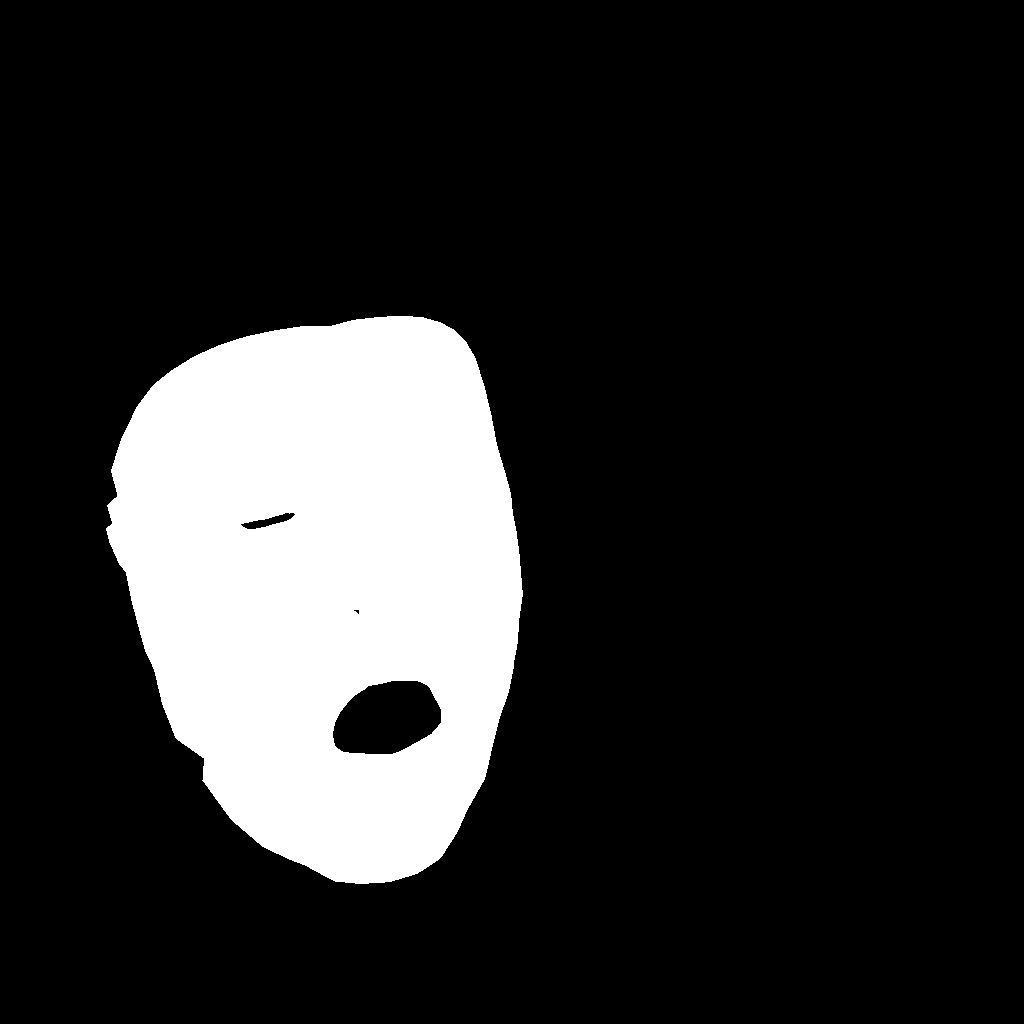}
    \includegraphics[width=0.23\columnwidth]{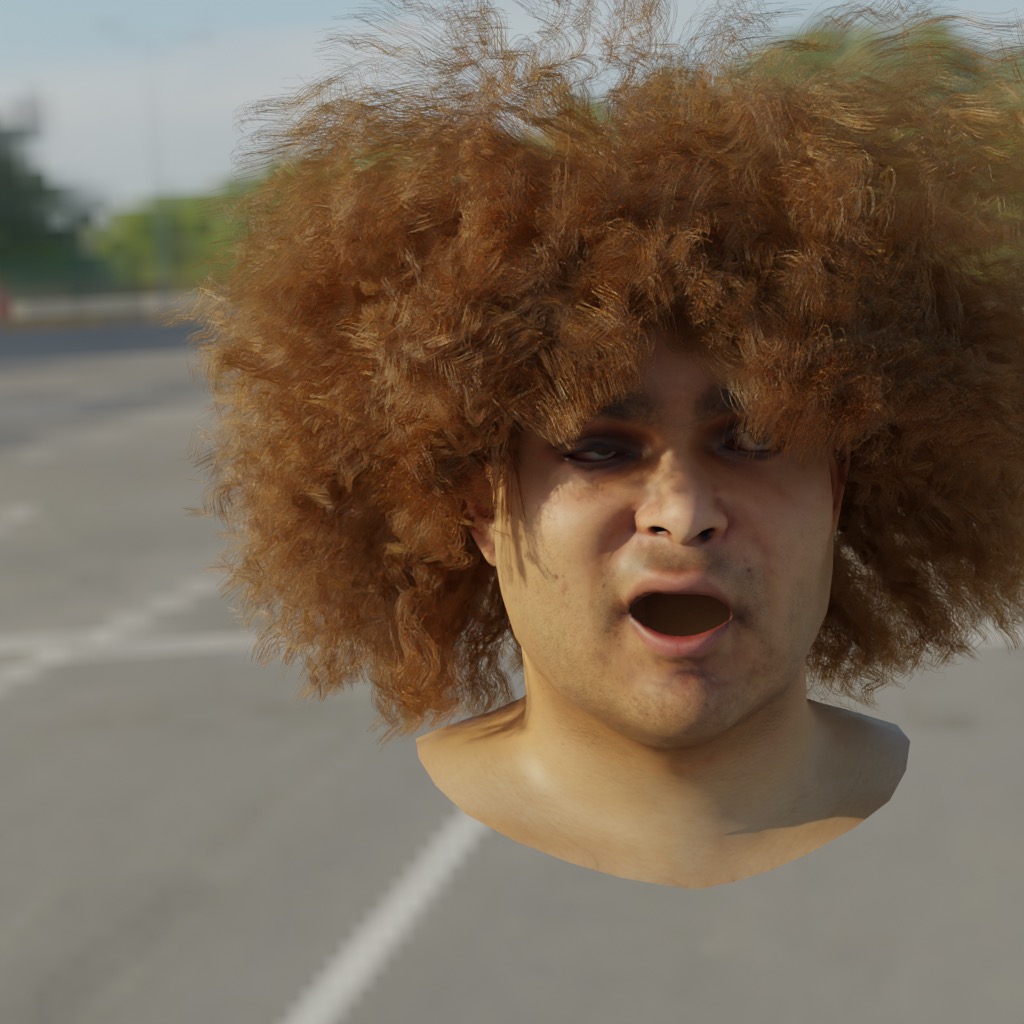}
    \includegraphics[width=0.23\columnwidth]{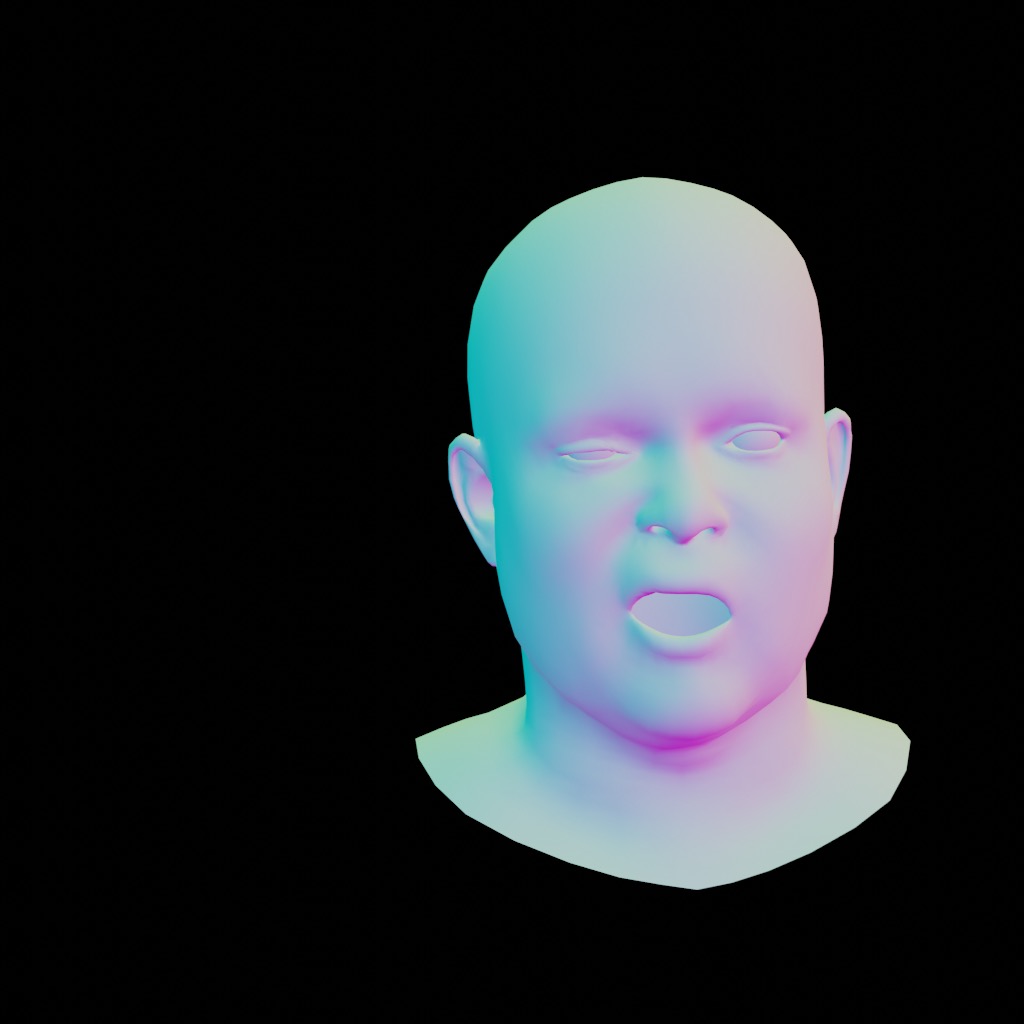}
    \includegraphics[width=0.23\columnwidth]{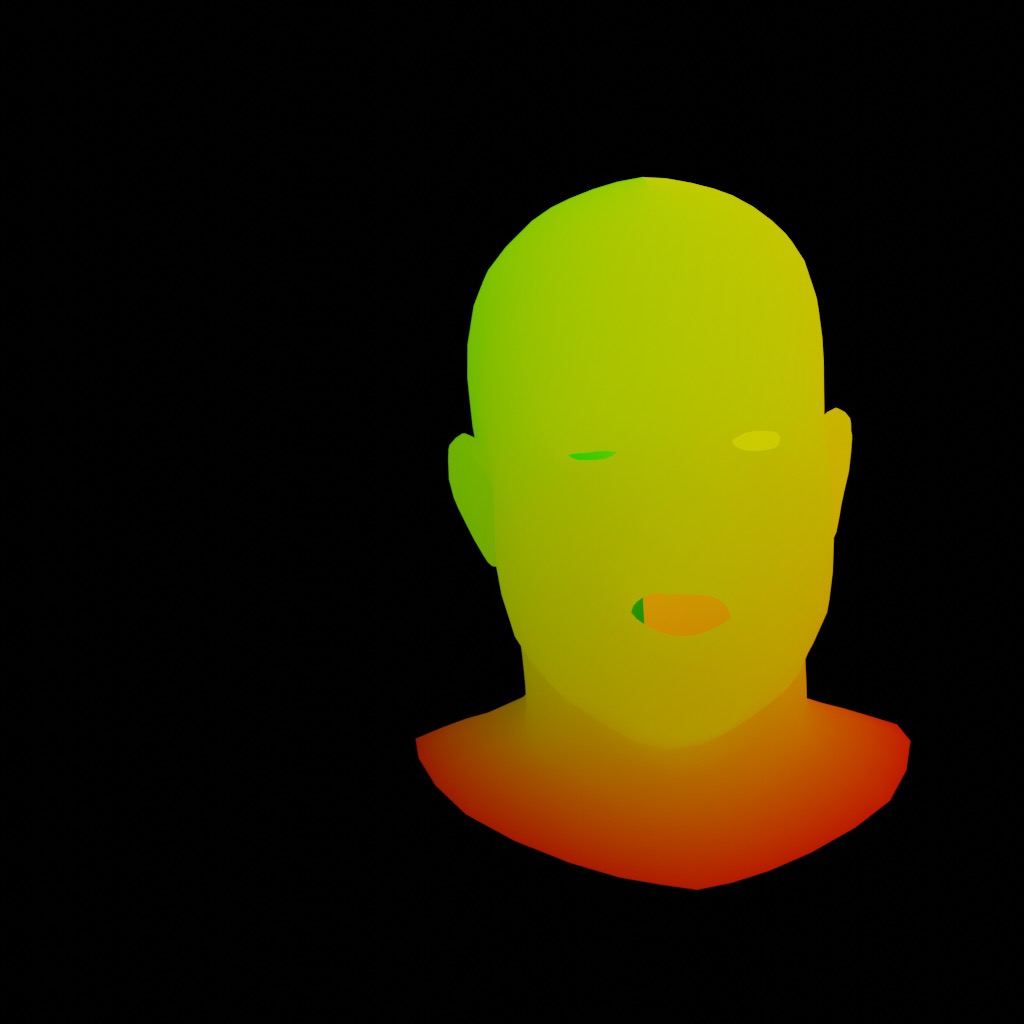}
    \includegraphics[width=0.23\columnwidth]{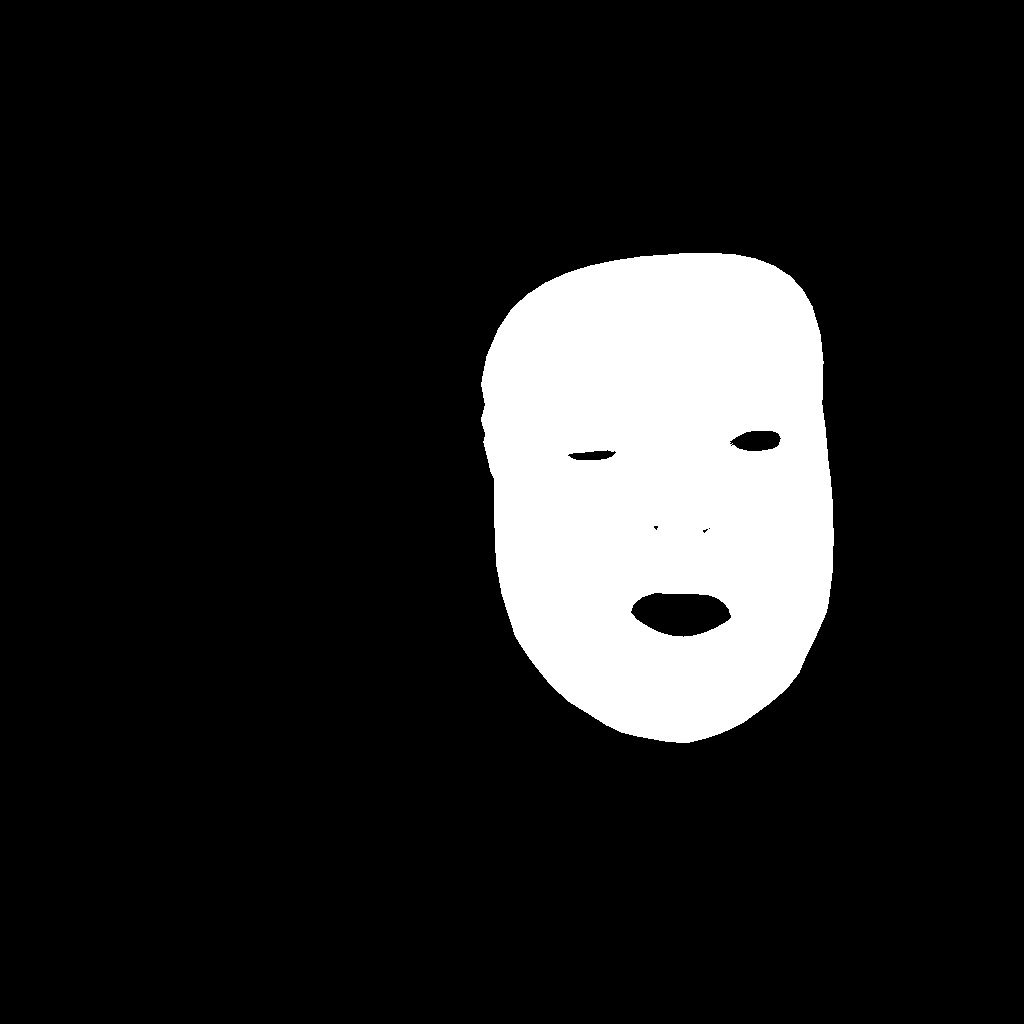}
    \includegraphics[width=0.23\columnwidth]{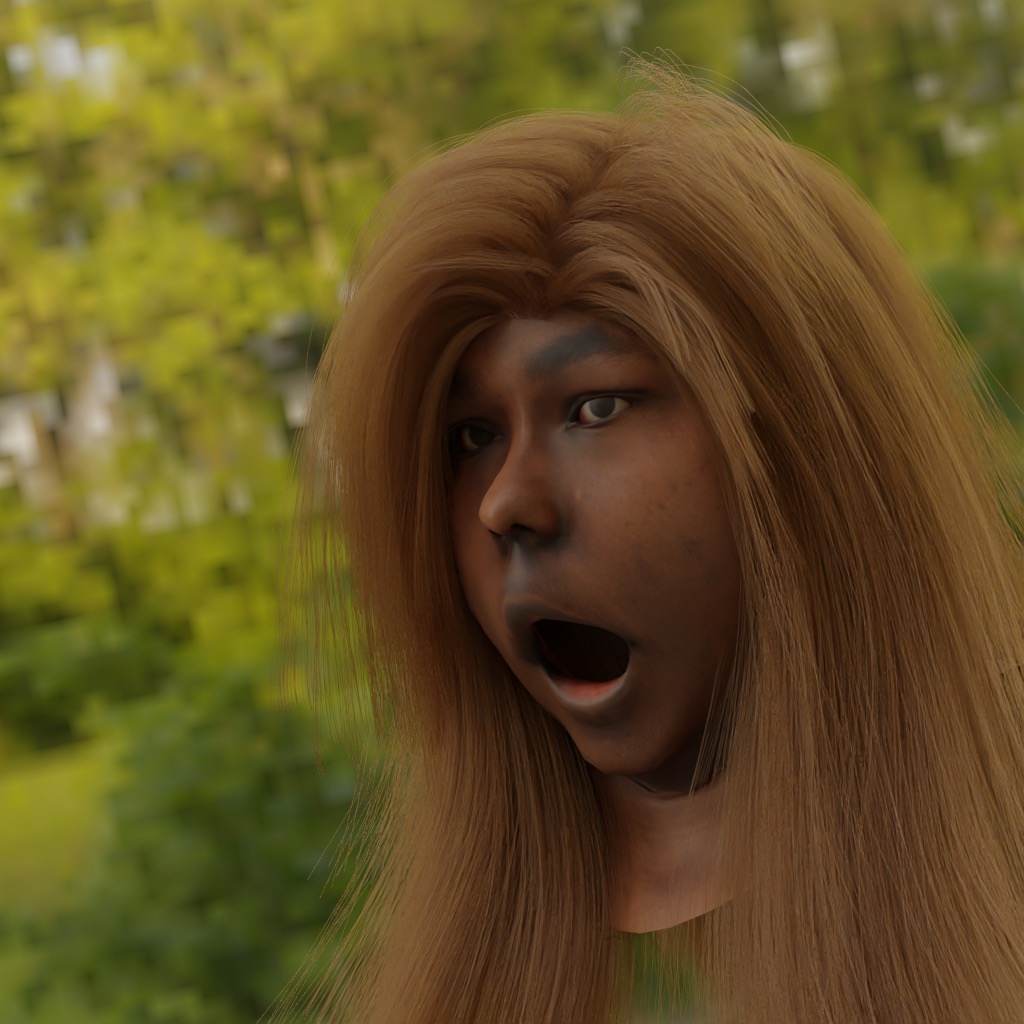}
    \includegraphics[width=0.23\columnwidth]{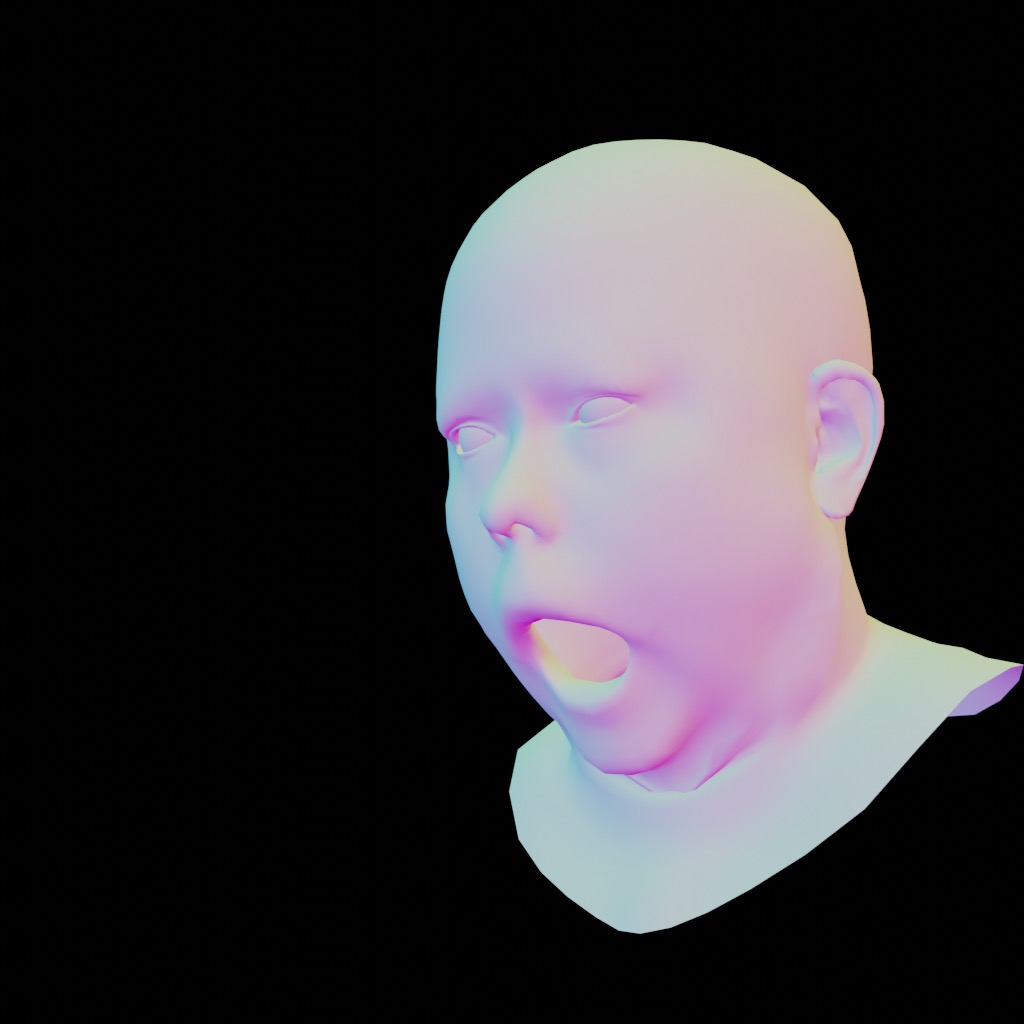}
    \includegraphics[width=0.23\columnwidth]{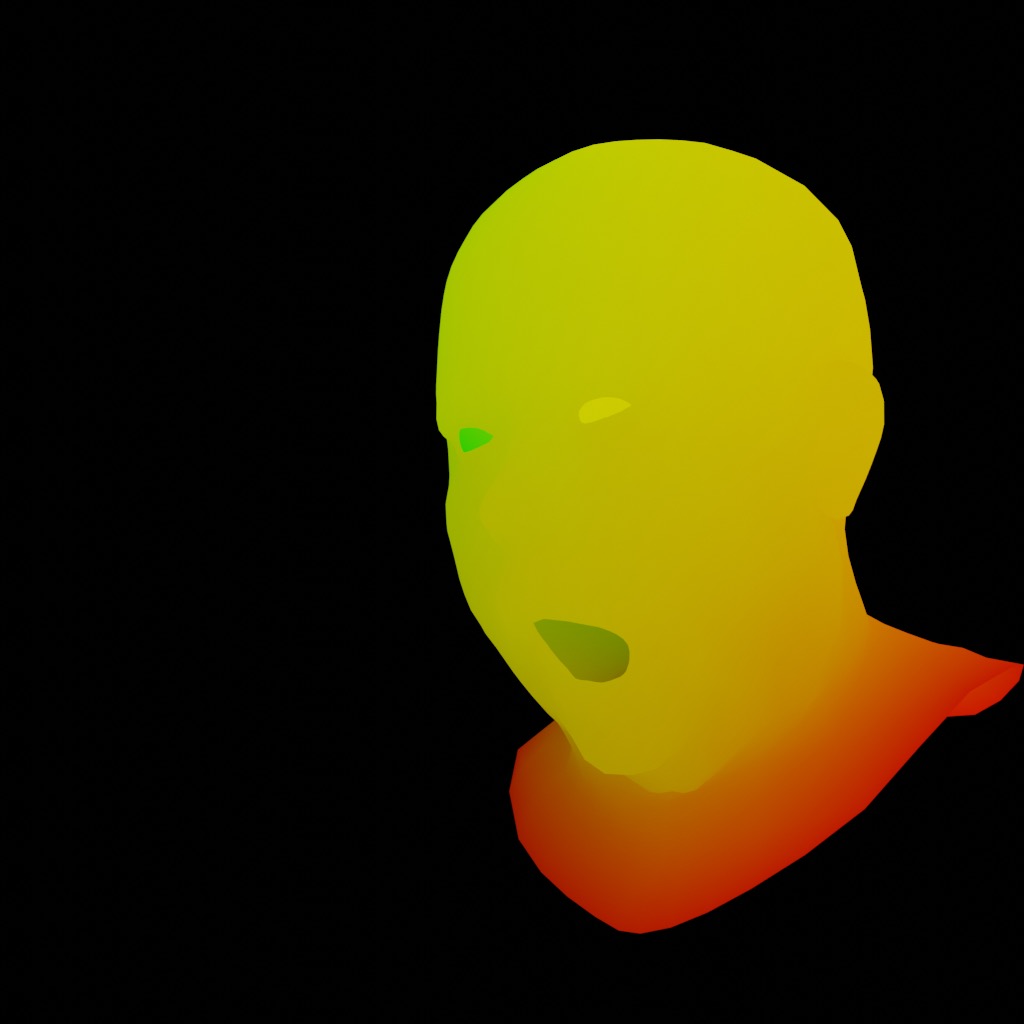}
    \includegraphics[width=0.23\columnwidth]{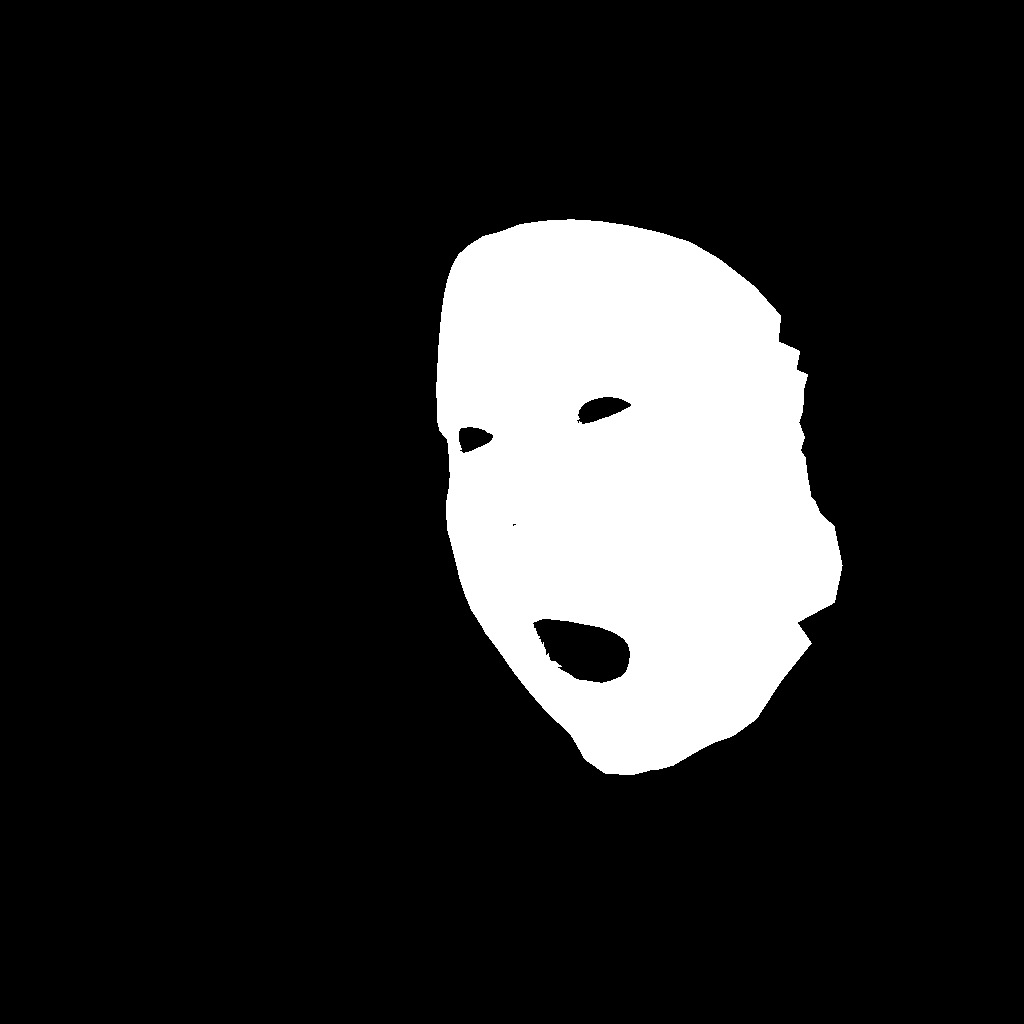}
    \includegraphics[width=0.23\columnwidth]{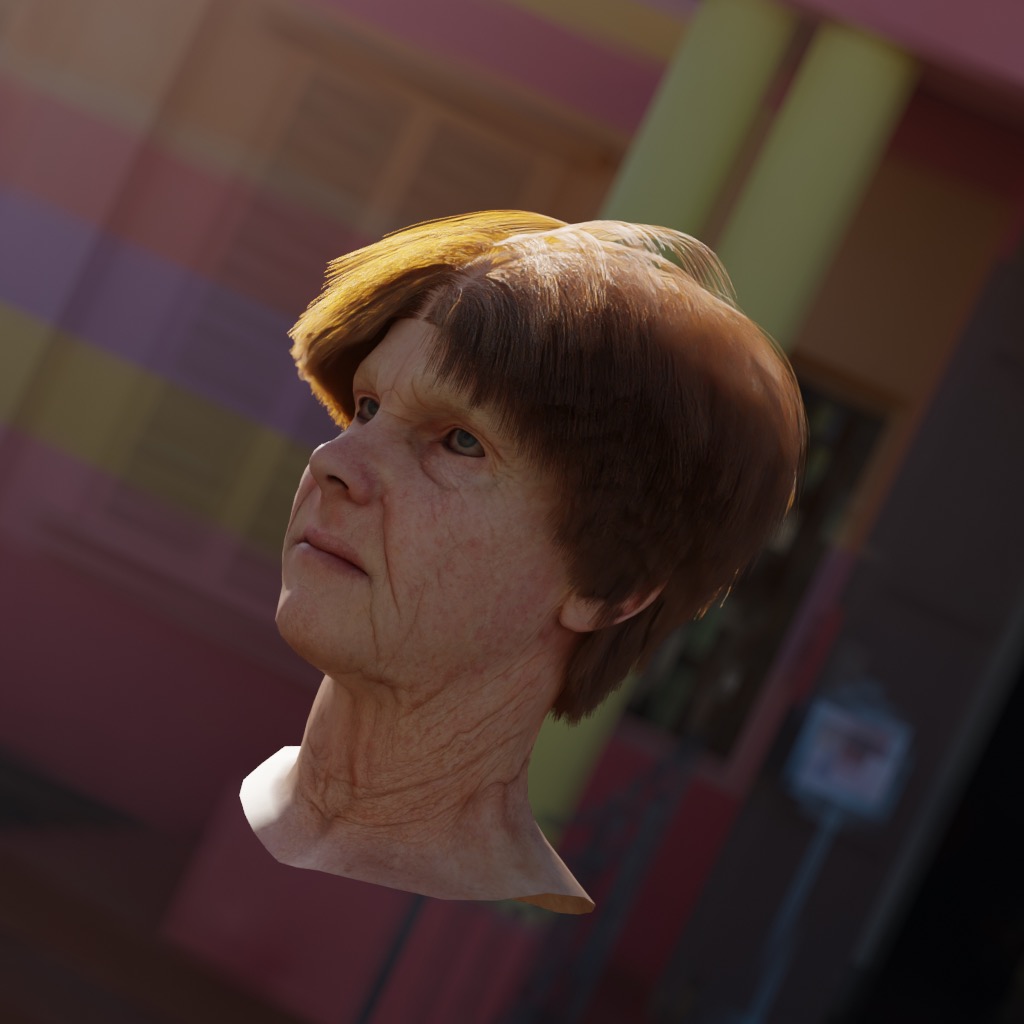}
    \includegraphics[width=0.23\columnwidth]{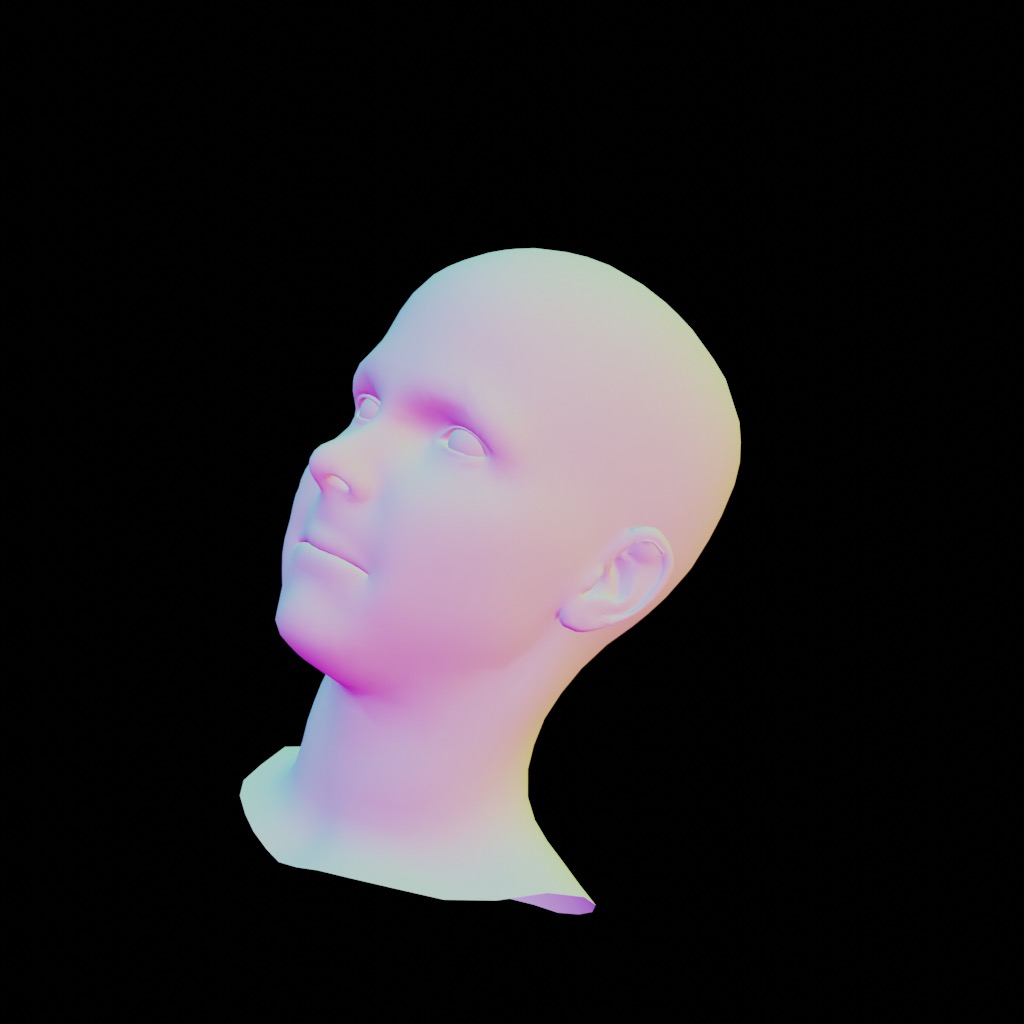}
    \includegraphics[width=0.23\columnwidth]{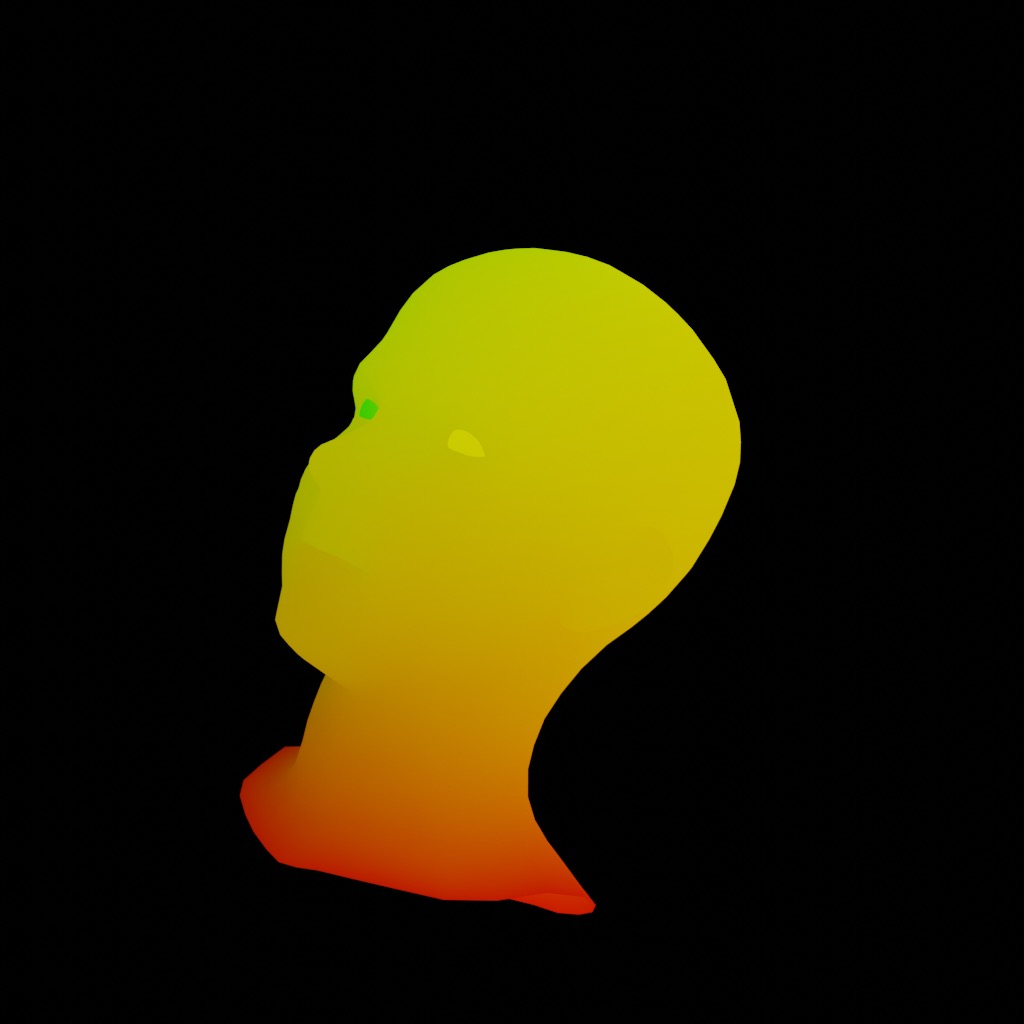}
    \includegraphics[width=0.23\columnwidth]{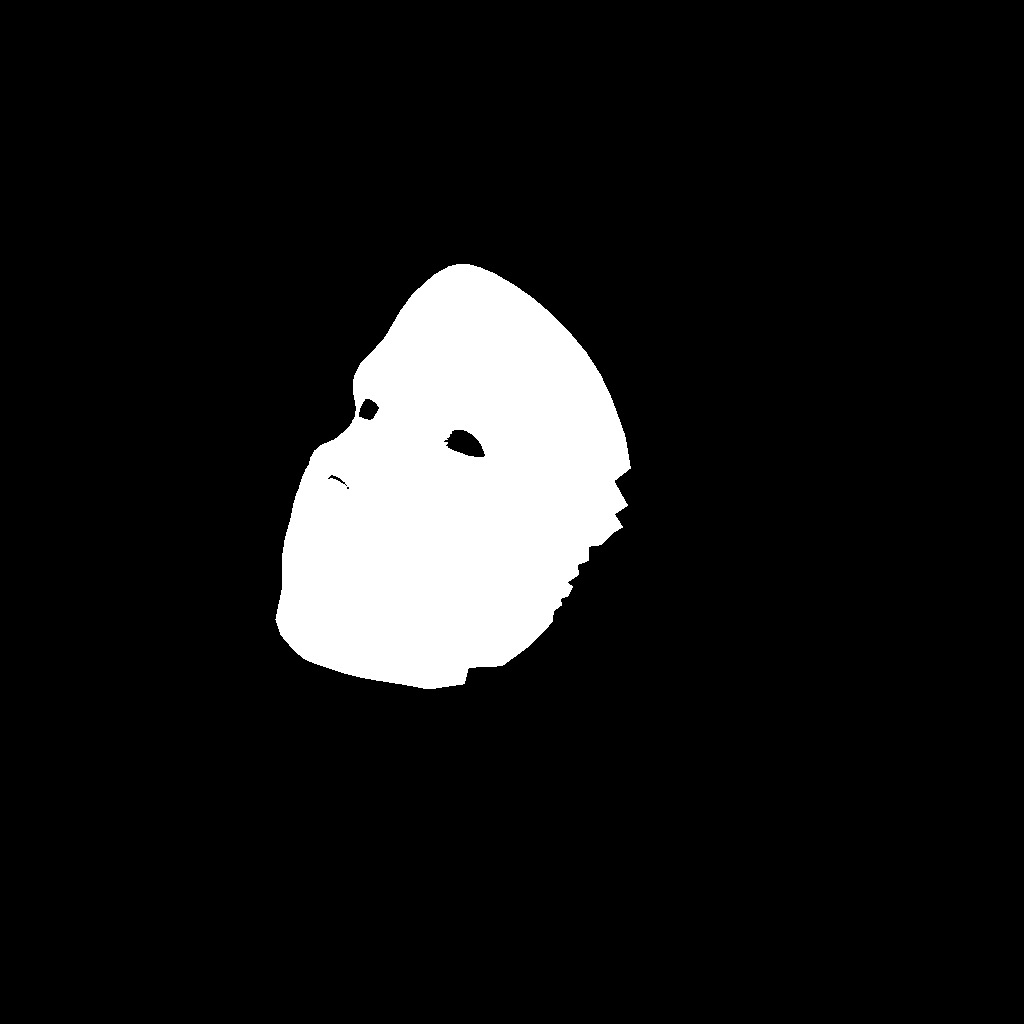}
    \includegraphics[width=0.23\columnwidth]{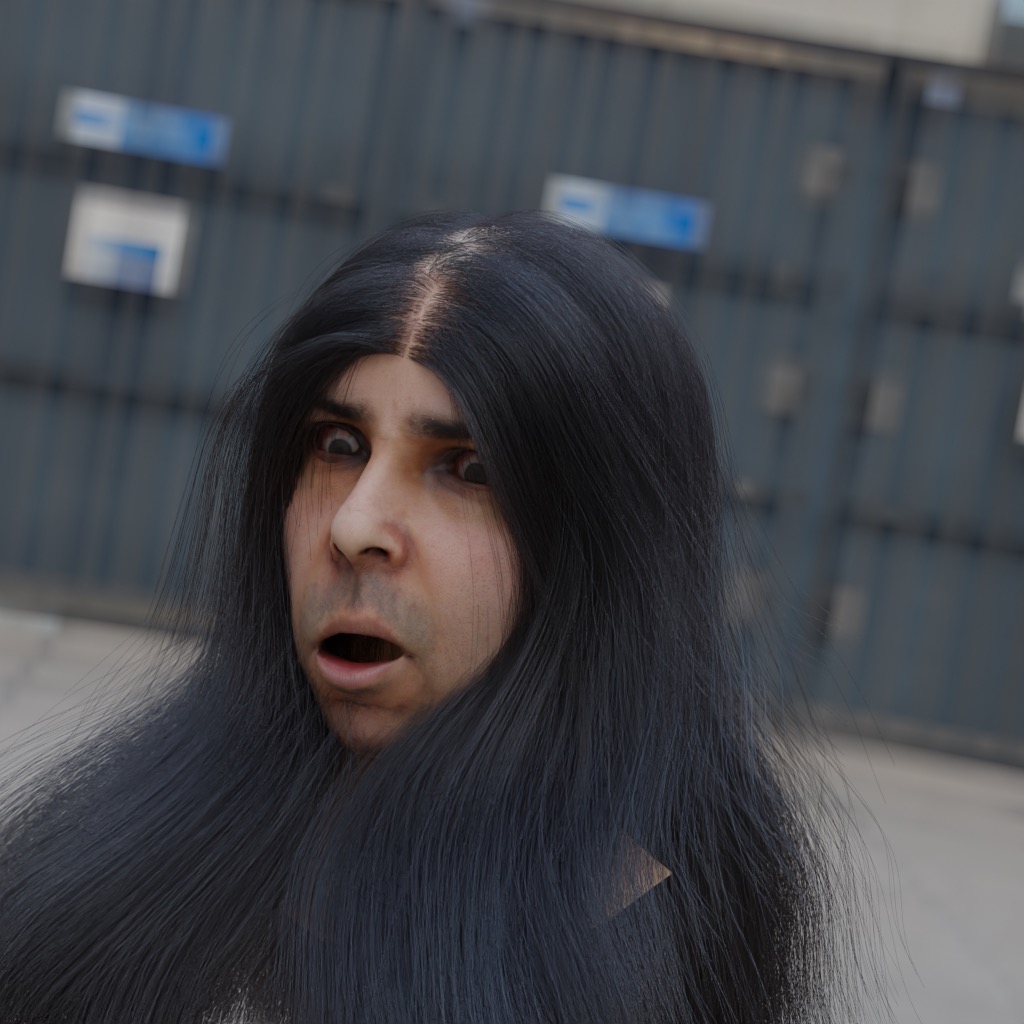}
    \includegraphics[width=0.23\columnwidth]{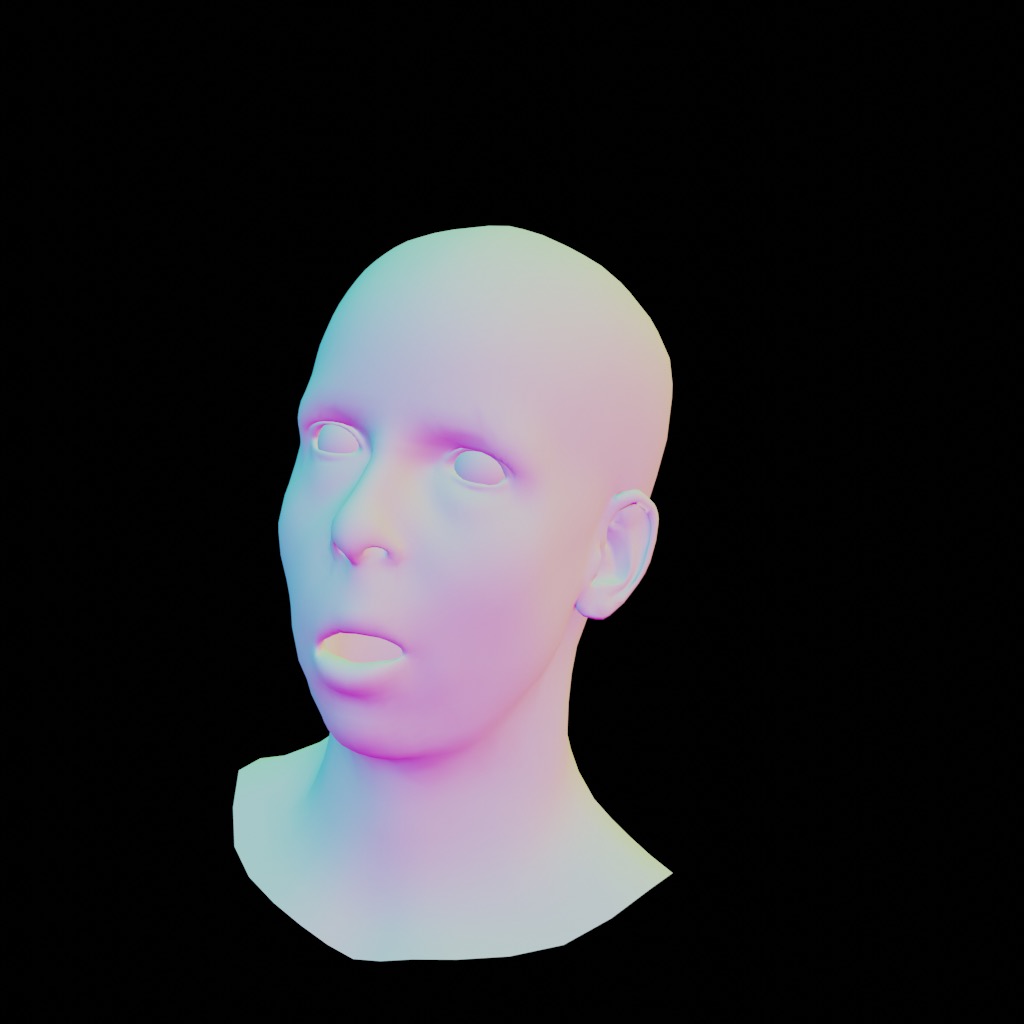}
    \includegraphics[width=0.23\columnwidth]{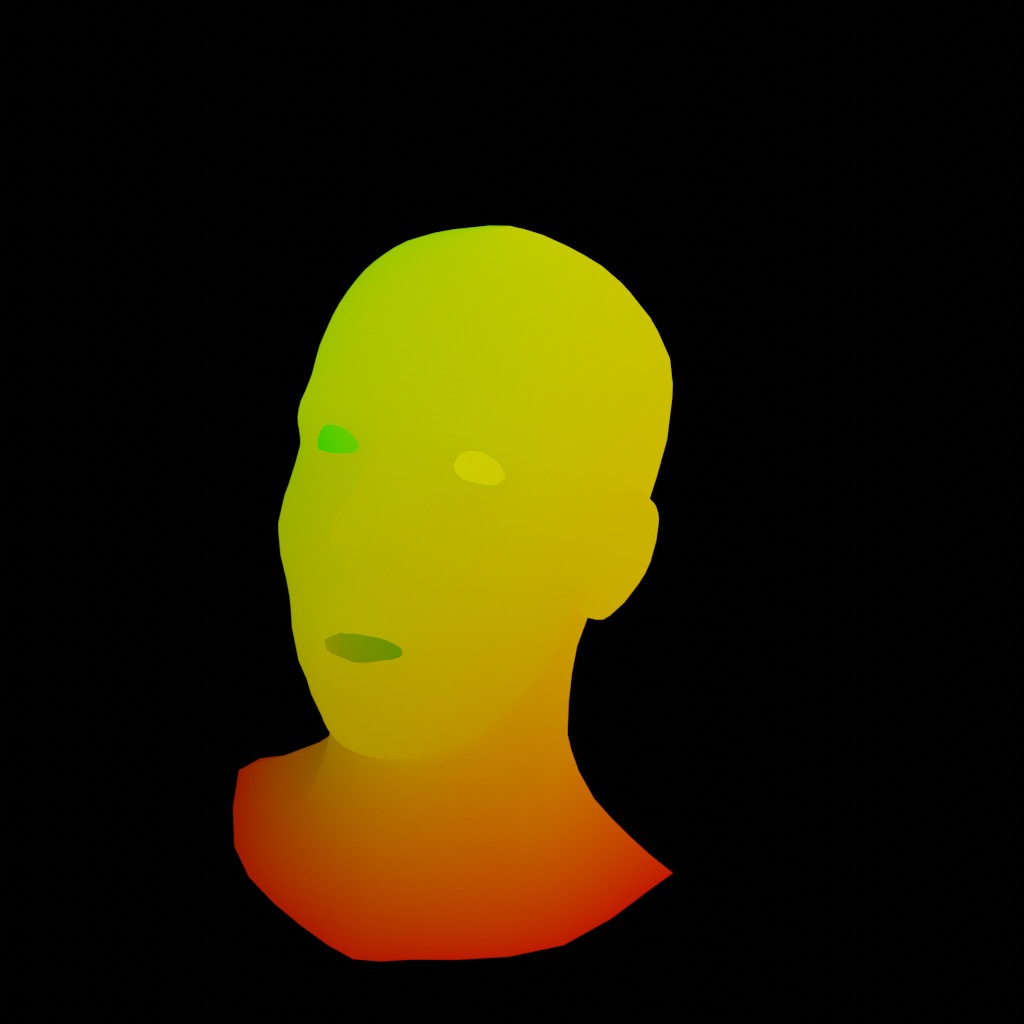}
    \includegraphics[width=0.23\columnwidth]{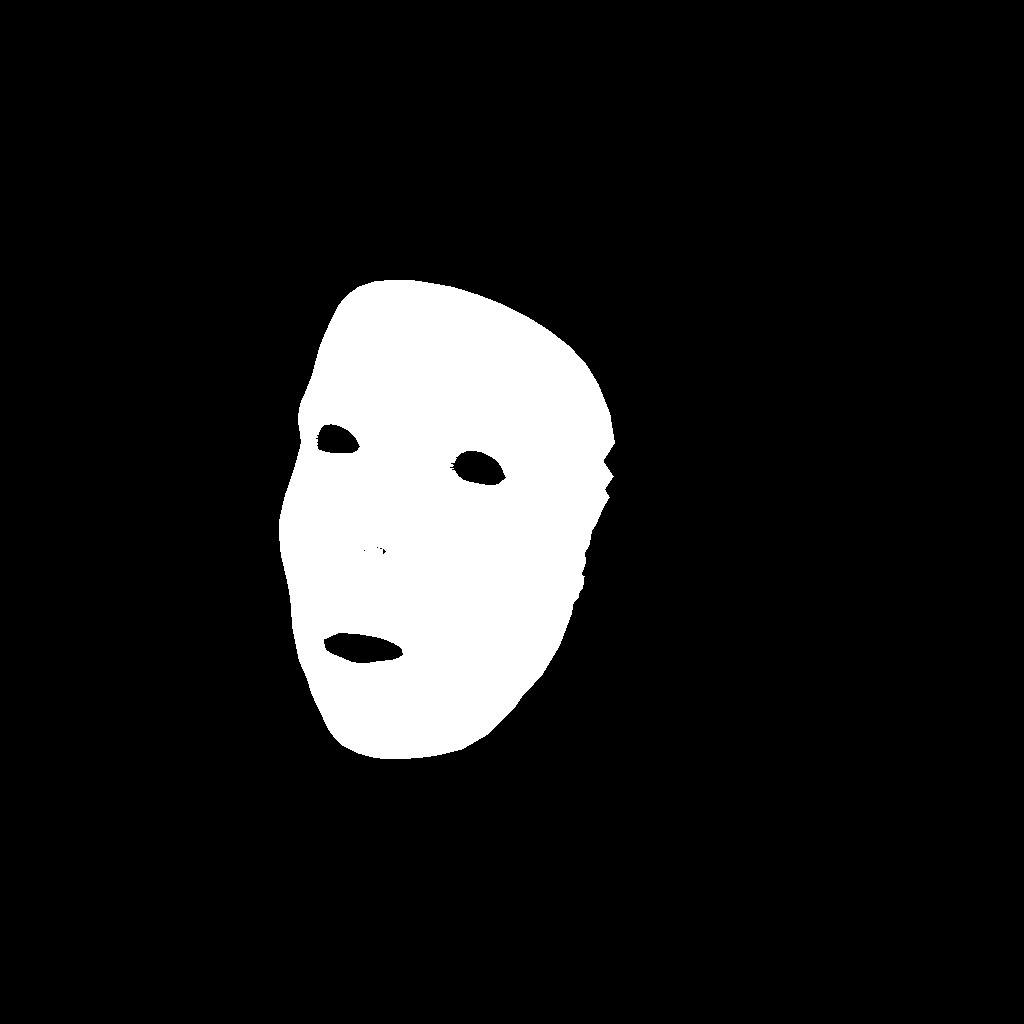}
    \includegraphics[width=0.23\columnwidth]{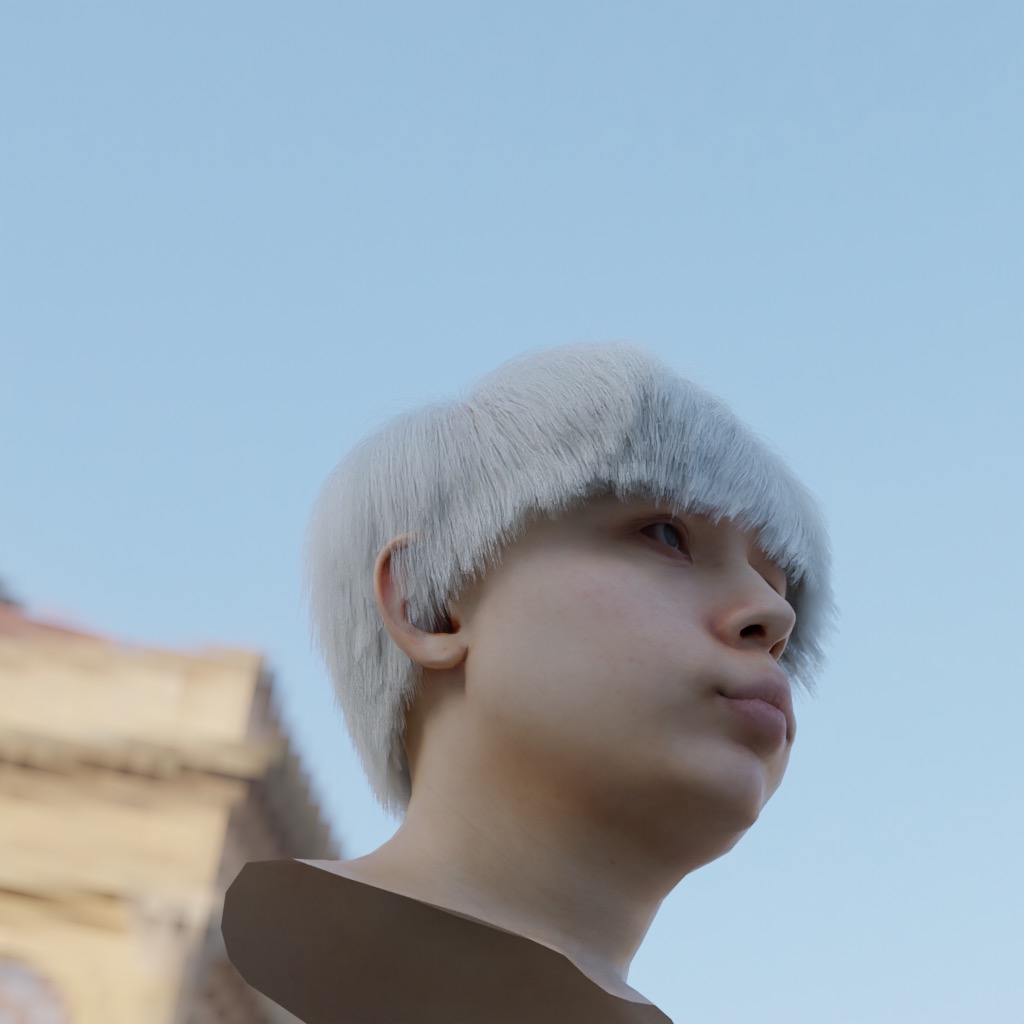}
    \includegraphics[width=0.23\columnwidth]{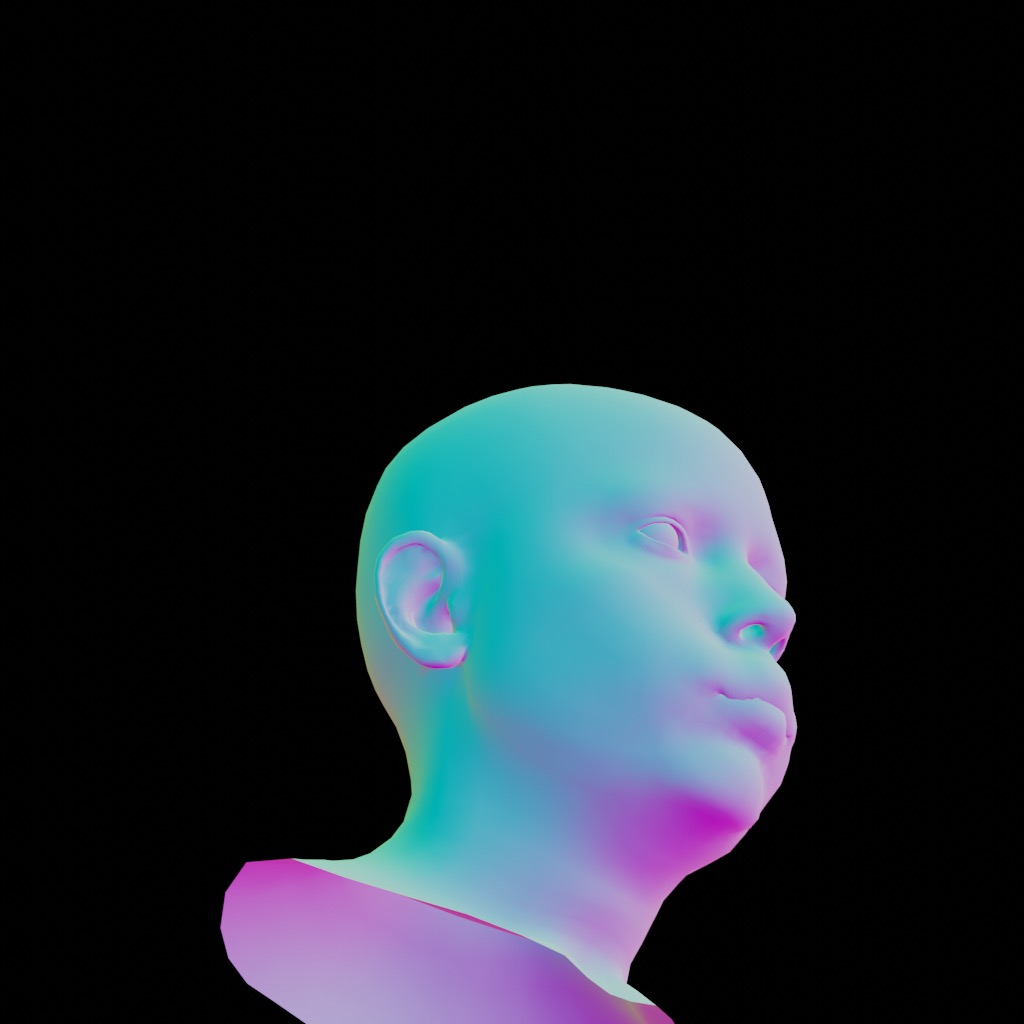}
    \includegraphics[width=0.23\columnwidth]{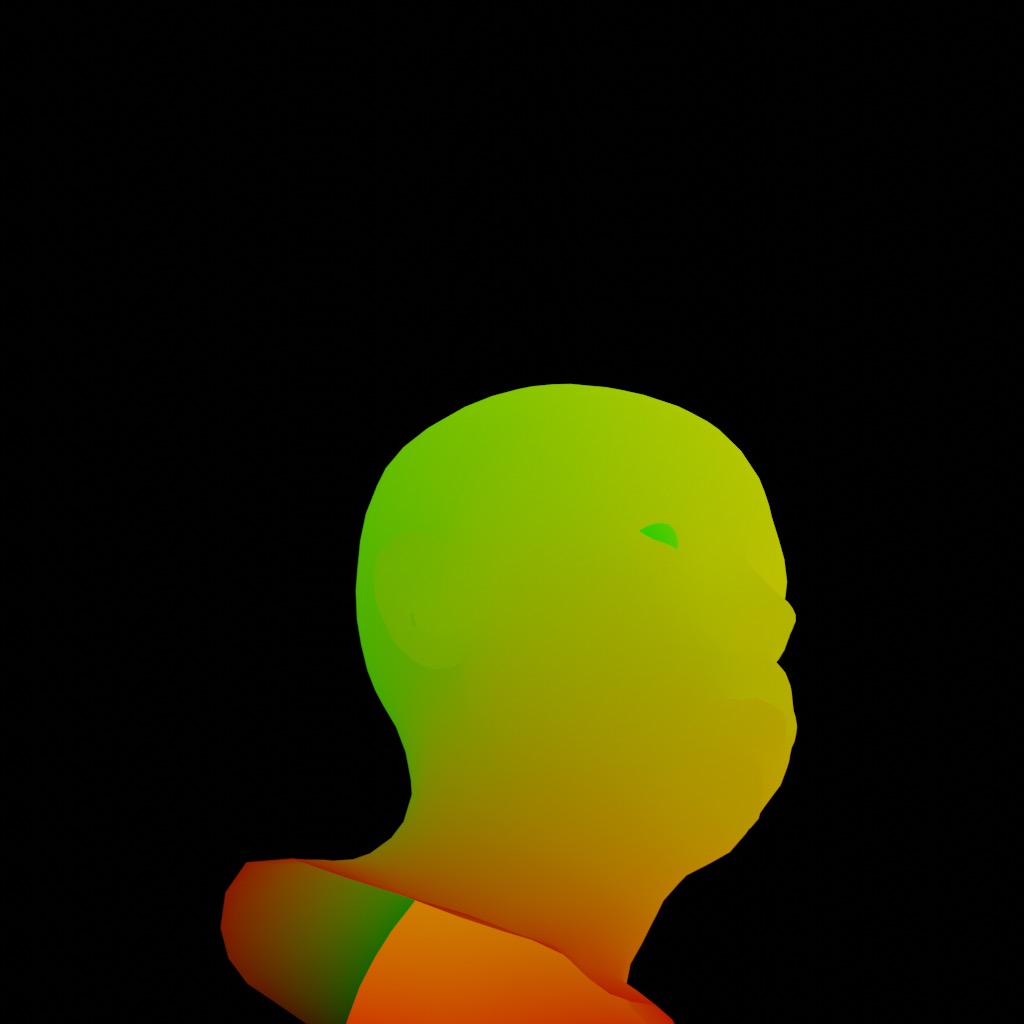}
    \includegraphics[width=0.23\columnwidth]{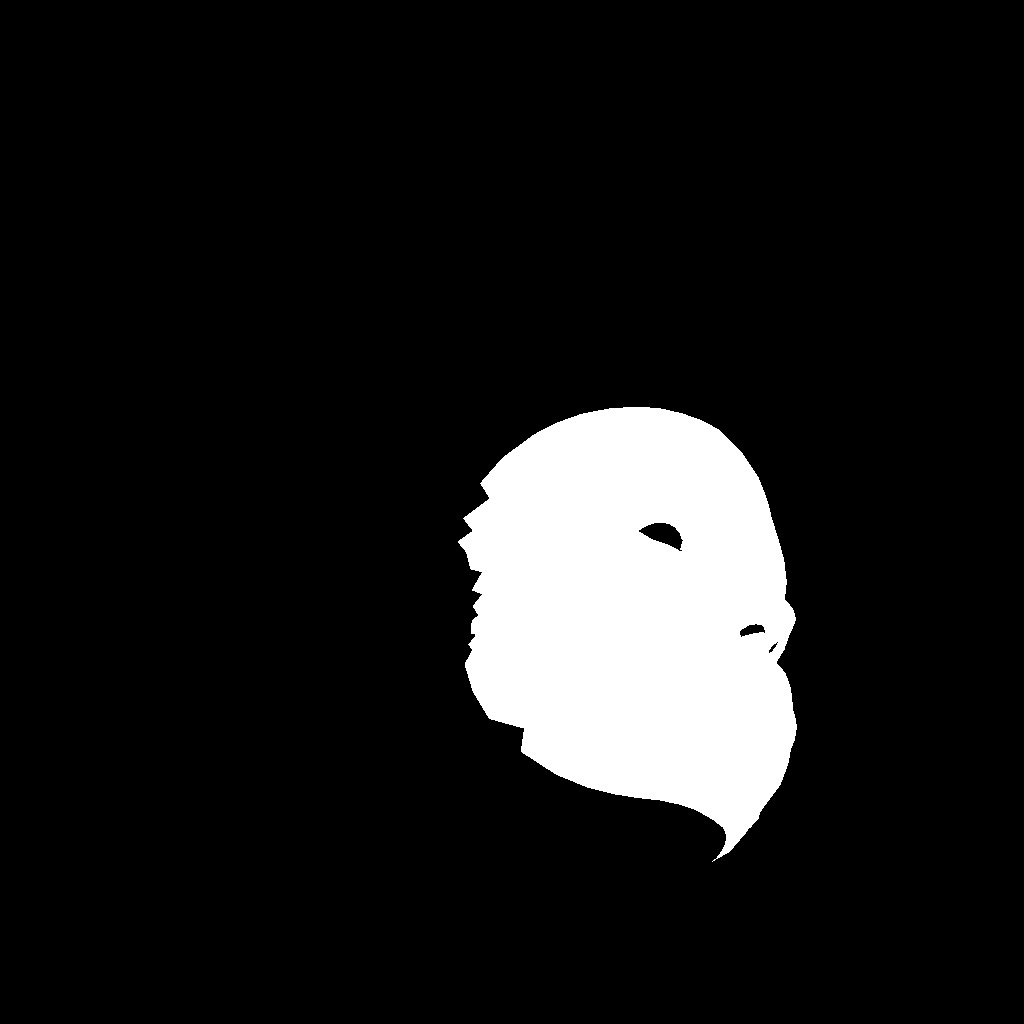}
    \includegraphics[width=0.23\columnwidth]{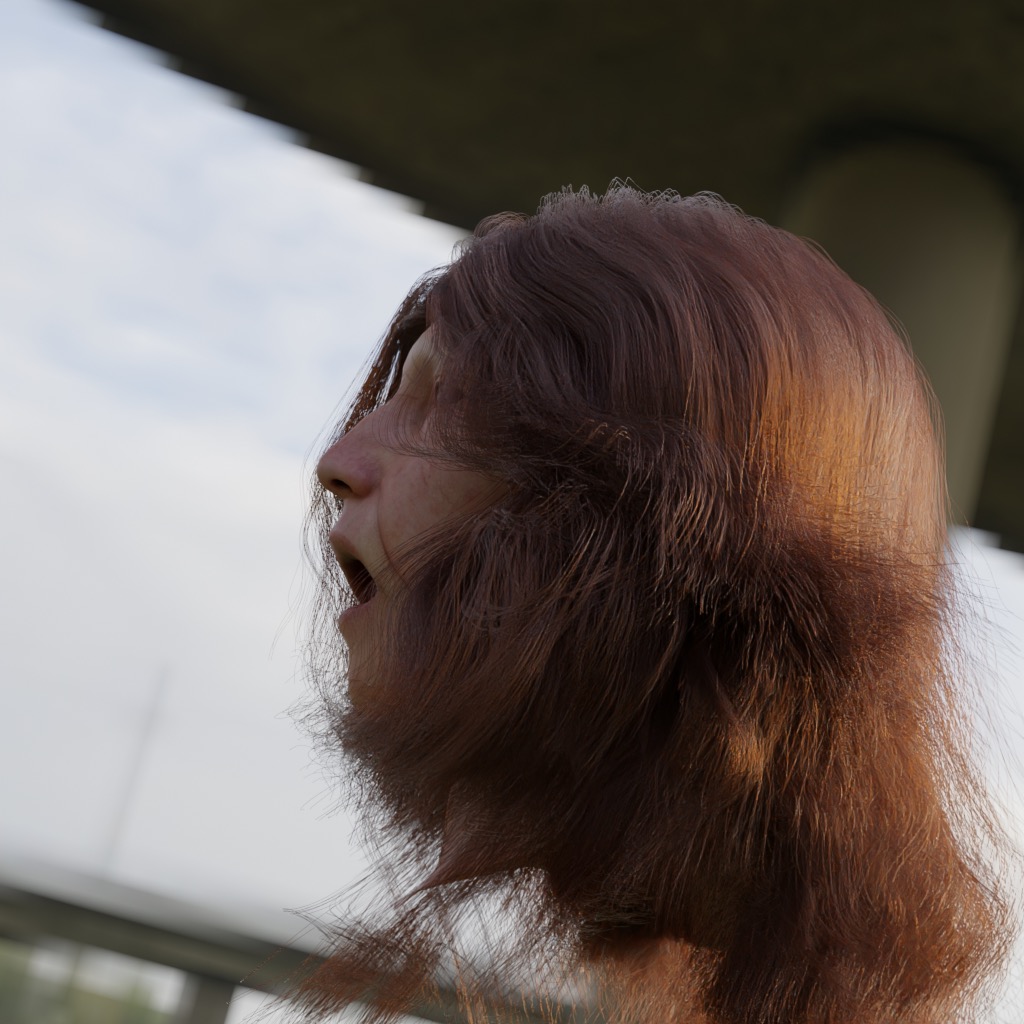}
    \includegraphics[width=0.23\columnwidth]{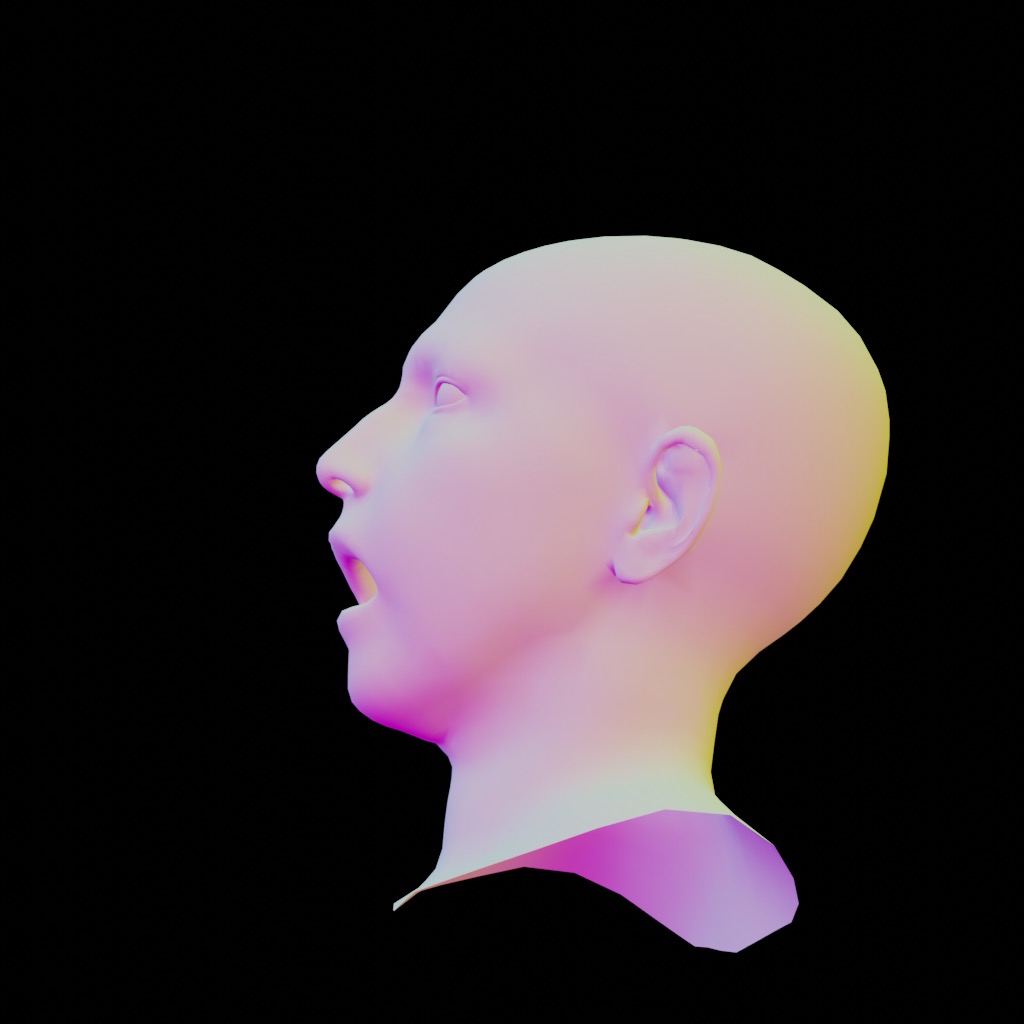}
    \includegraphics[width=0.23\columnwidth]{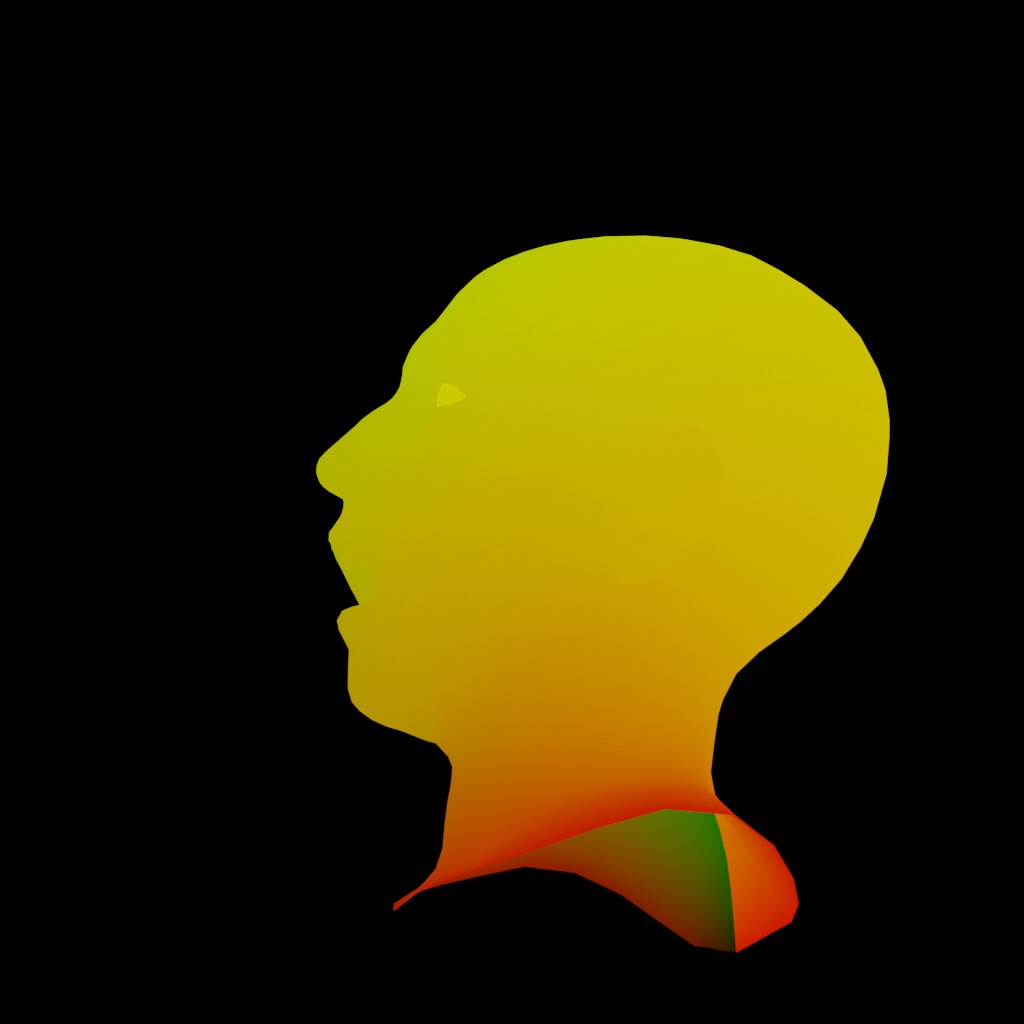}
    \includegraphics[width=0.23\columnwidth]{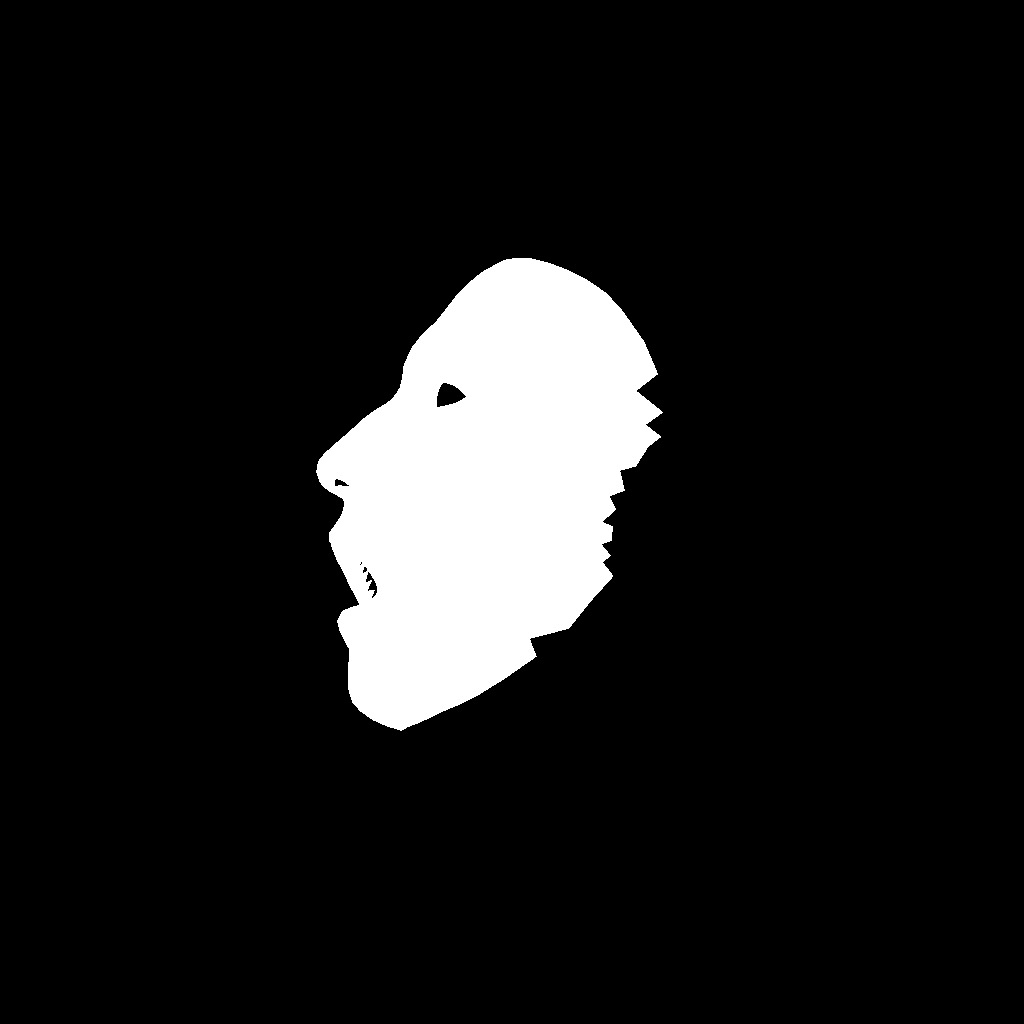}
    \includegraphics[width=0.23\columnwidth]{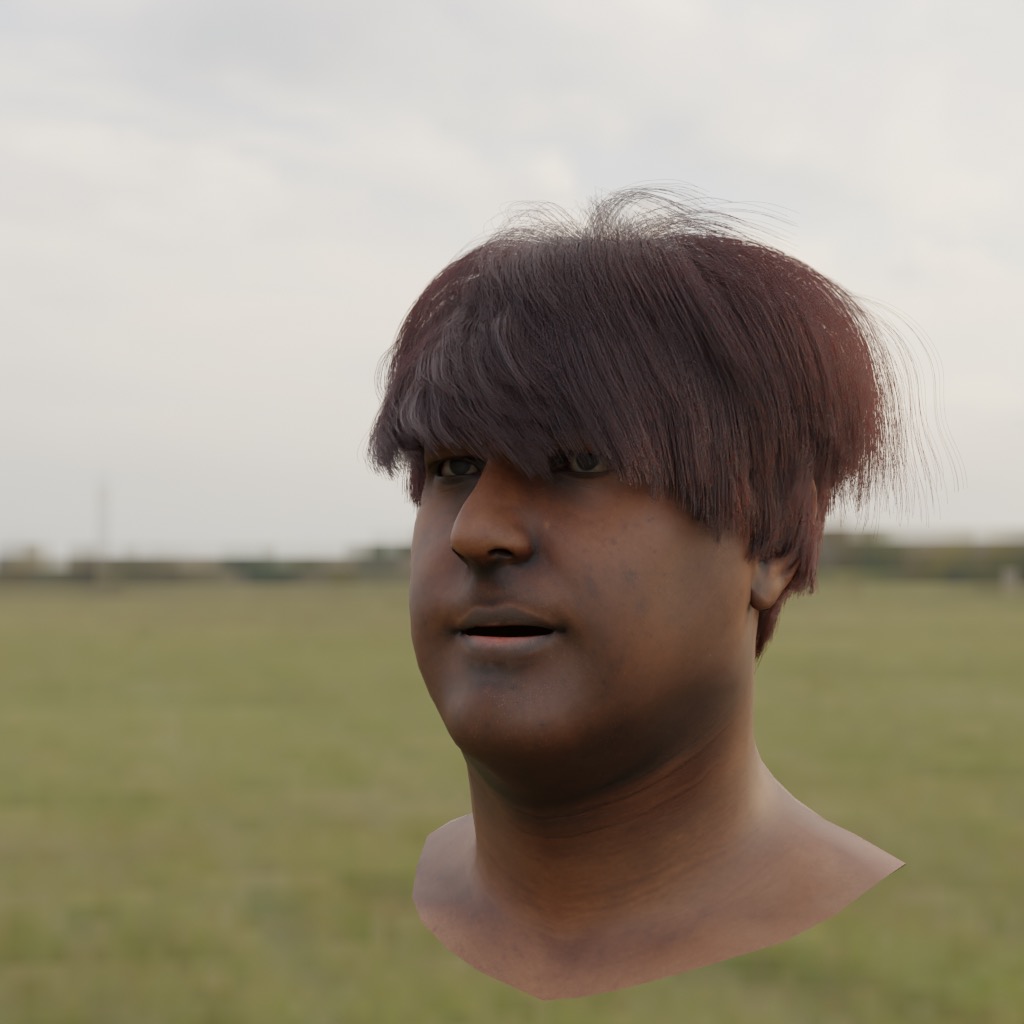}
    \includegraphics[width=0.23\columnwidth]{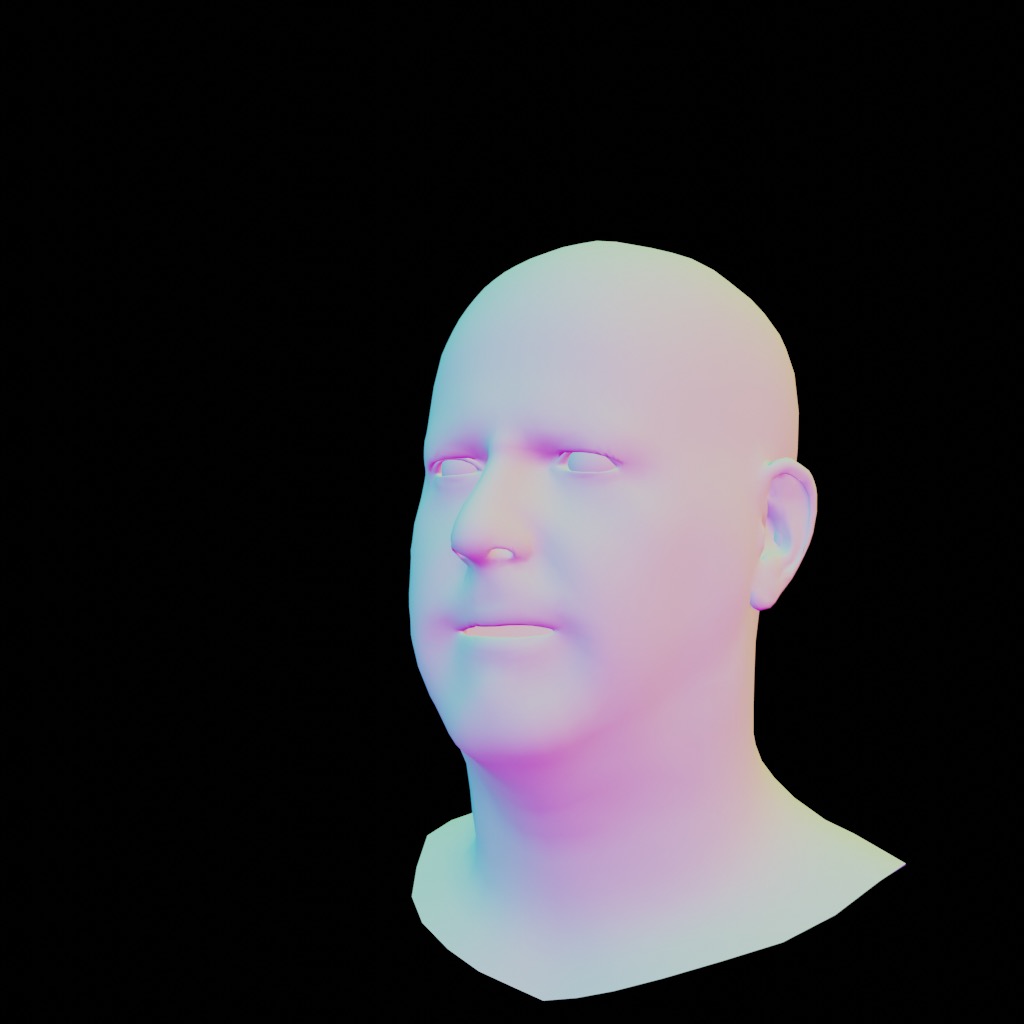}
    \includegraphics[width=0.23\columnwidth]{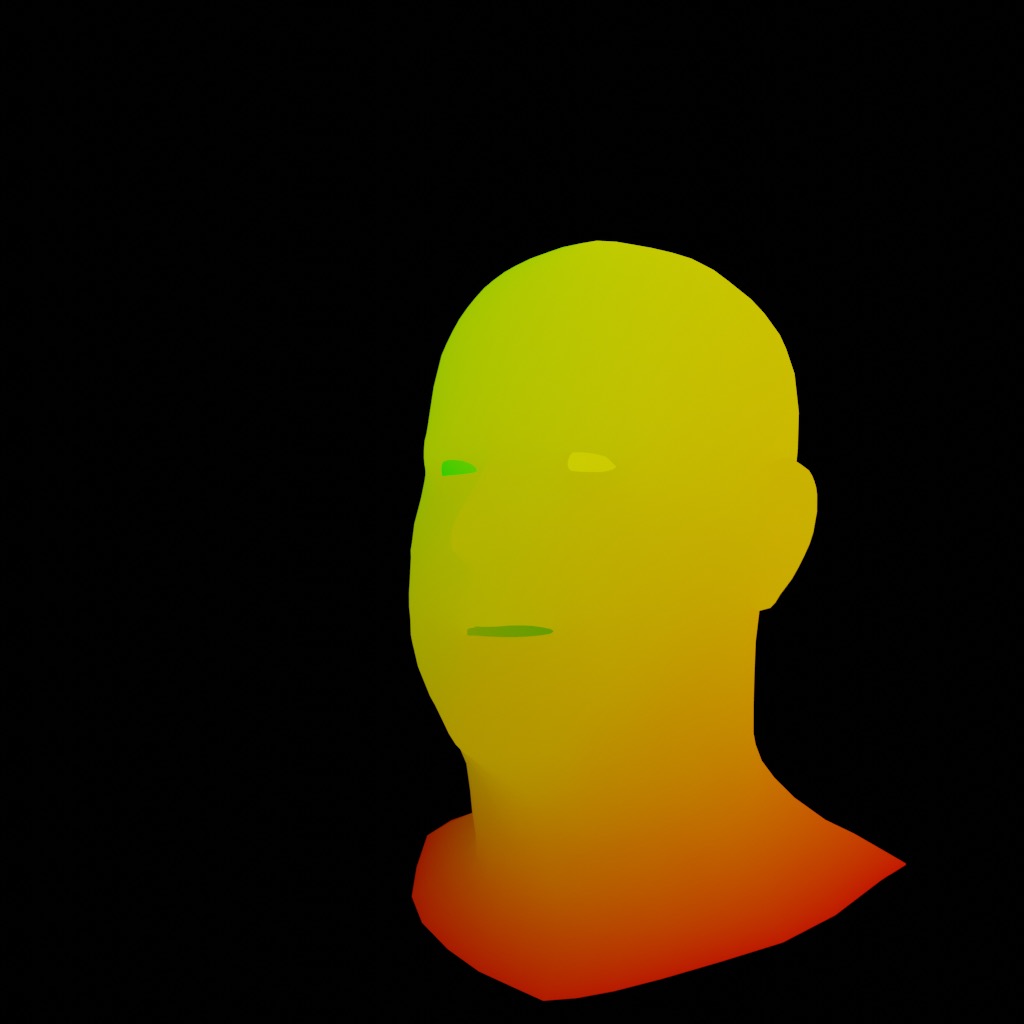}
    \includegraphics[width=0.23\columnwidth]{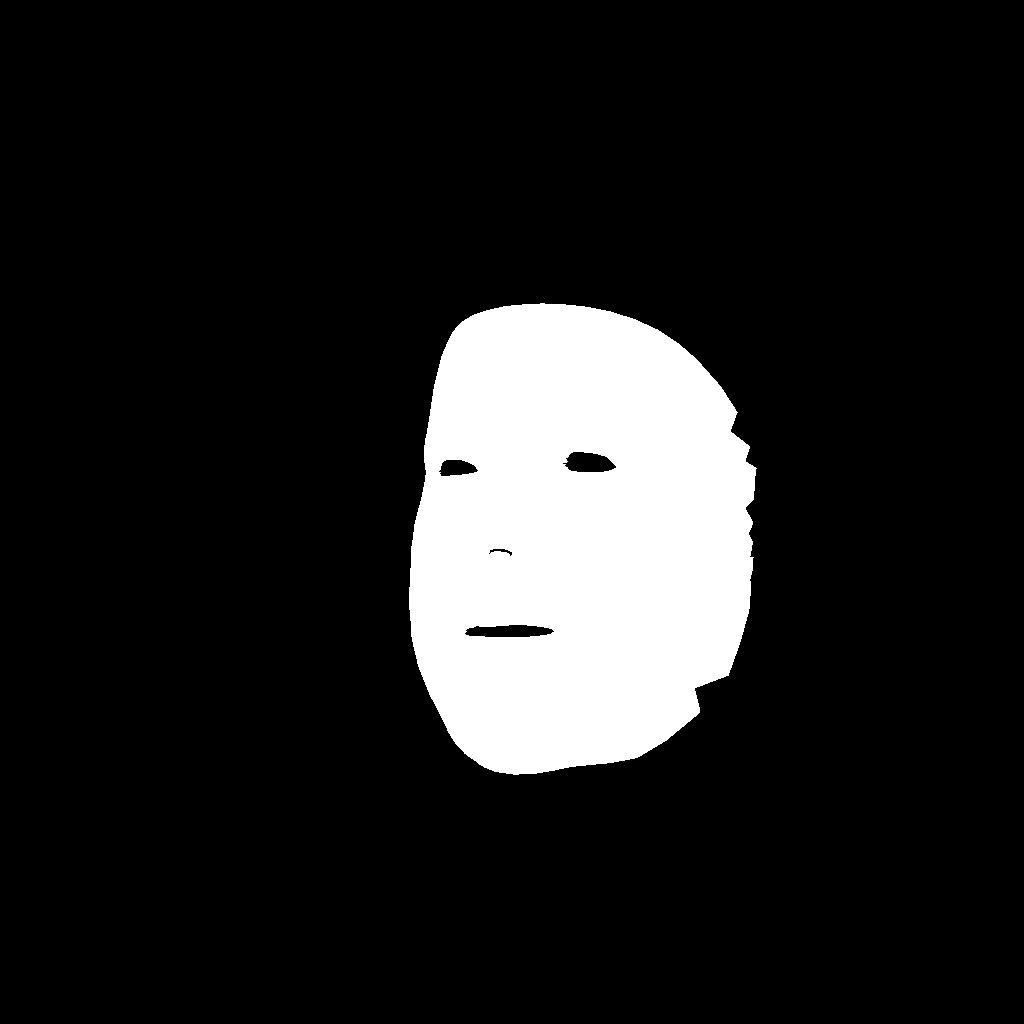}
    \includegraphics[width=0.23\columnwidth]{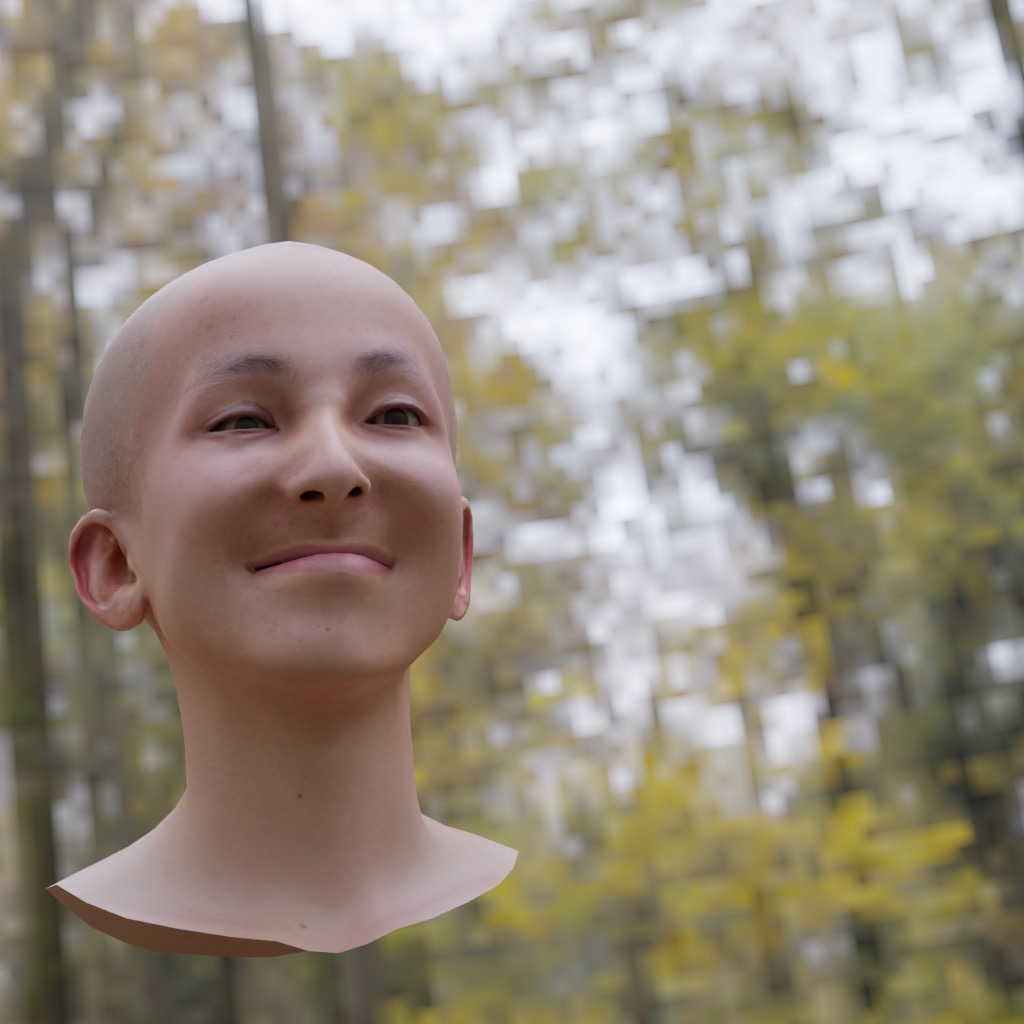}
    \includegraphics[width=0.23\columnwidth]{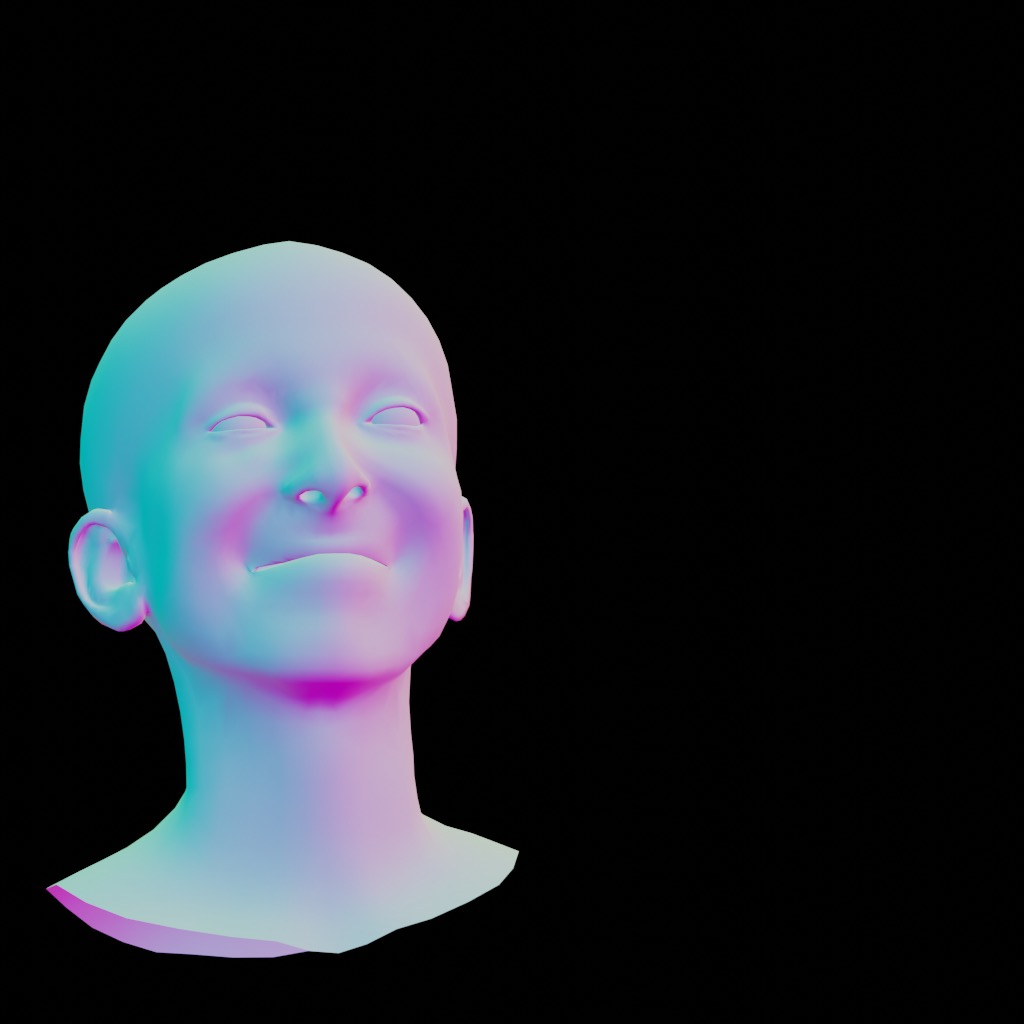}
    \includegraphics[width=0.23\columnwidth]{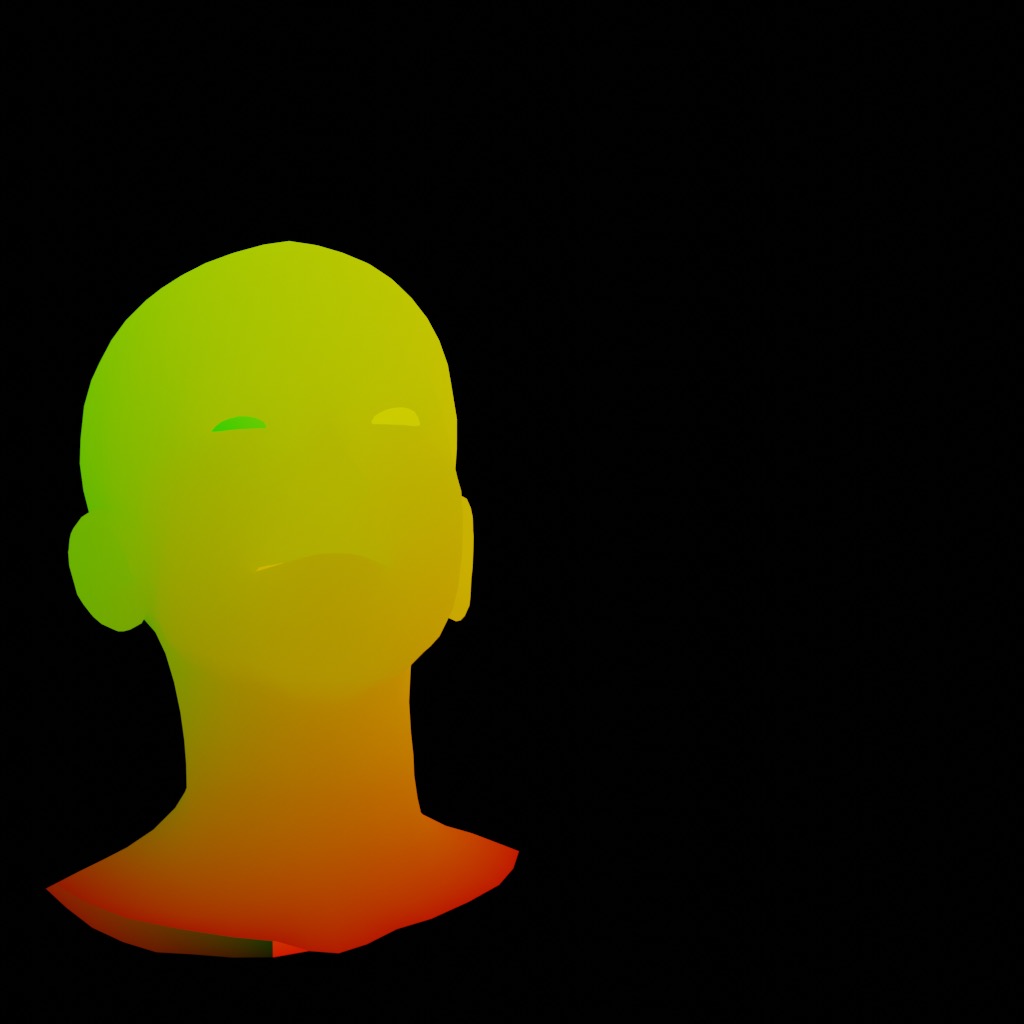}
    \includegraphics[width=0.23\columnwidth]{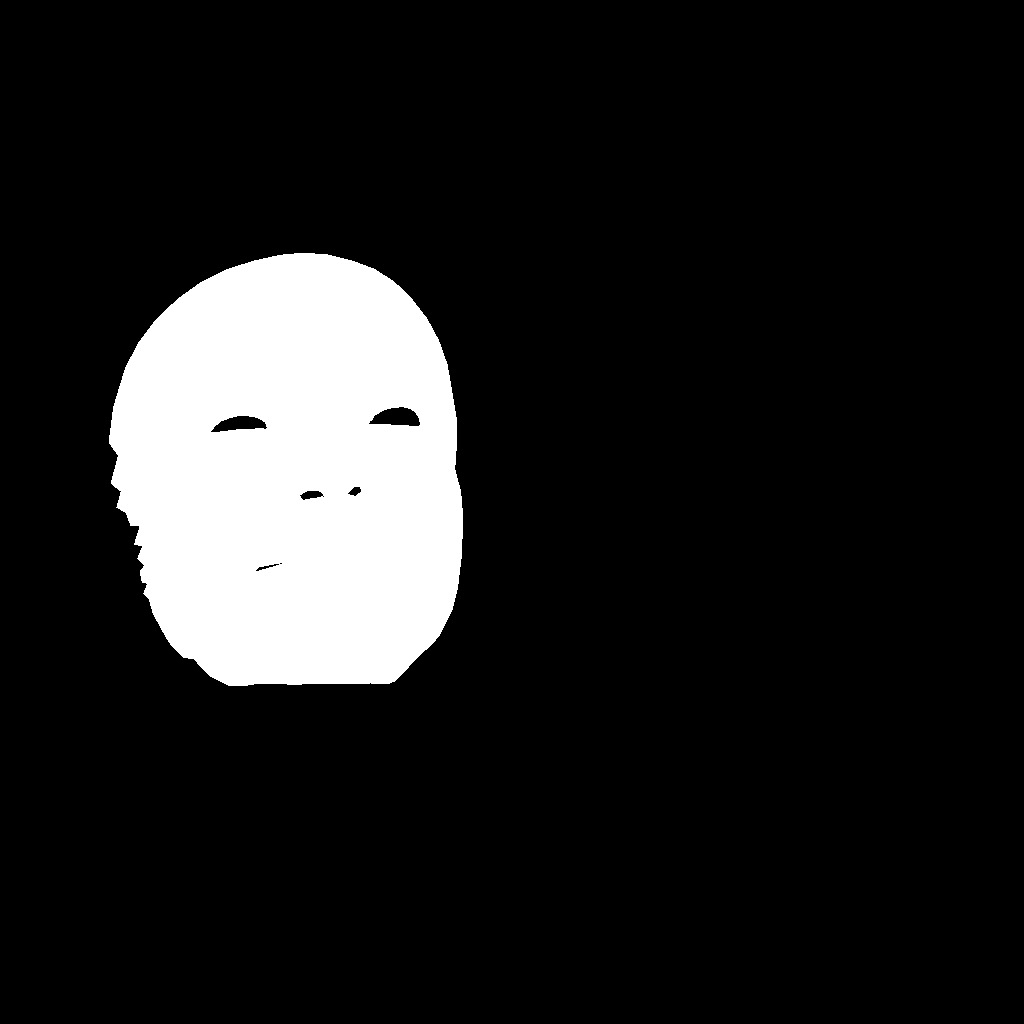}
    \includegraphics[width=0.23\columnwidth]{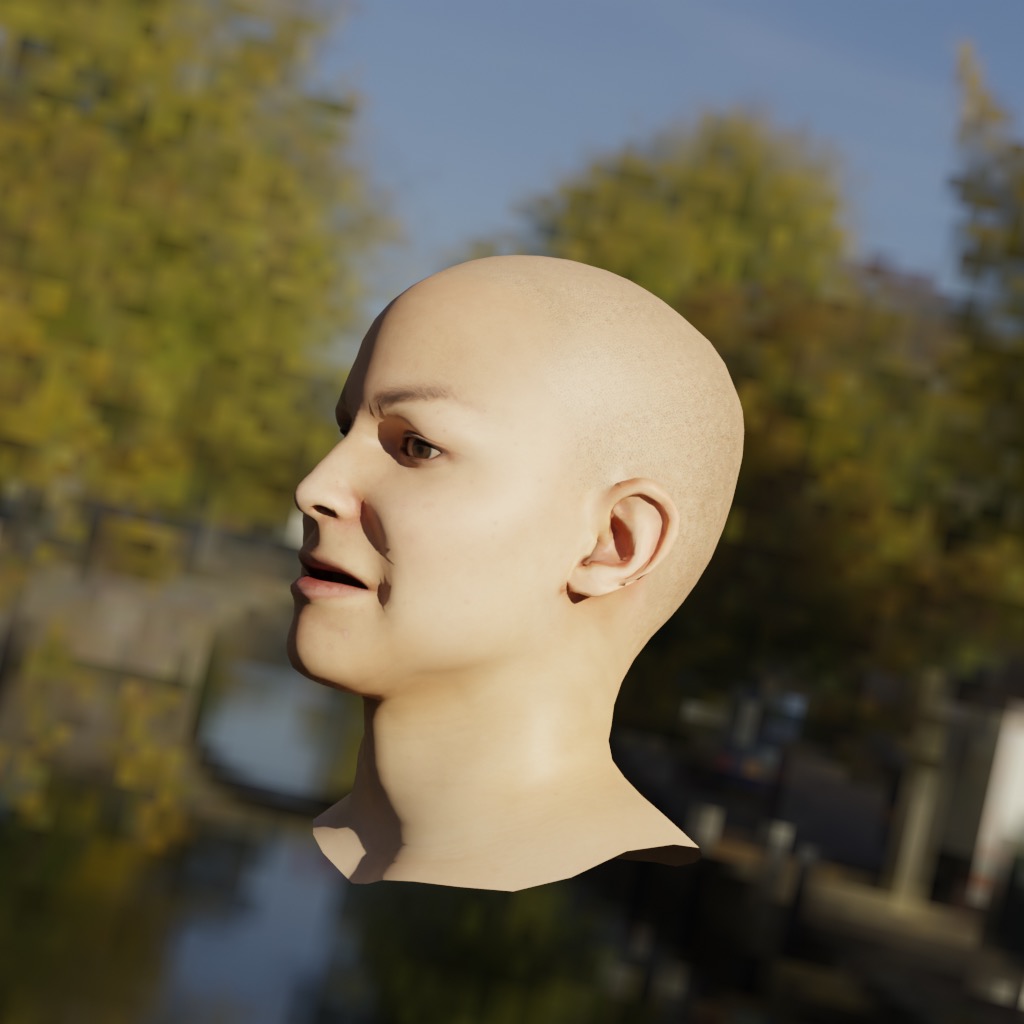}
    \includegraphics[width=0.23\columnwidth]{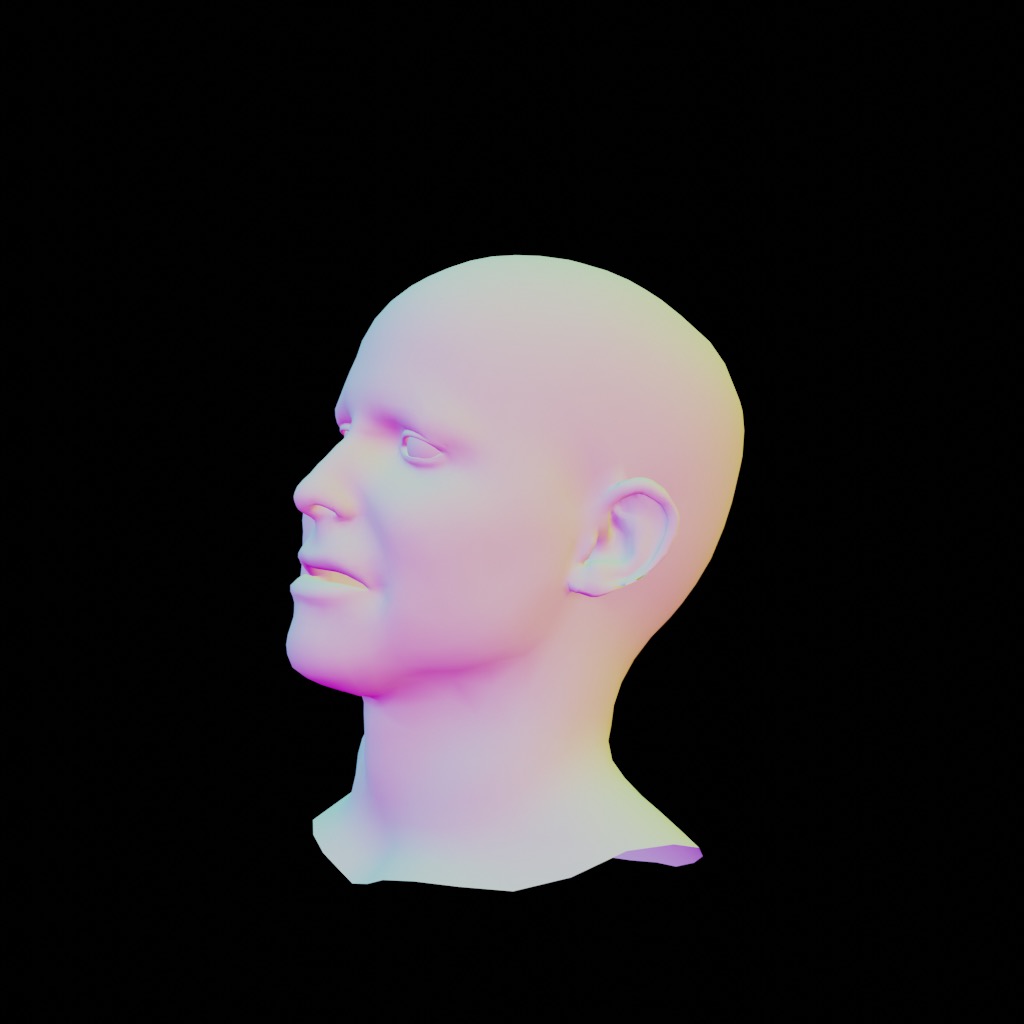}
    \includegraphics[width=0.23\columnwidth]{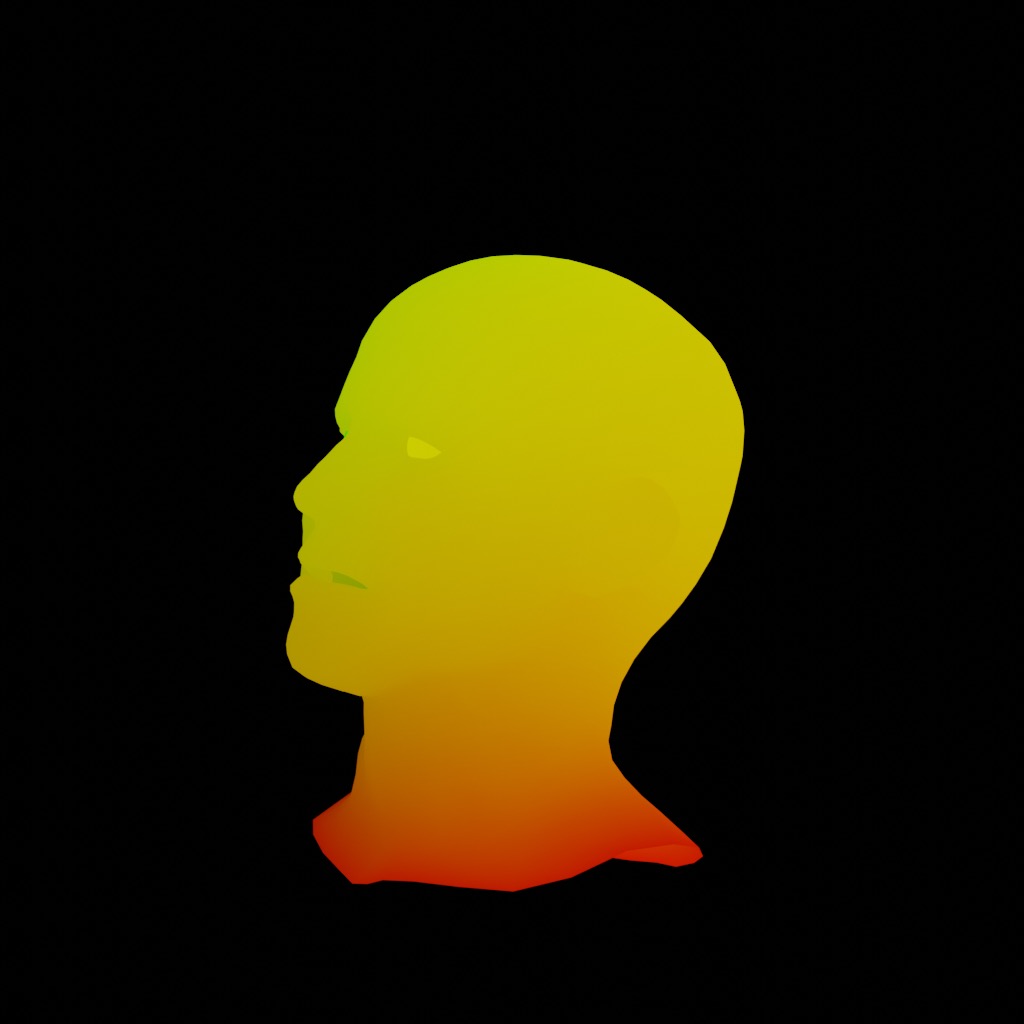}
    \includegraphics[width=0.23\columnwidth]{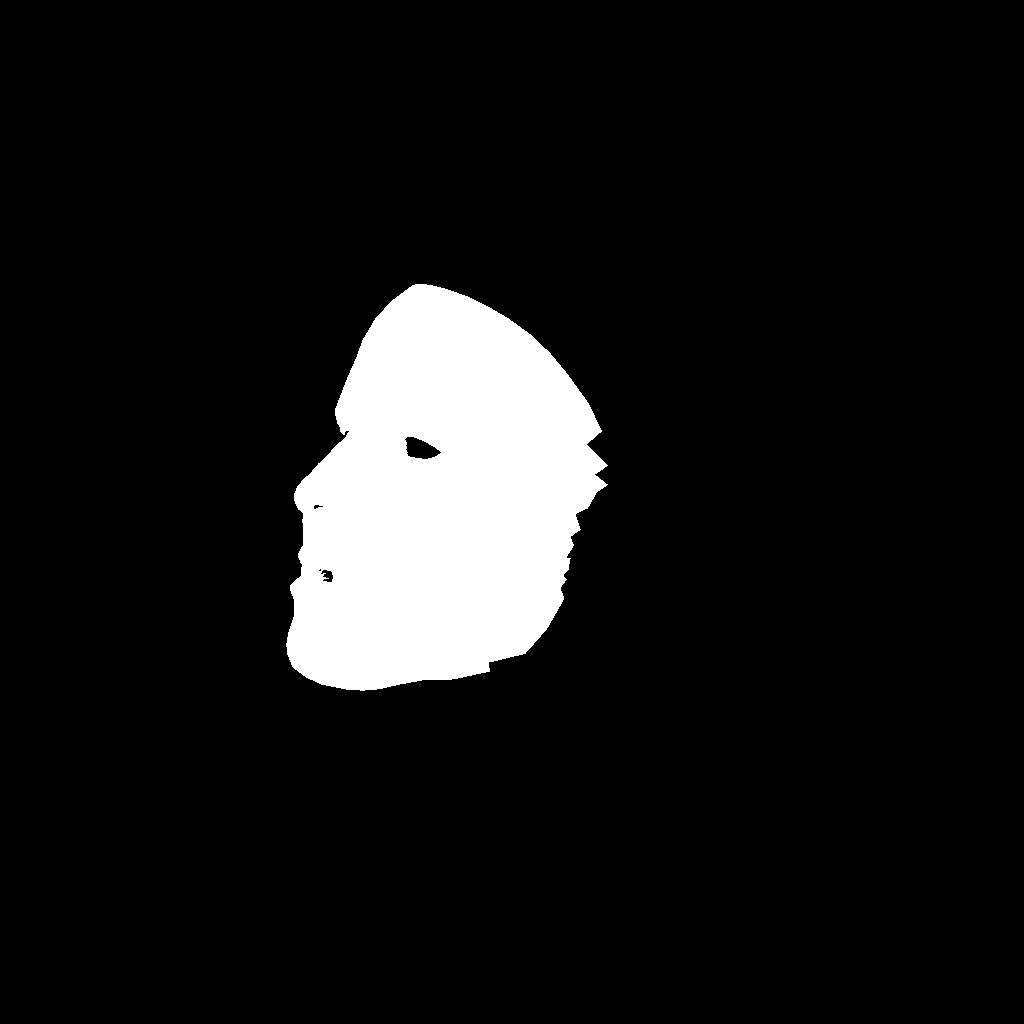}
    \includegraphics[width=0.23\columnwidth]{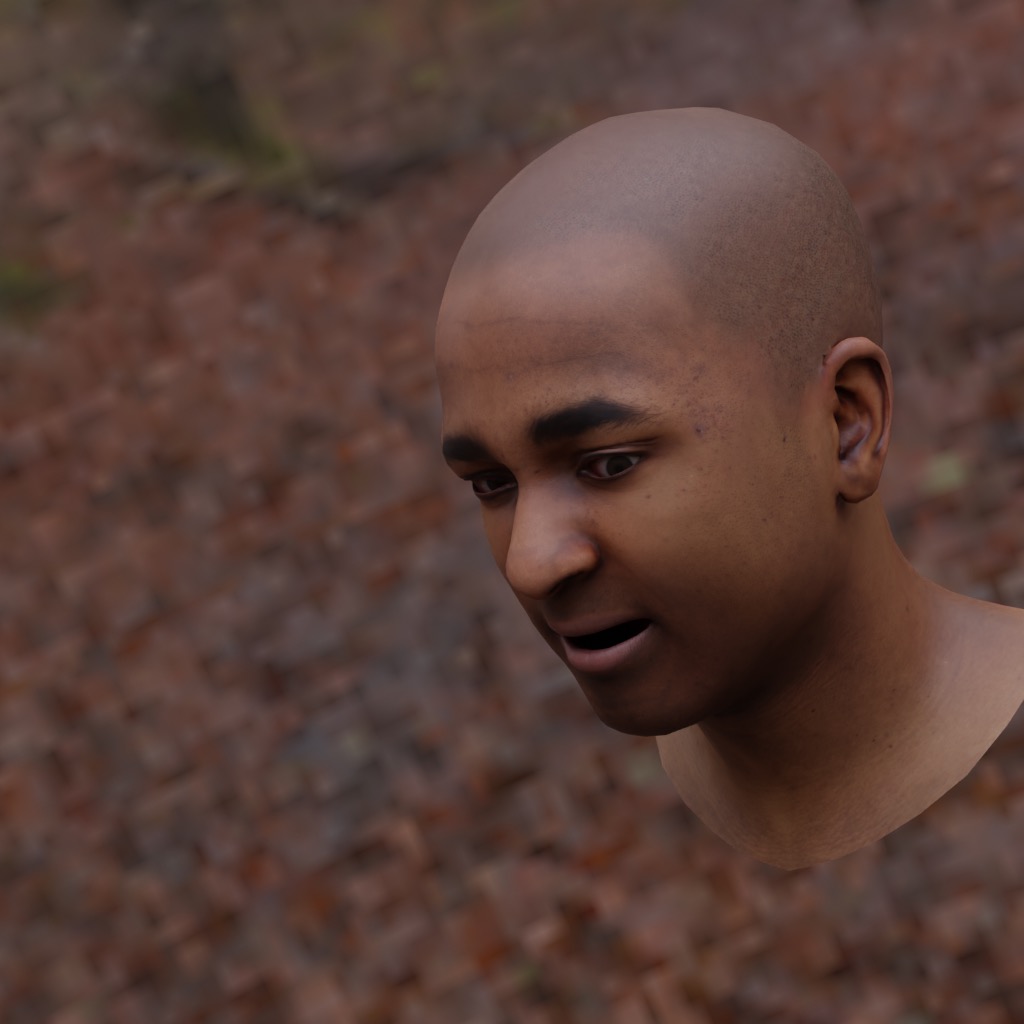}
    \includegraphics[width=0.23\columnwidth]{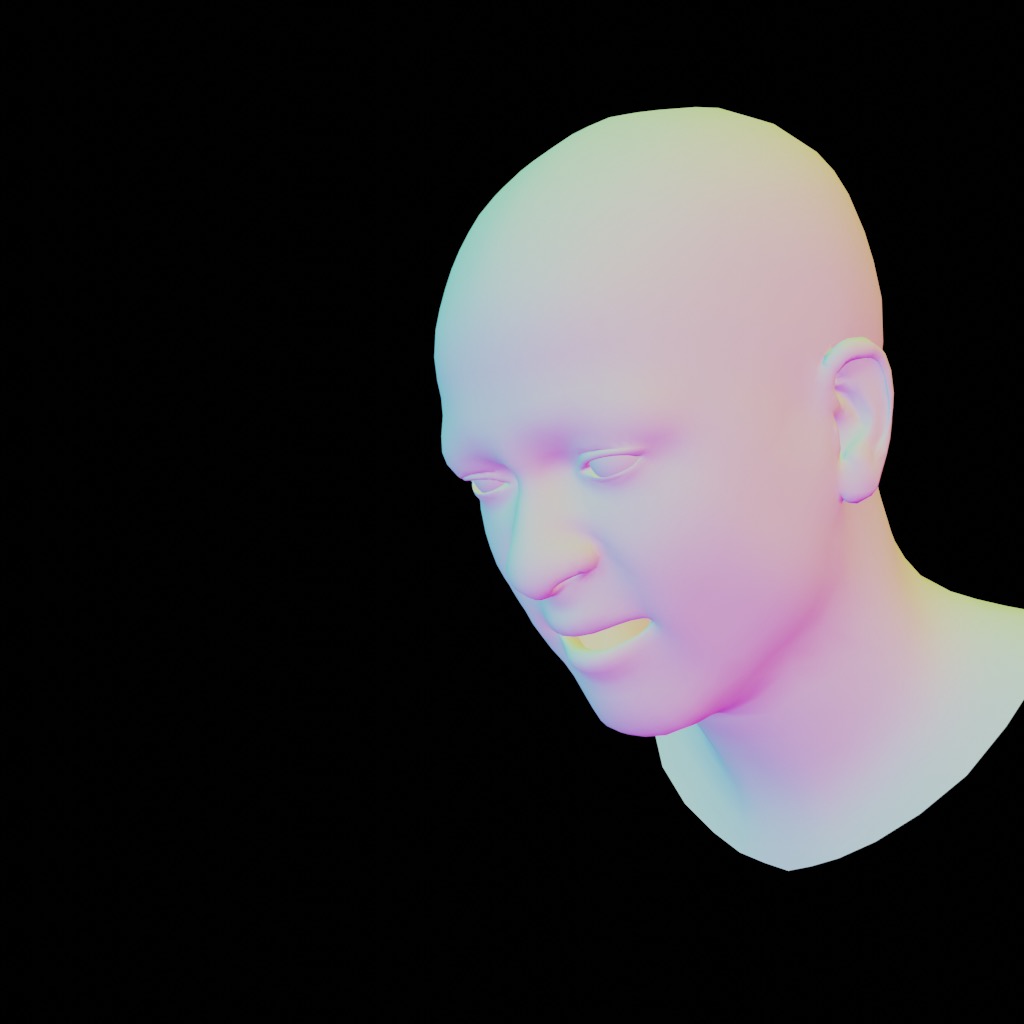}
    \includegraphics[width=0.23\columnwidth]{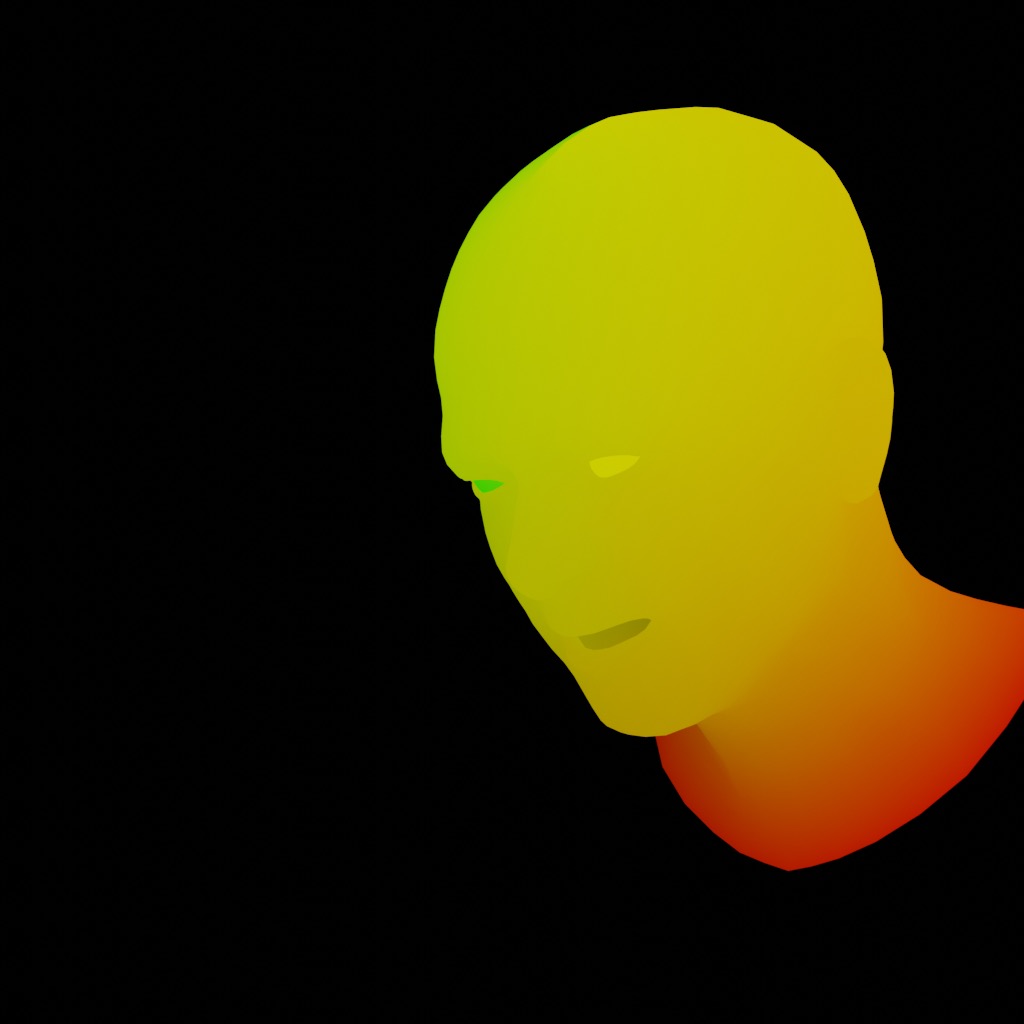}
    \includegraphics[width=0.23\columnwidth]{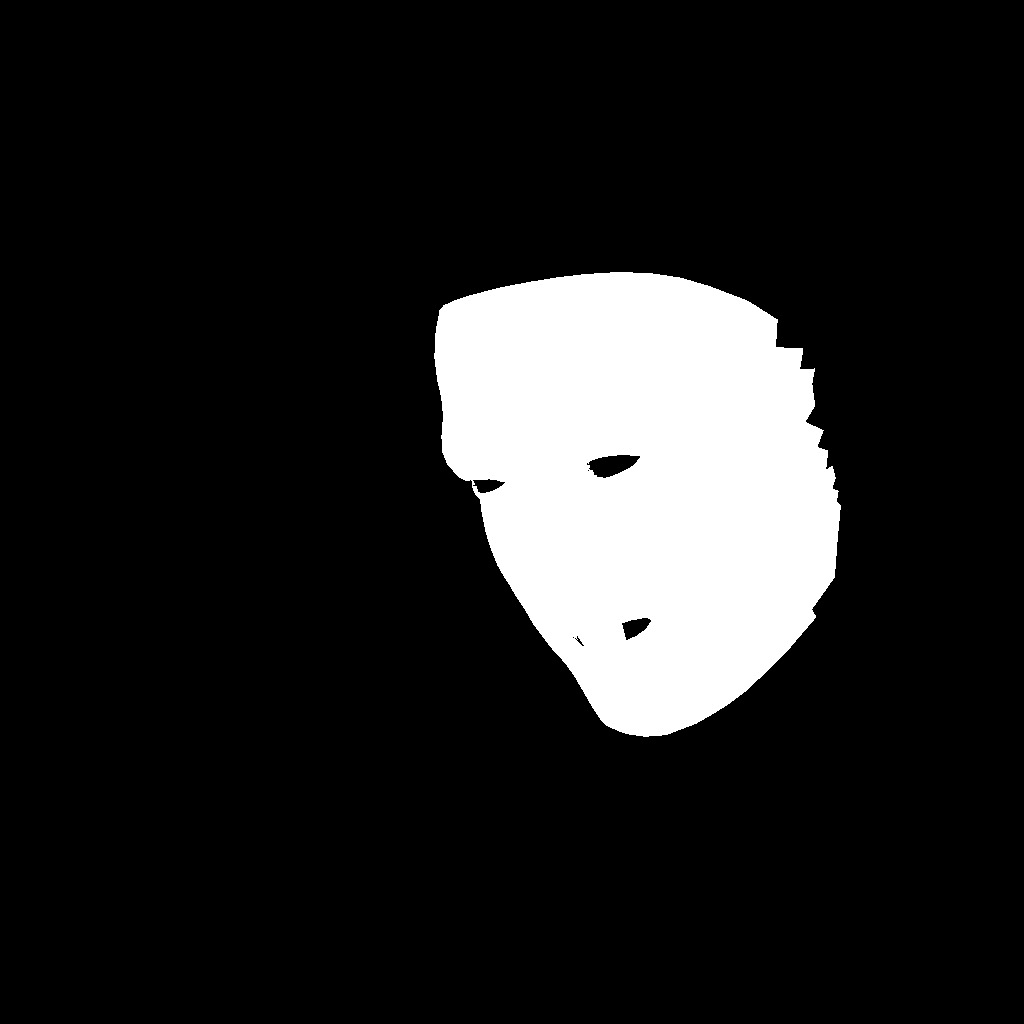}
    \includegraphics[width=0.23\columnwidth]{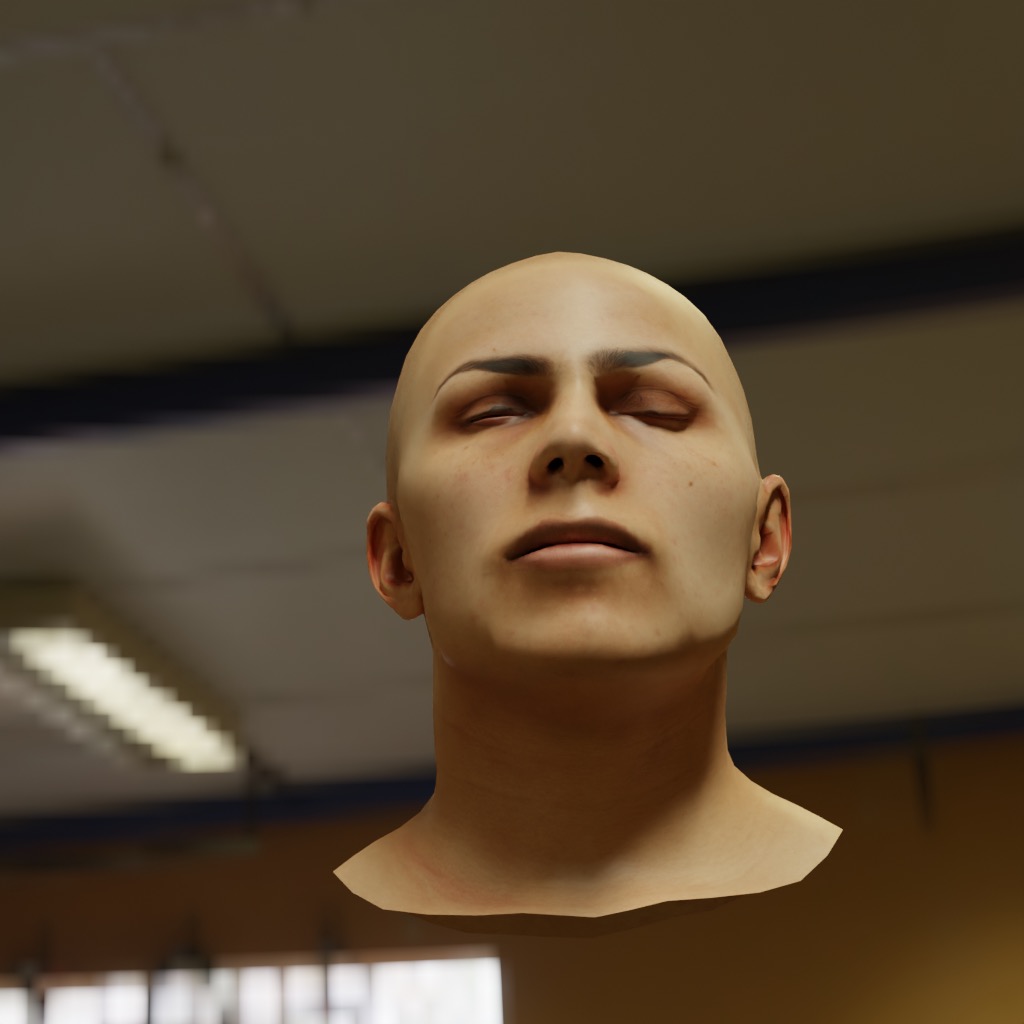}
    \includegraphics[width=0.23\columnwidth]{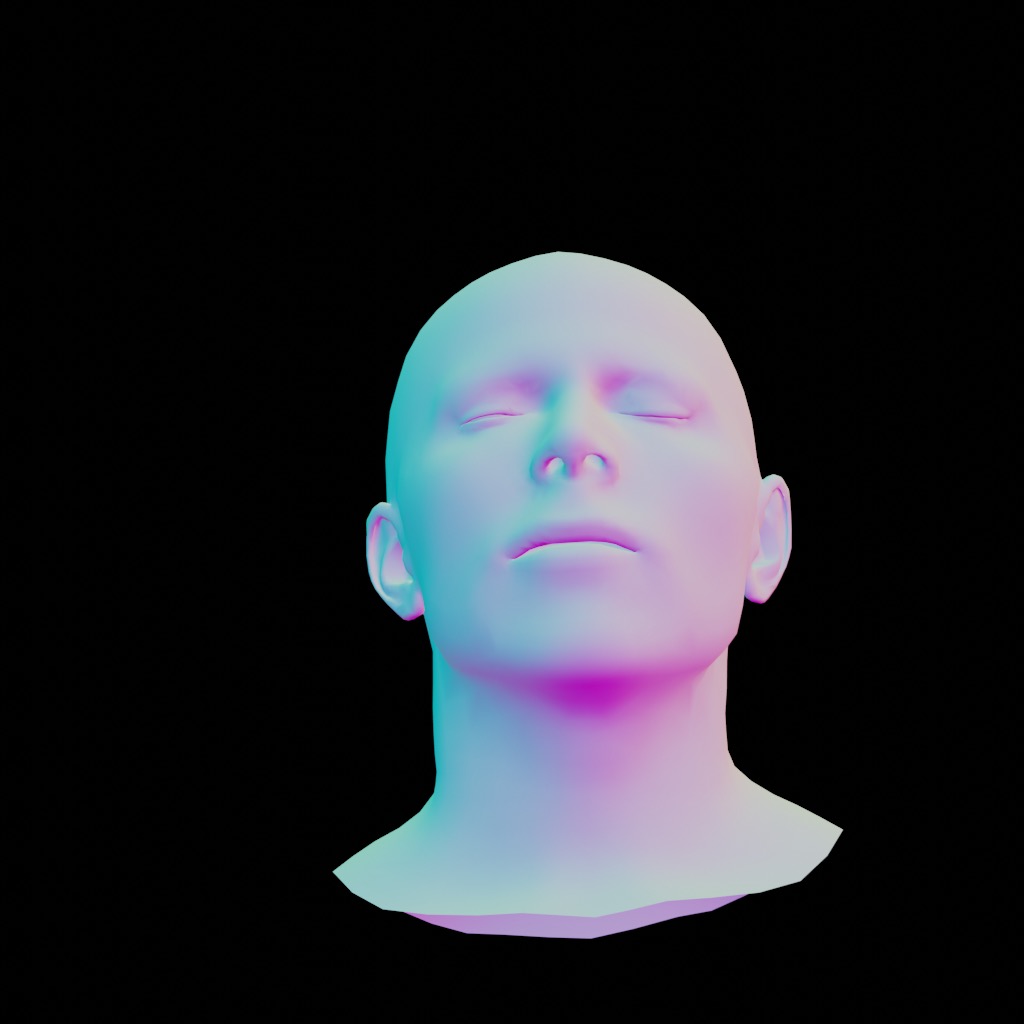}
    \includegraphics[width=0.23\columnwidth]{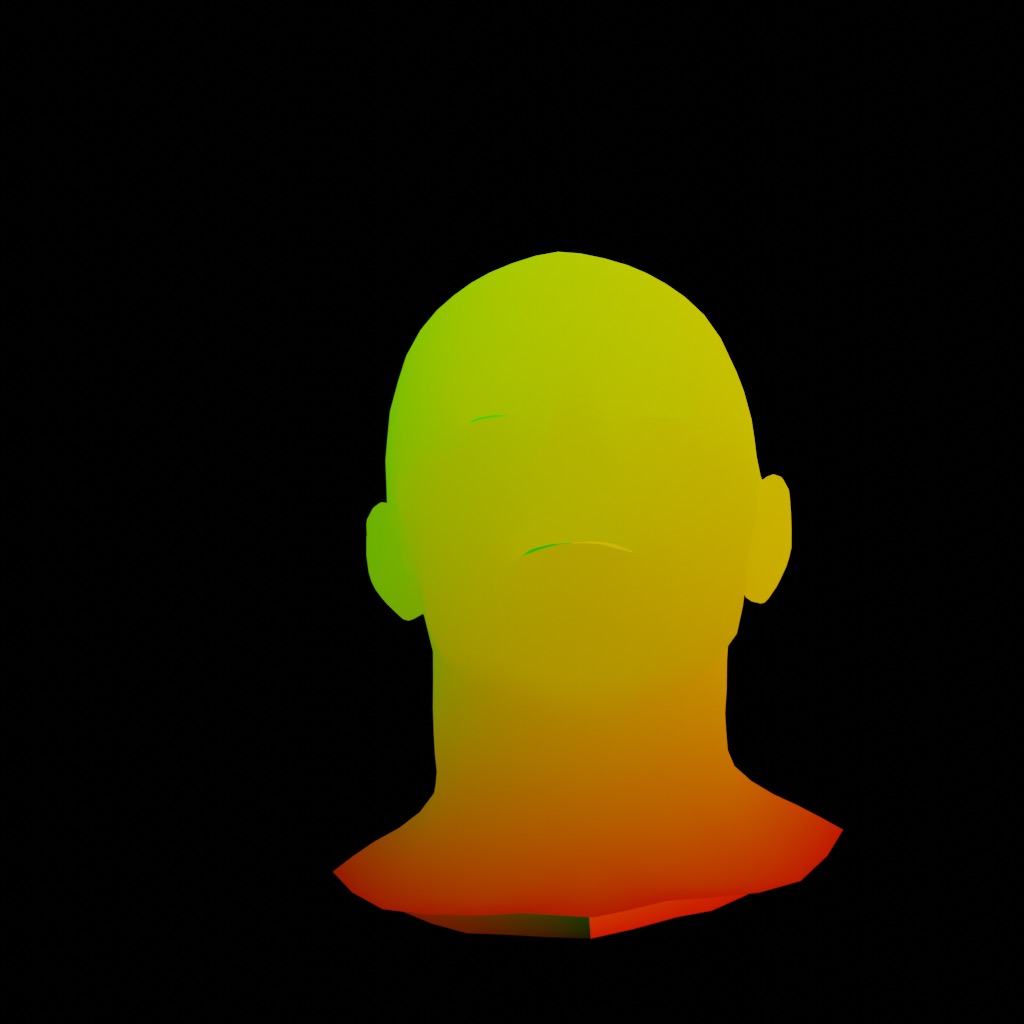}
    \includegraphics[width=0.23\columnwidth]{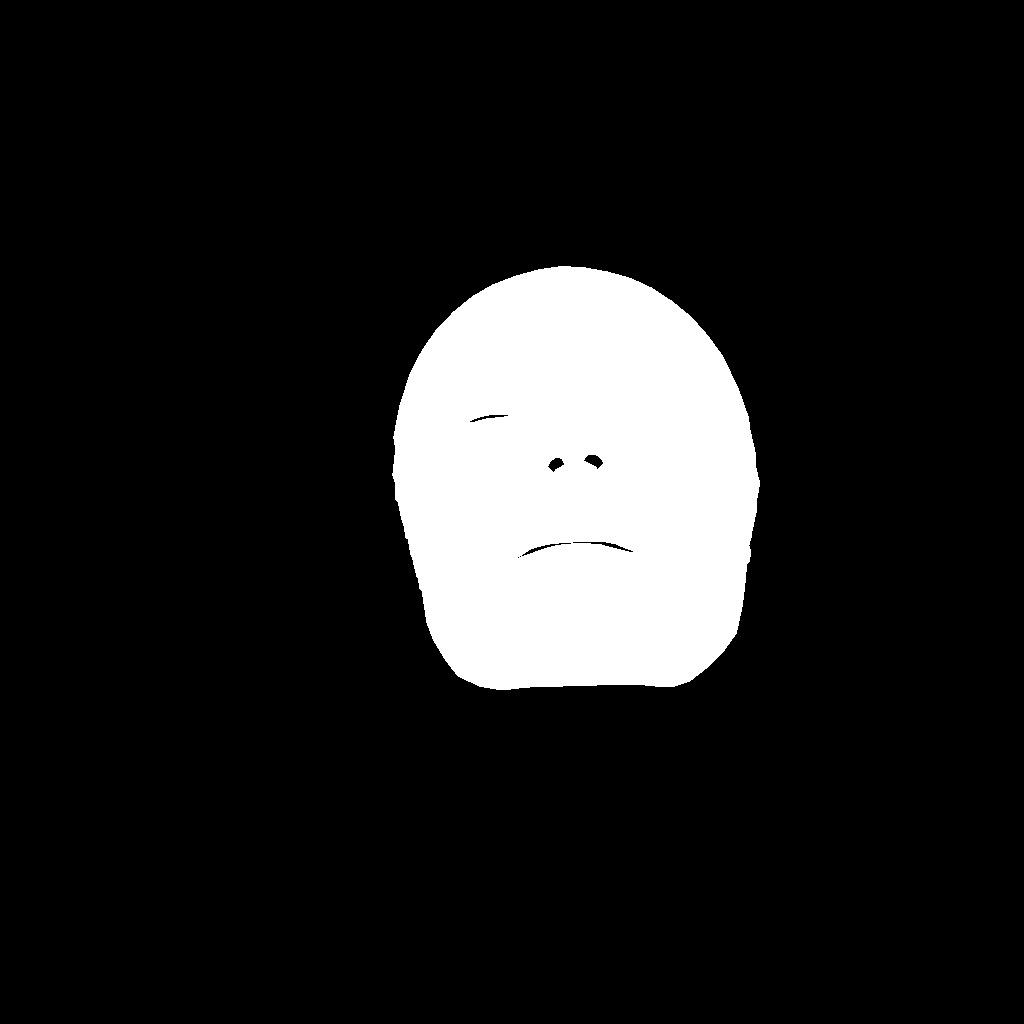}
    \includegraphics[width=0.23\columnwidth]{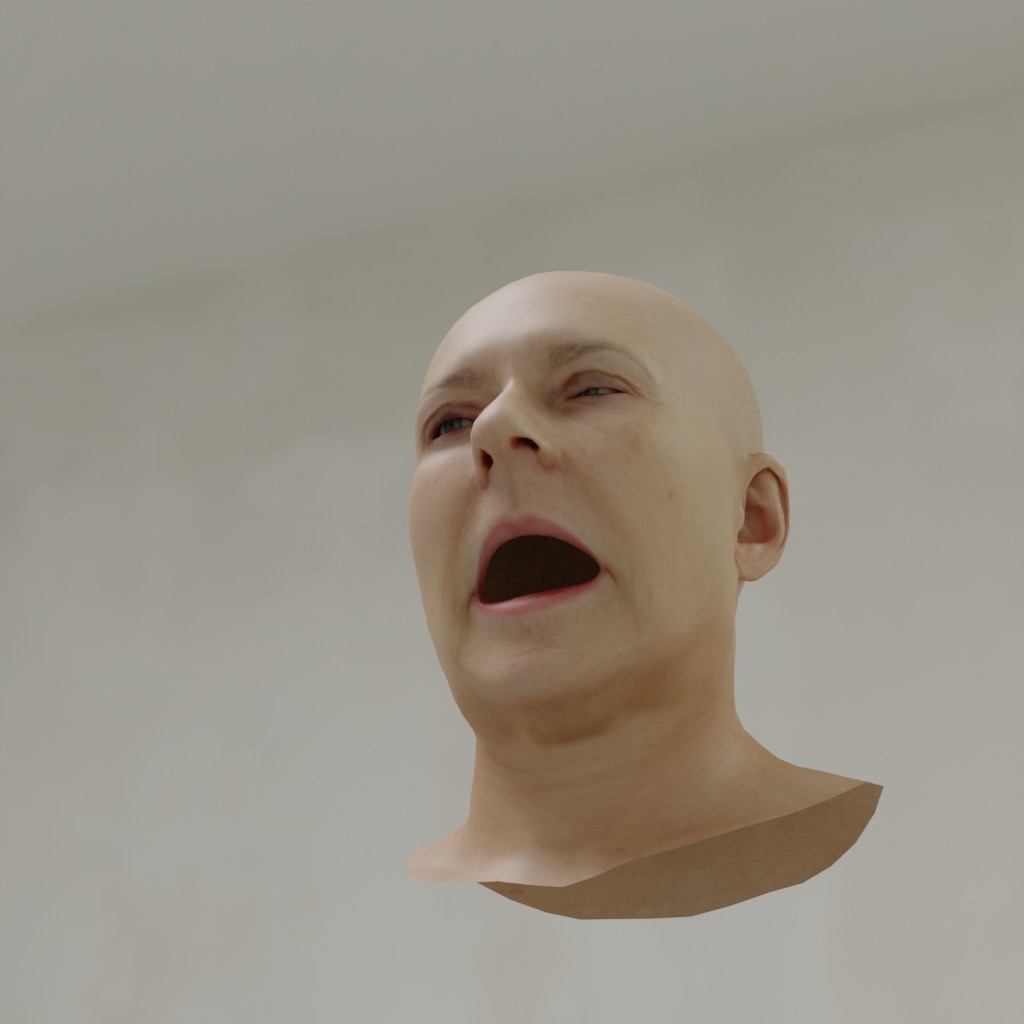}
    \includegraphics[width=0.23\columnwidth]{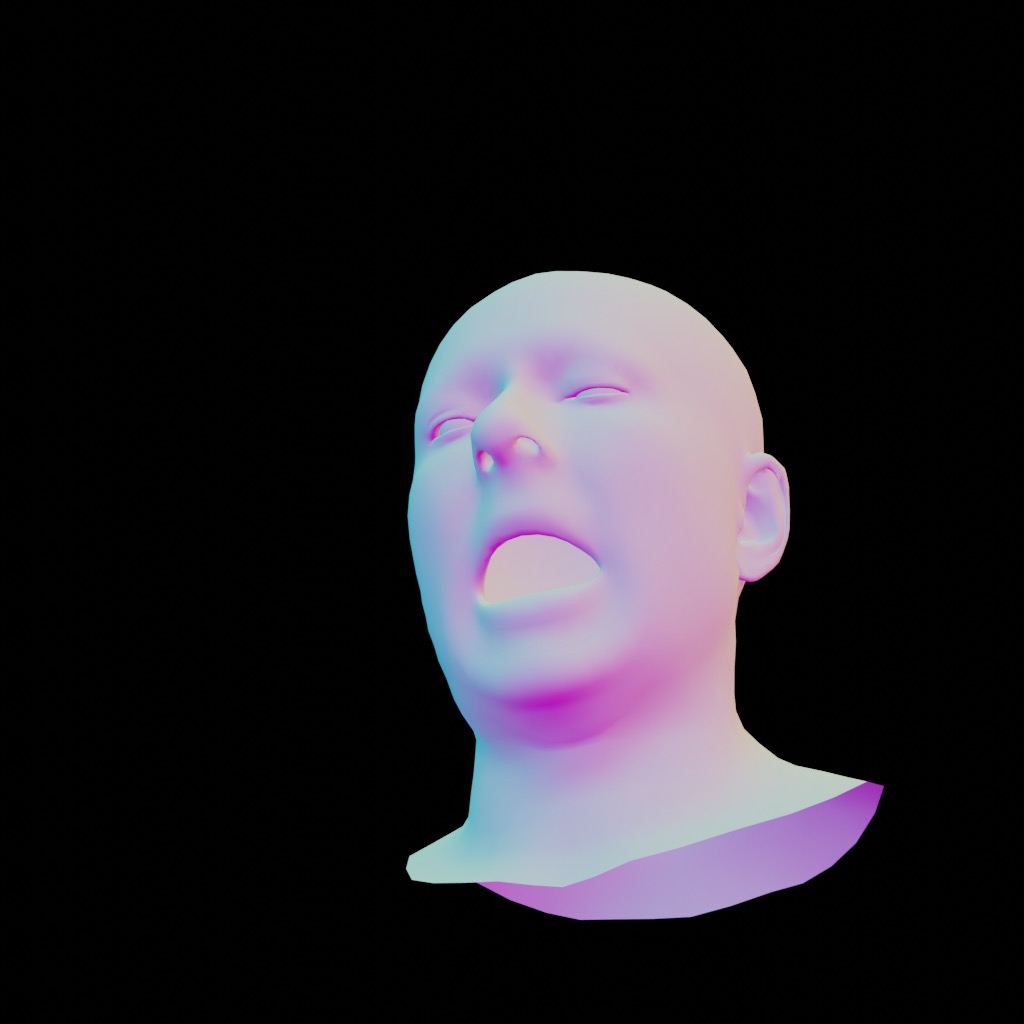}
    \includegraphics[width=0.23\columnwidth]{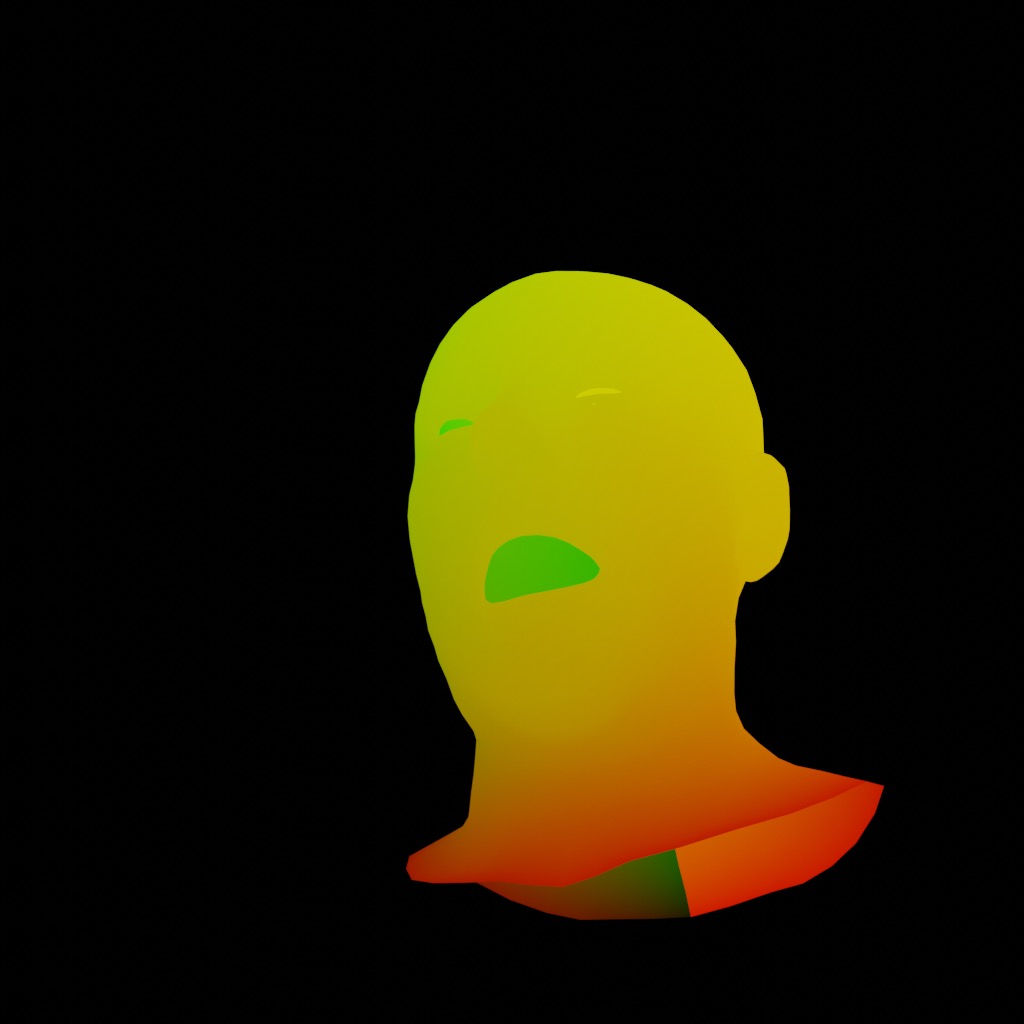}
    \includegraphics[width=0.23\columnwidth]{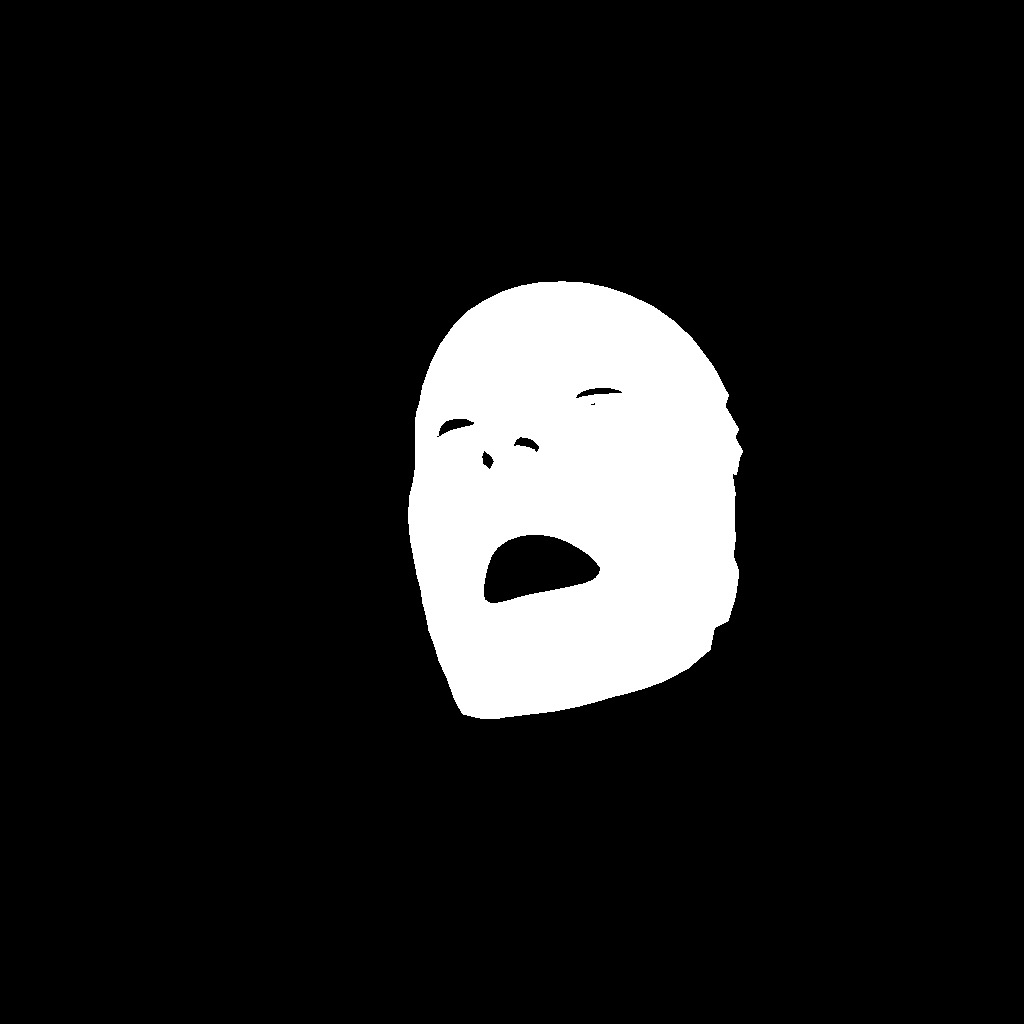}
    \includegraphics[width=0.23\columnwidth]{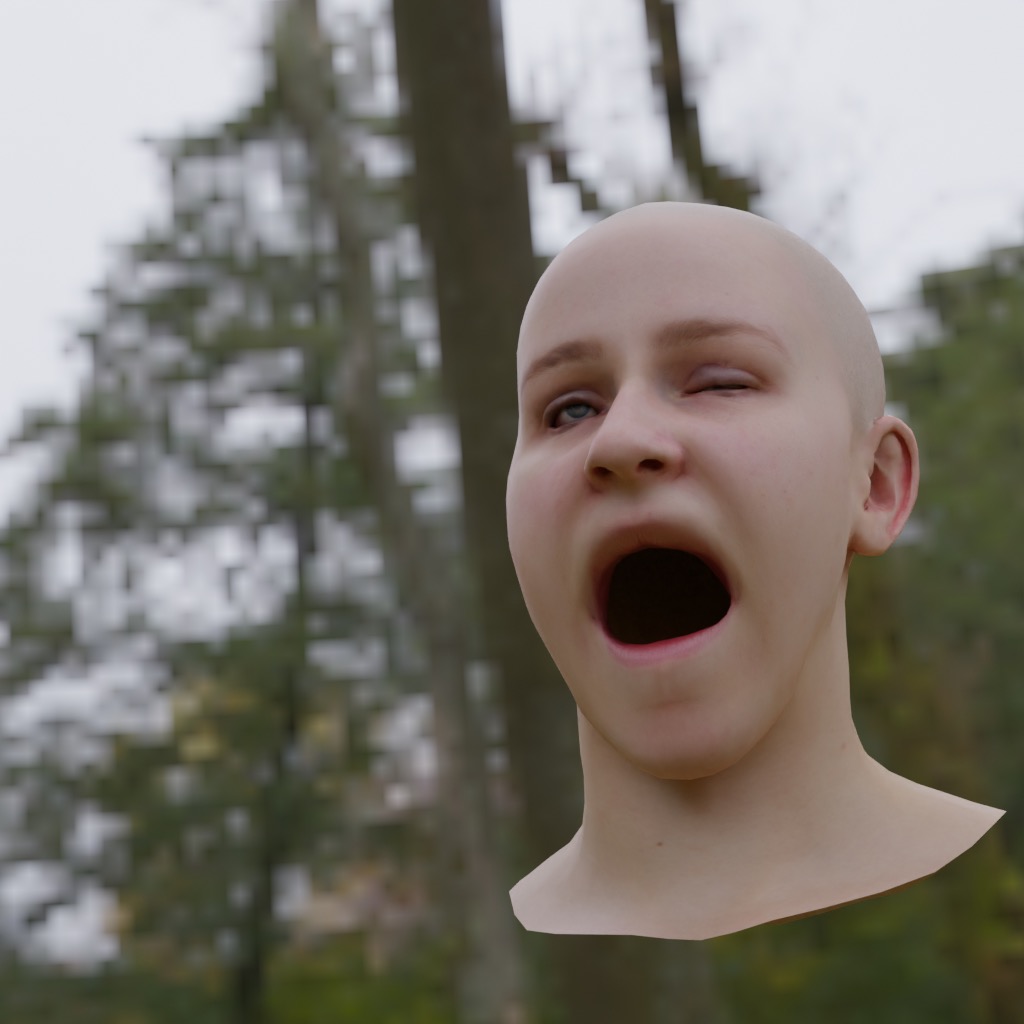}
    \includegraphics[width=0.23\columnwidth]{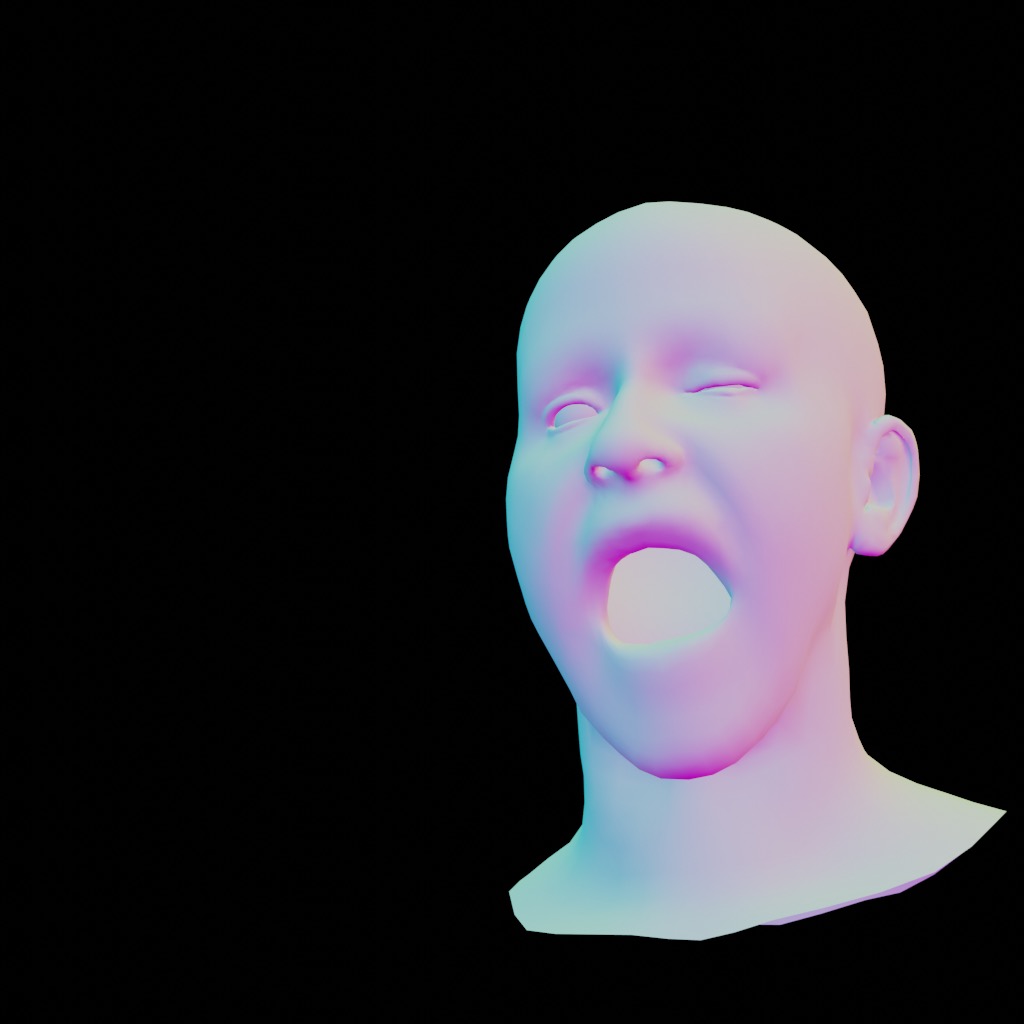}
    \includegraphics[width=0.23\columnwidth]{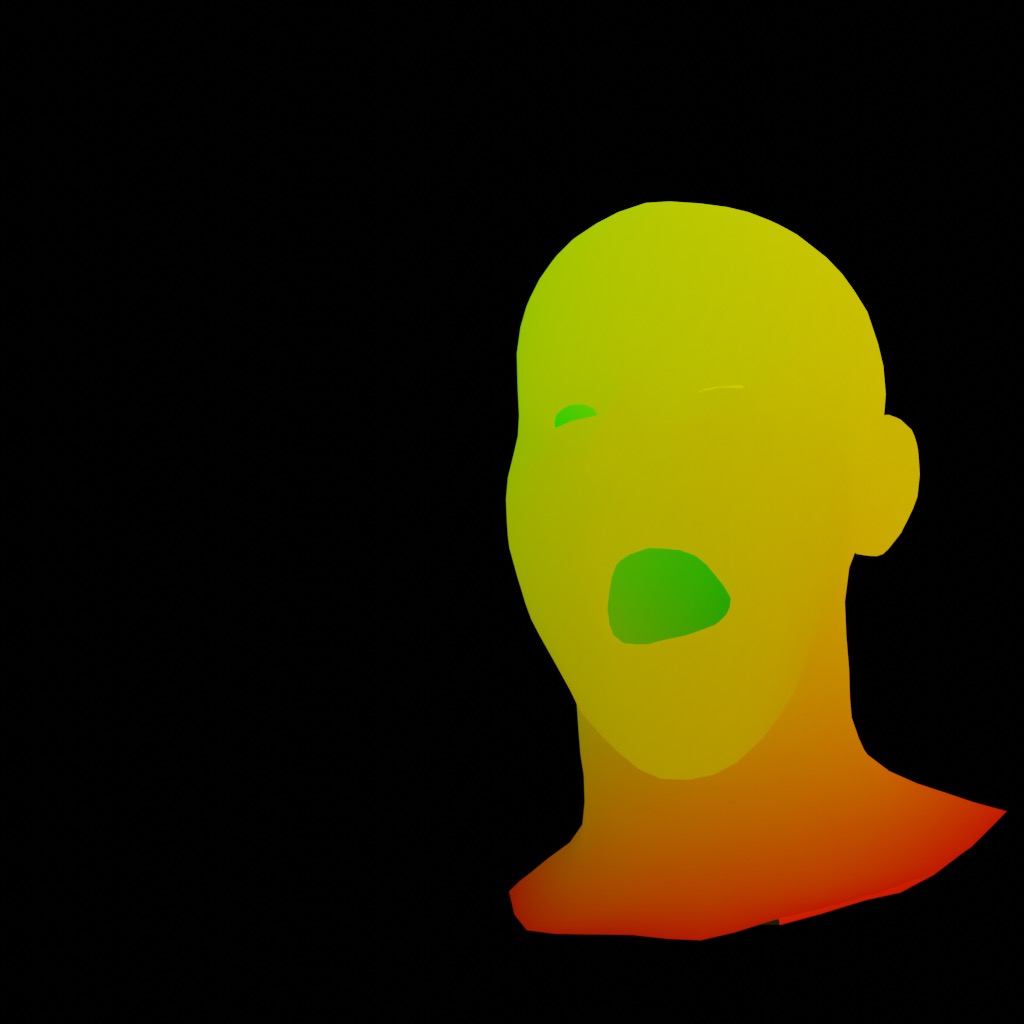}
    \includegraphics[width=0.23\columnwidth]{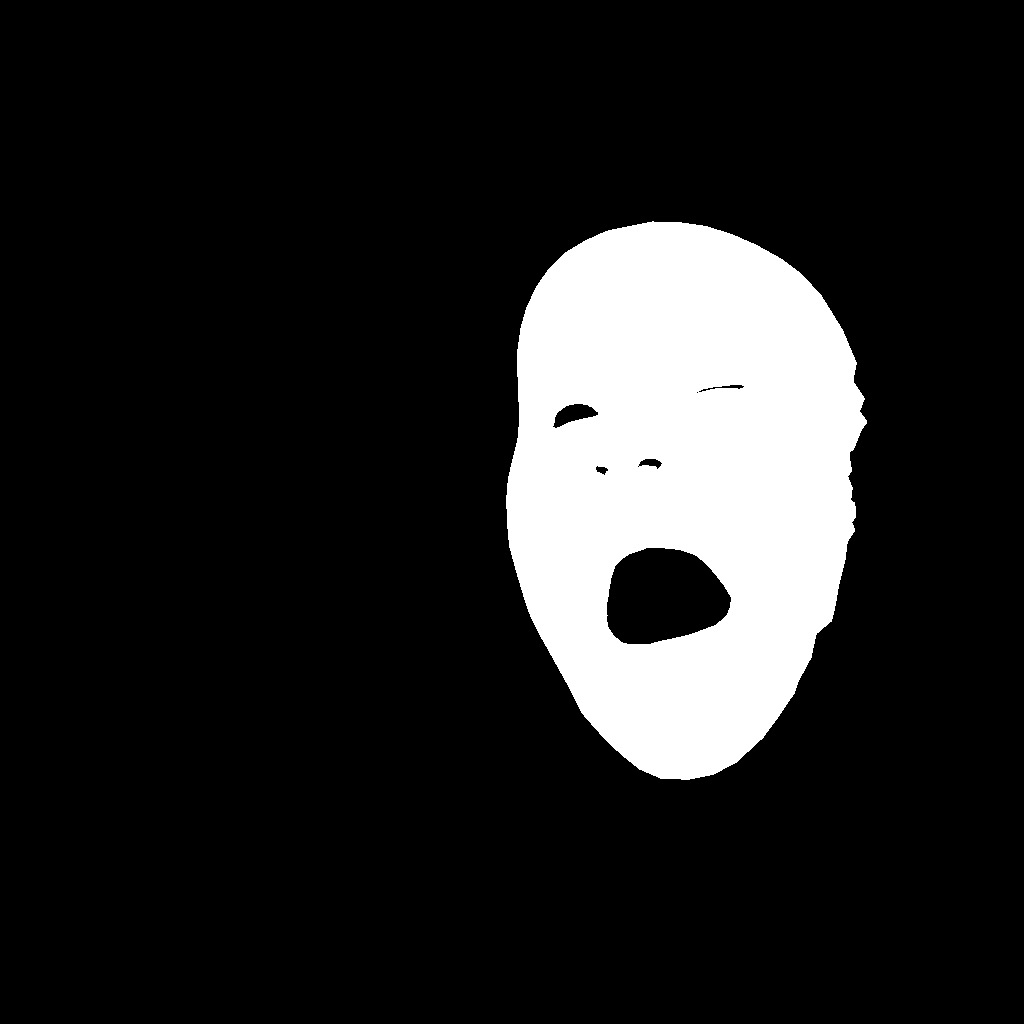}
    \includegraphics[width=0.23\columnwidth]{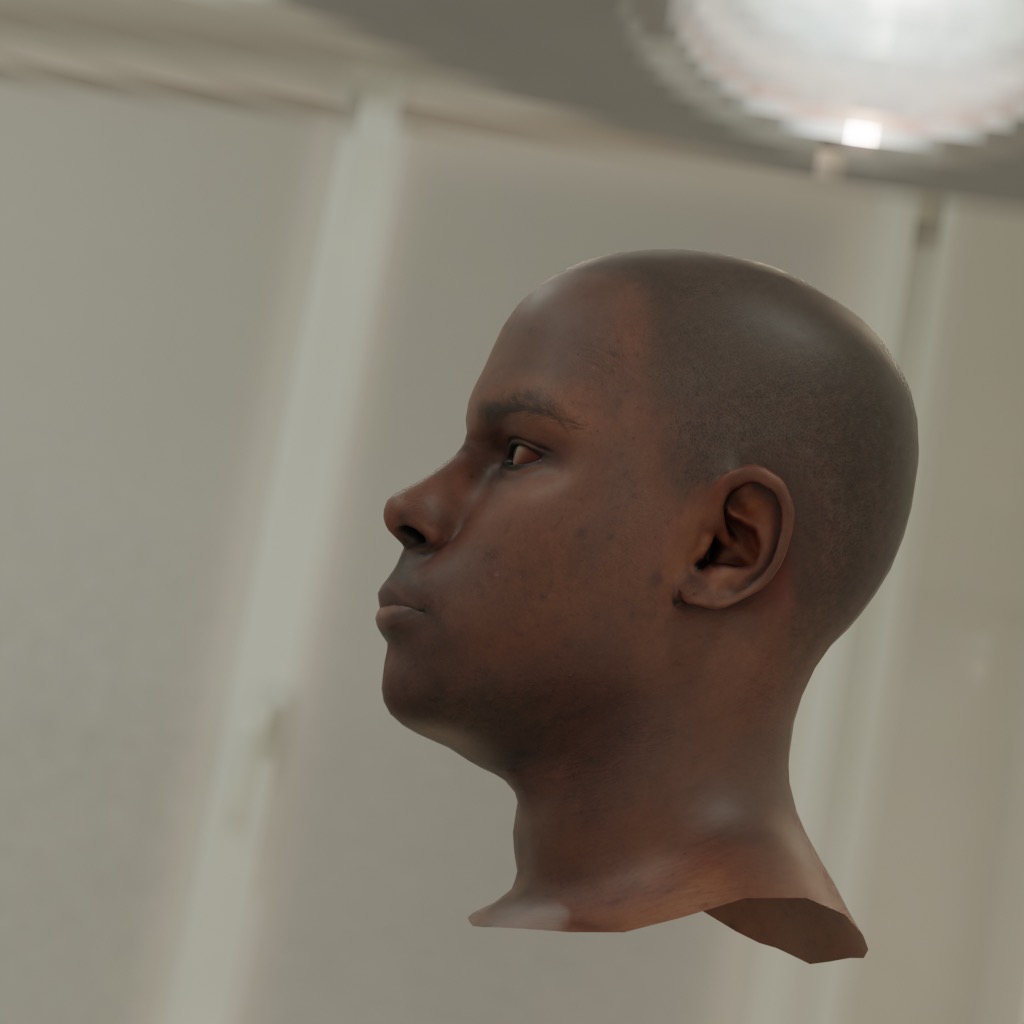}
    \includegraphics[width=0.23\columnwidth]{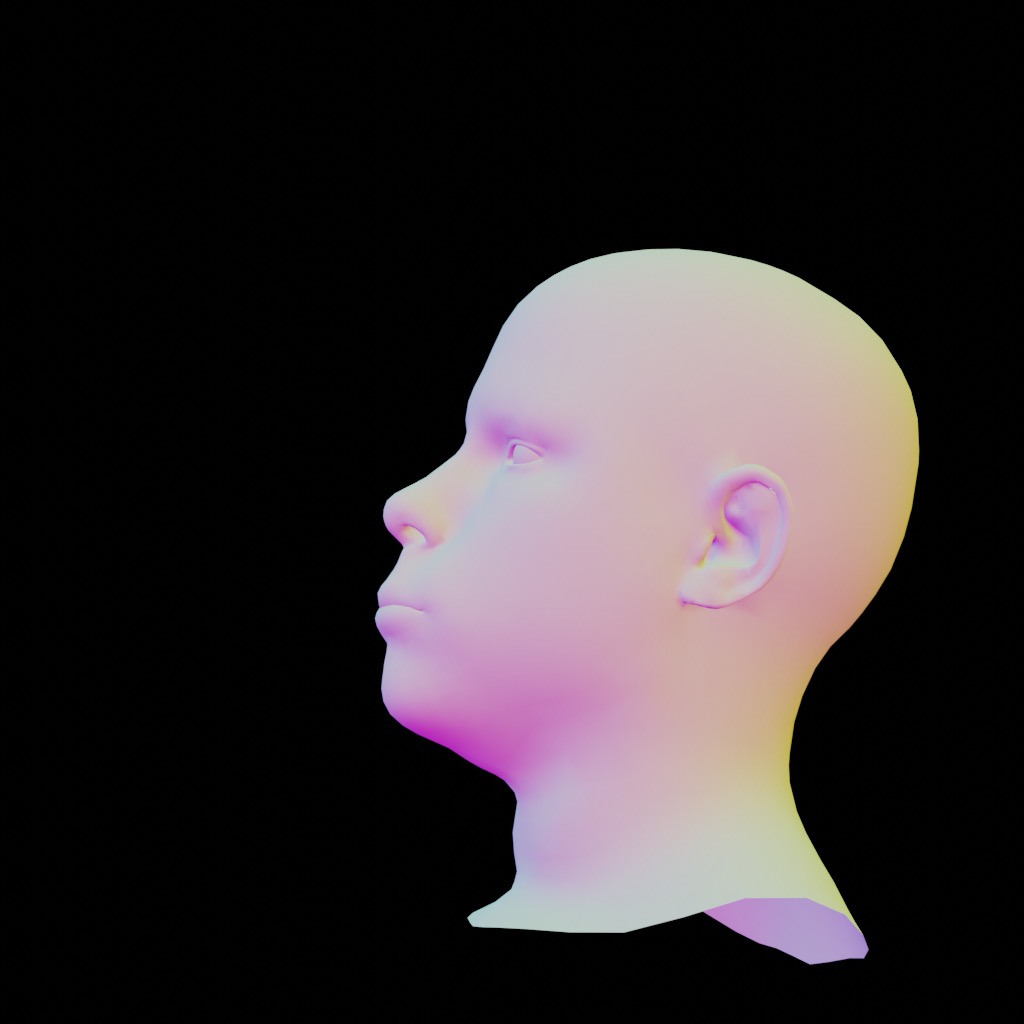}
    \includegraphics[width=0.23\columnwidth]{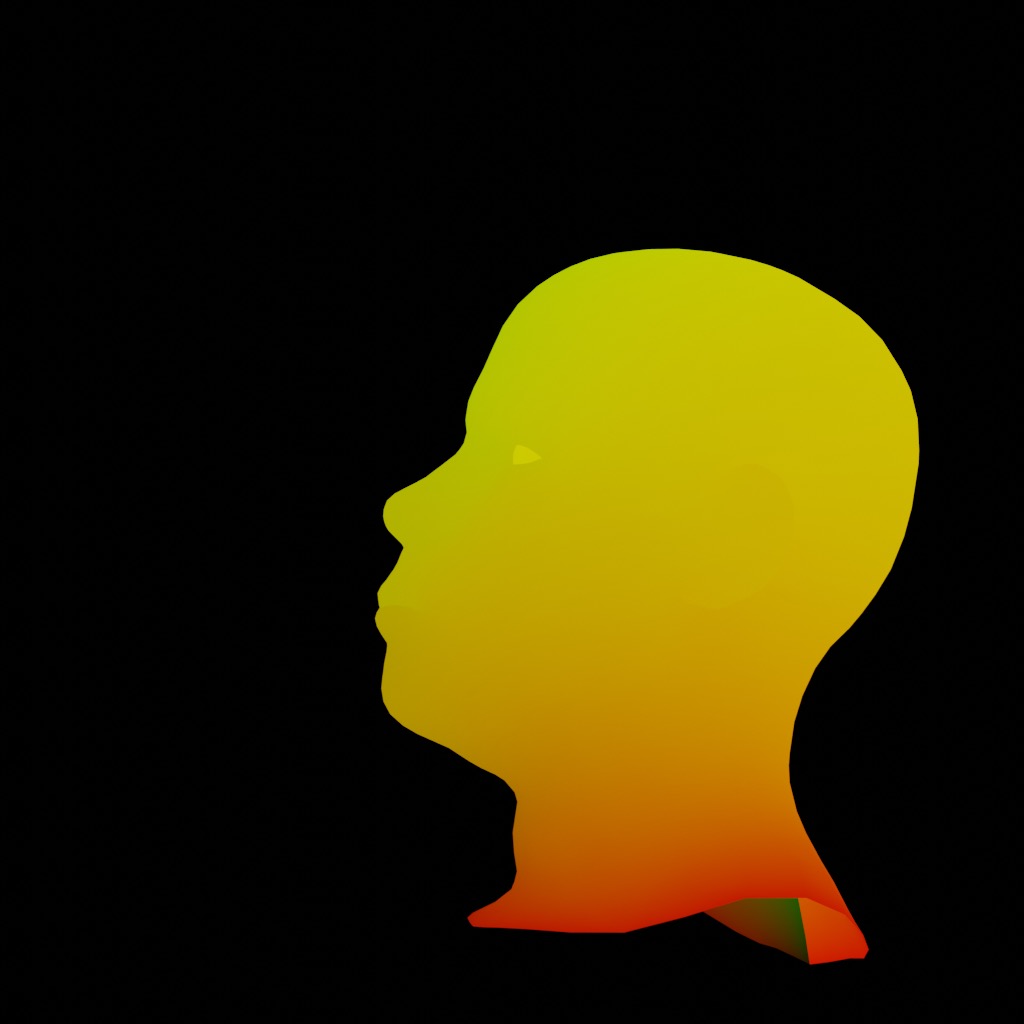}
    \includegraphics[width=0.23\columnwidth]{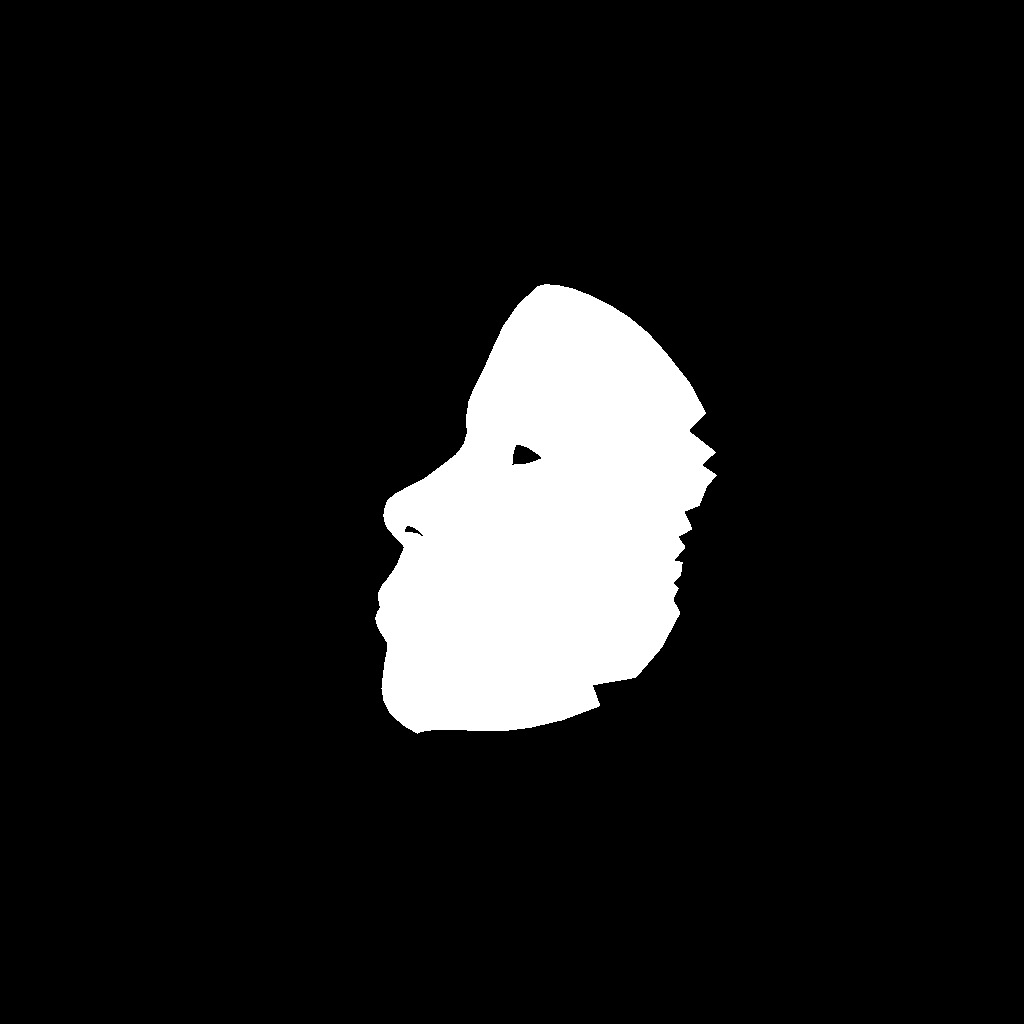}

    \caption{\textbf{Synthetic data.} From left to right: Path-traced RGB image, normal map, uv-map, binary face mask. We encourage the reader to zoom in on this figure in its digital form. }
    \label{fig:synthetic_images}
\end{figure*}

\section{Dense Landmark Tracker}
\label{sec:dlt}
\subsection{Synthetic Data With Blender}
\paragraph{Motivation.}
As described in the Method section of the main paper, our approach relies on an accurate dense landmark tracker for supervision. 
There are several ways to obtain high-quality dense landmark datasets:

(1) fitting 3D morphable models (3DDMs) to in-the-wild images, which is limited by the fact that unconstrained face reconstruction is still an open problem;

(2) deriving landmarks from registrations of high-quality 3D face scans captured in controlled studio environments, which we explicitly aim to avoid, as our goal is not to depend on registrations;

(3) synthetically generating data, which offers full control over the facial geometry, appearance, and deformation, and therefore provides perfect ground truth.

Given these considerations, we adopt the third option.

\paragraph{Generating Geometry.}
To generate the geometry we randomly sample a FLAME identity, a FLAME expression, FLAME jaw opening angle, eyeball rotation and the activation of a separate eye-blinking blendshapes. 
To generate hair, we use HAAR~\cite{Sklyarova2024haar} to sample 150 various hairstyles. 
For each generated face, we either keep it bald, or sample one of the hairstyles.

\paragraph{Albedo.}
We sample 54 (26 female and 28 male) very high quality textures purchased from 3D Scan Store, which cover all different skin tones and eye colors. 
These textures were re-topologized to be compatible with the FLAME UV-map. 
For each rendering, we uniformly sample one texture.

\paragraph{Illumination.}
For illumination, we employ high quality HDRI environment maps obtained from PolyHaven. They were captured in a variety of indoor and outdoor scenes and cover many possible illumination settings. 
For each rendering, we first sample one of 665 environment maps. 
Then, we randomly rotate the environment map about the vertical axis.

\paragraph{Camera position.}
We randomly sample the camera position to be in the frontal hemisphere of the rendered face. 
We also slightly vary the "look-at" point such that is not always looking at the center of the face. 
Furthermore, we perturb the camera distance and the focal length. 

\paragraph{Output.}
We use Blender \cite{blender} to render all the images. First, the Cycles path tracer renders the RGB image of the scene generated with the process described above. 
Then we remove hair (if the sample had any) and switch to the EEVEE renderer to render UV-map, normal map, and flame regions masks. 
The resulting images along with the mesh and camera parameters, provide perfect GT to train our dense landmark tracker. 
Figure \ref{fig:synthetic_images} shows a few example outputs.

\paragraph{Limitations.}
While our data is sufficient for our tasks, it does have a few limitations. First, FLAME models the face and neck only, which results in a floating-head appearance. Although this is not realistic, we do not observe any negative impact on model performance. Second, we do not model the mouth cavity, teeth, or tongue, as these components are not included in FLAME. Finally, our synthetic data does not include facial hair or accessories such as glasses, which may lead to reduced performance when these are present. Nevertheless, for our primary use case, namely landmark tracking for multi-view facial capture, subjects typically do not exhibit such variations, so this limitation has minimal practical impact.

\subsection{2D Landmark Tracker}
\label{sec:tracker_details}

Using the previously described synthetic dataset, we train a dedicated 2D landmark tracker. We employ a Vision Transformer (ViT-Large) backbone~\cite{dosovitskiy2020vit}, initialized with DINOv3 weights~\cite{DINOv3_2025}. To maintain the robust semantic features learned during pre-training while adapting to our task, we freeze the backbone parameters and inject Low-Rank Adaptation (LoRA)~\cite{hu2021lora} modules into the attention and feed-forward blocks. Specifically, we apply LoRA with rank $r=8$ and $\alpha=16$ to the query, key, value, and projection matrices in the attention layers, as well as the fully connected layers in the MLP blocks. 
The features are aggregated via global average pooling and passed to a lightweight regression head consisting of a LayerNorm and a final Linear projection that predicts the 2D normalized coordinates of the FLAME mesh vertices.

\begin{figure}[t]
    \centering
    \includegraphics[width=1.0\linewidth]{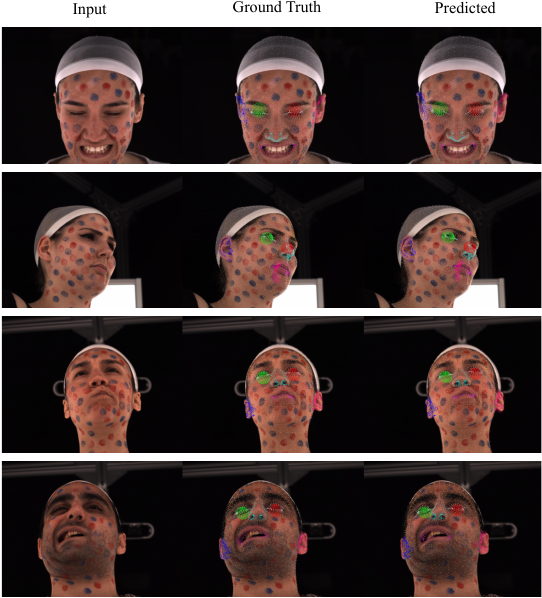}
    \caption{\textbf{Qualitative evaluation of the 2D landmark tracker on the FaMoS dataset.} From left to right: the input RGB image, the Ground Truth landmarks (obtained by projecting the available registration), and the landmarks predicted by our tracker. The vertices are color-coded by semantic region (e.g., green/red for eyes, pink for lips, blue for ears). Despite being trained exclusively on synthetic data, the tracker generalizes robustly to real-world images, accurately capturing dense correspondence across challenging expressions and poses.}
    \label{fig:synthetic_2d}
\end{figure}

\paragraph{Training and Evaluation.}
We train the tracker for 100 epochs on our generated synthetic dataset. The input images are resized to $512 \times 512$ pixels. We use the AdamW optimizer~\cite{loshchilov2017decoupled} with a learning rate of $10^{-4}$ and weight decay of $10^{-4}$. The learning rate is scheduled using a cosine annealing strategy with a linear warm-up. The model is supervised using a Mean Squared Error (MSE) loss between the predicted 2D coordinates and the ground-truth projected vertices from the synthetic rendering. To improve generalization to real-world data, we apply aggressive geometric and photometric image augmentations during training. 


To validate our 2D tracker, we perform a qualitative comparison against projected FLAME registrations (from a classical pipeline) on a subset of the FaMoS dataset. Figure~\ref{fig:synthetic_2d} visualizes the output of our tracker—trained exclusively on synthetic data—when applied to real-world samples. In our MOCHI pipeline, these 2D predictions serve as semantic regularizers, anchoring difficult regions such as the eyes and lips. As such, the tracker does not need to achieve pixel-perfect accuracy, but rather provide stable, semantically meaningful guidance.

\clearpage
\clearpage